\documentclass{article}

\usepackage[margin=1in]{geometry}

\usepackage{microtype}
\usepackage{graphicx}
\usepackage{subfigure}
\usepackage{booktabs} %

\usepackage{amsmath}

\usepackage[breaklinks,colorlinks=true,allcolors=blue,hyperfootnotes=false]{hyperref}
\usepackage[capitalise,noabbrev]{cleveref}
\crefname{equation}{}{}

\RequirePackage{mathtools}
\RequirePackage{dsfont}
\RequirePackage{delimset}

\newcommand{\ceil}[2][*]{\delim\lceil\rceil#1{#2}}

\newcommand{\indEvent}[2][*]{\mathds{1}{\brk[c]#1{#2}}}
\newcommand{\indFunc}[2][\infty]{\mathds{I}_{\brk[s]{#2,#1}}}
\newcommand{\indFuncAt}[3][\infty]{\indFunc[#1]{#2}\brk*{#3}}

\newcommand{\EE}[2][]{\mathbf{E}_{#1}{#2}}
\newcommand{\EEBrk}[2][*]{\mathbf{E}\delim{[}{]}#1{#2}}
\newcommand{\RR}[1][]{\mathbb{R}^{#1}}

\newcommand{\PP}[2][*]{\mathbb{P}\brk#1{#2}}

\ifx\argmax\undefined
    \DeclareMathOperator*{\argmax}{arg\,max}
\fi
\ifx \argmin\undefined
    \DeclareMathOperator*{\argmin}{arg\,min}
\fi

\DeclarePairedDelimiterX\setDef[1]\lbrace\rbrace{\def\given{\;\delimsize\vert\;}#1}
\newcommand{\seqDef}[4][*]{\brk[c]#1{#2}_{#3}^{#4}}

\def\noi{\noindent}
\usepackage{amsthm}
\usepackage{amssymb}
\usepackage{thmtools}

\declaretheoremstyle[
	    spaceabove=\topsep, 
	    spacebelow=\topsep, 
	    bodyfont=\normalfont\itshape,
    ]{theorem}

\declaretheorem[style=theorem,name=Theorem]{theorem}

\declaretheoremstyle[
	    spaceabove=\topsep, 
	    spacebelow=\topsep, 
	    bodyfont=\normalfont,
    ]{definition}

\declaretheoremstyle[
        spaceabove=\topsep, 
        spacebelow=\topsep, 
        bodyfont=\normalfont,
        notefont=\normalfont\bfseries,
        notebraces={}{},
        qed=$\blacksquare$, 
    ]{proofstyle}
\declaretheorem[style=proofstyle,numbered=no,name=Proof]{proof}

\declaretheorem[style=theorem,name=Lemma]{lemma}

\declaretheorem[style=theorem,name=Proposition]{proposition}

\declaretheorem[style=theorem,numbered=no,name=Theorem]{theorem*}
\declaretheorem[style=theorem,numbered=no,name=Lemma]{lemma*}
\declaretheorem[style=theorem,numbered=no,name=Corollary]{corollary*}
\declaretheorem[style=theorem,numbered=no,name=Proposition]{proposition*}
\declaretheorem[style=theorem,numbered=no,name=Claim]{claim*}
\declaretheorem[style=theorem,numbered=no,name=Fact]{fact*}
\declaretheorem[style=theorem,numbered=no,name=Observation]{observation*}
\declaretheorem[style=theorem,numbered=no,name=Conjecture]{conjecture*}

\declaretheorem[style=definition,name=Definition]{definition}
\declaretheorem[style=definition,name=Remark]{remark}

\declaretheorem[style=definition,name=Assumption]{assumption}

\declaretheorem[style=definition,numbered=no,name=Definition]{definition*}
\declaretheorem[style=definition,numbered=no,name=Remark]{remark*}
\declaretheorem[style=definition,numbered=no,name=Example]{example*}
\declaretheorem[style=definition,numbered=no,name=Question]{question*}
\declaretheorem[style=definition,numbered=no,name=Assumption]{assumption*}

\newcommand{\rev}[2]{#1}

\newcommand{\simple}{simple}
\newcommand{\Simple}{Simple}

\newcommand{\CVAR}[1][\alpha]{CVaR_{#1}}
\newcommand{\VAR}[1][\alpha]{VaR_{#1}}
\newcommand{\UBSR}[1][\alpha]{UBSR_{#1}}
\newcommand{\EDRMabbrv}{EDPM}
\newcommand{\UUCB}{\RHat-UCB}

\newcommand{\XtPi}[1][t]{X_{{\policy}, {#1}}}
\newcommand{\Xti}[2][i]{X_{{#1}, {#2}}}
\newcommand{\Xt}[1][t]{X_{{#1}}}

\newcommand{\XtPiOrd}[2][\policy]{X^*_{{#1},{#2}}}

\newcommand{\policiesSet}{\Pi}

\newcommand{\policy}[1][]{\pi^{#1}}
\newcommand{\policyAt}[2][]{\policy[#1]_{#2}}
\newcommand{\statPolicy}[1][p]{\pi^{#1}}
\newcommand{\statPolicyAt}[2][p]{\statPolicy[#1]_{#2}}
\newcommand{\optPolicyAt}[1]{\policyAt[*]{#1}\brk*{\infty}}
\newcommand{\optPolicy}{\policy[*]\brk*{\infty}}
\newcommand{\optPolicyTAt}[1]{\policyAt[*]{#1}\brk*{T}}
\newcommand{\optPolicyT}[1][T]{\policy[*]\brk*{#1}}

\newcommand{\Rbase}[1][]{\tilde{U}^{#1}}
\newcommand{\Rt}[1][]{\Rbase[#1]_{t}}
\newcommand{\RtFunc}[2][]{\Rt[#1]\brk*{#2}}
\newcommand{\RPi}[1][\policy]{{U}_{#1}}
\newcommand{\RPiNamed}[2][\policy]{\RPi[{#1}]^{#2}}
\newcommand{\RHat}[1][]{{U}^{#1}}
\newcommand{\RHatFunc}[2][]{\RHat[{#1}] \brk1{#2}}
\newcommand{\DRHat}[1][]{\partial \RHat[{#1}]}
\newcommand{\DRHatAt}[2][]{\DRHat[{#1}] \brk1{#2}}
\newcommand{\DRHatAtOn}[3][]{\DRHatAt[{#1}]{#2} \cdot {#3}}

\newcommand{\FHatt}[1][t]{\hat{F}_{#1}}
\newcommand{\FHattFunc}[2][t]{\hat{F}_{#1} \brk*{#2}}
\newcommand{\FHatPi}[2][\policy]{\hat{F}_{#2}^{#1}}
\newcommand{\FProxPi}[2][\policy]{{F}_{#2}^{#1}}
\newcommand{\Fp}[1][p]{F_{#1}}
\newcommand{\FpOpt}{F_{p^*}}
\newcommand{\Fi}[1][i]{F^{\brk*{{#1}}}}
\newcommand{\FiOpt}{F^{(i^*)}}

\newcommand{\simplex}{\Delta}

\newcommand{\funcSpace}{L_{\norm{\cdot}}}
\newcommand{\DistSet}{\mathcal{D}}
\newcommand{\DistSetDelta}{\DistSet^{\Delta}}

\newcommand{\EmpDistSetT}{\hat{\DistSet}_t}

\newcommand{\Filtration}{\mathcal{H}}
\newcommand{\FiltrationAt}[1]{\Filtration_{#1}}
\newcommand{\actionSet}{\mathds{K}}

\newcommand{\piAt}[2][i]{\hat{p}_{#1}\brk*{#2}}
\newcommand{\piT}[1][i]{\piAt[#1]{T}}

\newcommand{\tauIAt}[2][i]{\tau_{#1} \brk*{#2}}
\newcommand{\tauI}[1][i]{\tauIAt[#1]{T}}

\newcommand{\regret}[1][\policy]{R_{#1}\brk*{T}}
\newcommand{\pseudoRegret}[1][\policy]{\bar{R}_{#1}\brk*{T}}

\newcommand{\contMod}{\omega}
\newcommand{\contModFunc}[1]{\contMod \brk1{#1}}
\newcommand{\polyContModCoeff}{b}
\newcommand{\polyContModDeg}{q}
\newcommand{\lipConst}{L}
\newcommand{\pSmooth}{\beta}
\newcommand{\concentrationConst}{\upsilon}
\newcommand{\levelSet}[2][\alpha]{L_{#1}\brk*{#2}}

\newcommand{\phiFunc}{\phi}
\newcommand{\phiFuncAt}[1]{\phiFunc\brk*{#1}}
\newcommand{\phiInvFunc}{\phi^{-1}}
\newcommand{\phiInvFuncAt}[1]{\phiInvFunc\brk*{#1}}
\newcommand{\Di}{\Delta_i}

\newcommand{\linFunctionals}{L\brk*{\funcSpace,\RR}}

\newcommand{\varBa}{b_{\alpha}}
\newcommand{\varMa}{M_{\alpha}}

\newcommand{\ucbEvent}[2]{V_{#1}^{#2}}
\newcommand{\ucbEventInv}[2]{\overline{V_{#1}^{#2}}}

\newcommand{\distSetDiameter}{D}
\newcommand{\gapRatio}{\rho}

\newcommand{\varDeg}{d}

\newcommand{\pLinGap}{\eta}
\newcommand{\MT}{J_T}

\newcommand{\hComp}{h}
\newcommand{\hCompFunc}[2][1]{\hComp\brk#1{#2}}

\usepackage{xspace}
\usepackage{enumitem}

\usepackage{todonotes}
\usepackage{natbib}
\bibpunct{(}{)}{;}{a}{,}{,}

\newcommand{\acks}[1]{\section*{Acknowledgments.}#1}

\newcommand{\baseStyle}{}
\newcommand{\ABSTRACT}[1]{\begin{abstract} #1 \end{abstract}}
\newcommand{\KEYWORDS}[1]{\textbf{Key words:} #1}
\newenvironment{APPENDICES}[1]
    {
    \appendix
    }
    { 
    }

\title{A General Framework for Bandit Problems Beyond Cumulative Objectives}
\author{
    {\sf Asaf Cassel}
    \thanks{School of Computer Science, Tel Aviv University; \texttt{acassel@mail.tau.ac.il}.}
	\and
	{\sf Shie Mannor}
	\thanks{Faculty of Electrical Engineering, Technion, Israel Institute of Technology; \texttt{shie@technion.ac.il},
	Nvidia Research; \texttt{smannor@nvidia.com}}
	\and
	{\sf Assaf  Zeevi}
	\thanks{Graudate School of Business, Columbia University; \texttt{assaf@gsb.columbia.edu}}
}

\begin{document}

\ifdefined\baseStyle
    \maketitle
\fi

\ABSTRACT{
    The stochastic multi-armed bandit (MAB) problem is a common  model for sequential decision problems. In the standard setup, a decision maker has to choose at every instant between several competing arms, each of them provides a scalar random variable, referred to as a ``reward.'' Nearly all research on this topic considers the total cumulative reward as the criterion of interest. This work focuses on other natural objectives that cannot be cast as a sum over rewards, but rather more involved functions of the reward stream. Unlike the case of  cumulative criteria, in the problems we study here the oracle policy, that knows the problem parameters a priori and is  used to ``center" the regret, is not trivial. 
    We provide a systematic approach to such problems, 
    and derive general conditions under which the oracle policy is sufficiently tractable to facilitate the design of optimism-based (upper confidence bound) learning policies. These conditions elucidate an interesting interplay between the arm reward distributions and the performance metric.  Our main findings are illustrated for several commonly used objectives such as conditional value-at-risk, mean-variance trade-offs, Sharpe-ratio, and more.
}


\ifdefined\morStyle
    \maketitle
\fi

\section{Introduction} 
\label{sec:intro}

Consider a sequential decision making problem where at each stage  one of  $K$ independent alternatives is to be selected. When choosing alternative $i$ at stage $t$ (also referred to as time $t$), the decision maker receives a reward $\Xt$ that is distributed according to some \emph{unknown} distribution $\Fi$, $i=1,\ldots,K$ and is independent of $t$. (Where unambiguous, we avoid indexing $X_t$ with $i$, and leave that implicit; the information will be encoded in the policy that governs said choices, which will be detailed in what follows.) At time $t$, the decision maker has accumulated a vector of rewards $(\Xt[1],\ldots,\Xt)$. In our setting, performance criteria are defined by a function $\Rbase$ that maps the reward vector to a real-valued number.
As $\Rbase(X_1,\ldots,X_t)$ is a random quantity, we consider the accepted notion of expected performance, i.e., $\EE{\Rbase(\Xt[1],\ldots,\Xt)}$, assuming this expectation exists and is finite.  
An oracle, with full knowledge of the arms' distributions, will make a sequence of selections based on this information so as to maximize the expected performance criterion. This serves as a benchmark for any other policy which does not have such information a priori, and hence needs to learn it on the fly. The gap between the former (performance of the oracle) and the latter (performance of the policy) represents the usual notion of regret in the learning problem.   

The most ubiquitous  performance criterion in the literature concerns the long run average reward, which involves the empirical mean, $\Rbase[ave](\Xt[1], \ldots, \Xt) = \frac{1}{t}\sum_{s=1}^{t}\Xt[s]$. In this case, the oracle rule, that maximizes the expected value of the above, just samples from the distribution with the highest mean value, namely, it selects $i^* \in \argmax \{\int x d\Fi(x)\}$. Learning algorithms for such problems date back to Robbins' paper \cite{robbins1952some} and were extensively studied subsequent to that. In particular, the seminal work of \cite{lai1985asymptotically} establishes that the normalized (per epoch)  regret in this problem, when the arms are ``well separated,"  cannot be made smaller than $\mathcal{O}(\log T / T)$,  and there exist learning algorithms that achieve this regret by maximizing a confidence bound modification of the empirical mean. (When the arms are not well separated, equivalent statements hold with order-$1/\sqrt{T}$.) Since then, this class of policies has come to be known as UCB, or upper confidence bound policies.  Some strands of literature that have emerged from this include \cite{auer2002finite} (non-asymptotic analysis of  UCB-policies), \cite{maillard2011finite} (empirical confidence bounds or KL-UCB), \cite{agrawal2012analysis} (Thompson sampling based algorithms), and various works which consider an adversarial formulation (see, e.g., \cite{auer1995gambling}).

\textbf{Main research questions.}    
In  this paper we are interested in studying the above problem for more general {\it path dependent} objectives that are of interest beyond  just the vanilla average. Many of these objectives bear an interpretation as ``risk criteria'' insofar as they focus on a finer probabilistic nature of the primitive distributions than the mean,  and typically  relate to the spread or tail behavior. Examples include: the so-called \emph{Sharpe ratio}, which is the ratio between the mean and standard deviation; {\em  value-at-risk} ($\VAR$) which focuses on the $\alpha$ percentile of the distribution (with $\alpha$ small); or a close counterpart that integrates (averages) the values out in the tail beyond that point known as the {\em expected shortfall} (or conditional value at risk; $\CVAR$). The last  example is of further interest as it belongs to the class of \emph{coherent} risk measures which has various attractive properties from the risk theory perspective. A discussion thereof is beyond the scope of this paper; cf.~\cite{artzner1999coherent} for further details. In our problem setting, the above criteria are applied via the function $\Rbase$ to the empirical observations, and then the decision maker seeks, as before, to optimize its expected value. A typical example where such criteria may be of interest is that of clinical trials (one of the original motivations for the development of the MAB framework).  More specifically, suppose several new drugs are sequentially tested on individuals who share similar characteristics. If we consider average performance, we may conclude that the best choice is a drug with a non-negligible fatality rate but a high success rate. If we wish to control the fatality rate then using $\CVAR$ for example may be appropriate.

While some of the above mentioned criteria have been examined in the decision making and learning literature (see  references below), the analysis tends to be driven by specific properties of the criterion in question, and is very much done on a case-by-case basis. One of the purposes of this paper is to present a more unified approach to a large set of such problems. One of the main obstacles that arises under the non-cumulative criteria  is the more complicated structure of the oracle rule. In particular,  unlike the case of the mean objective, here the oracle rule need not select the same arm throughout the horizon of the problem. This presents further obstacles in identifying and characterizing a learning policy, as most such blueprints call for minimizing regret by mimicking the oracle rule. To that end, as our analysis will flesh out, under suitable conditions the oracle policy can be approximated (asymptotically) by a {\it \simple\space policy}, that is, one that statically selects a single arm. This simplification can be leveraged to address the  {\it learning problem} which becomes more tractable and amenable to optimism-based design principles.  
It is therefore of interest to understand and characterize in what instances  does  this simplified structure exist. This is one of the main thrusts of the paper.

\textbf{Main contributions of this paper.} 
In this paper we consider a general approach to the analysis of performance criteria where the oracle policy is a simple policy. We identify a class of criteria that we term \emph{Empirical Distribution Performance Measures} (\EDRMabbrv). In particular, let $\hat{F}$ be the \emph{empirical distribution} of the vector $(\Xt[1],\ldots,\Xt)$, i.e., $\hat{F}(y)$ is the fraction of rewards less or equal to real valued $y$. An \EDRMabbrv\space evaluates performance by means of a function $\RHat$, which maps $\hat{F}$ to $\RR$, i.e., $\RHat(\hat{F}) = \Rbase(\Xt[1],\ldots,\Xt)$. Alternatively, $\RHat$ may also serve to evaluate the distributions of the random variables $\Xt[s]$ ($s=1,\ldots,t$). These evaluations may be aggregated to form a different type of performance criteria that we term ``pseudo regret,'' and consider as an intermediate learning goal. Our main results provide easily verifiable explicit conditions that characterize the asymptotic behavior of the oracle rule, and culminate in a $UCB$-type learning algorithm with either $\mathcal{O}(\log T/T)$ or $\mathcal{O}(1/\sqrt{T})$ normalized regret (depending on the properties of $\RHat$ and the arm distributions). 

\rev
{
While our framework encompasses many existing performance measures, there are some notable examples that are left out. One is the Maximum Drawdown, which measures the maximum decline from the highest historical peak in the reward sequence. This criterion cannot be expressed by our framework as it depends on the order of the reward sequence. Other criteria such as certain distortion risk measures and cumulative prospect theory can be expressed within our framework, however, they do not satisfy some of its requirements, hence our results are not directly applicable. Whether these could be incorporated into our framework remains open to future work. See further details in \cref{sec:limitations}.
}
{}

\textbf{Previous works on bandits that concern path-dependent and risk criteria}.
Sequential performance measures of the type considered here were previously studied in \cite{sani2012risk}, which considered the Mean-Variance of the sequence and presented the MV-UCB, and MV-DSEE algorithms, and \cite{vakili2016risk,vakili2015mean}, which complete the regret analysis of said algorithms and also consider performance under Value at Risk. \cite{icml2020_1567} also consider the Mean-Variance and give a Thompson sampling-based method. Additionally, \cite{agrawal2019bandits} consider a constrained bandit problem with a concave objective. This may, for example, capture the Mean-Variance setting,  however, it is not the main focus  of the paper and the results are restricted to order $\mathcal{O}(\sqrt{T})$ regret (or order-$1/\sqrt{T}$ normalized regret in our setting).  

Other works consider simpler performance measures that are more closely related to our notion of pseudo regret.
\cite{galichet2013exploration} present the MaRaB algorithm which uses $\CVAR$ in its implementation, however, they analyze the average reward performance, and do so under the assumptions that $\alpha = 0$, and the assumption that the  $\CVAR$ and average optimal arms coincide.
\cite{tamkindistributionally} also consider $\CVAR$ and give a sample-wise optimistic algorithm.
\cite{bhat2019concentration} consider the concentration of risk measures and apply it exclusively to bound $\CVAR$ pseudo regret. It should be noted that some of their findings, and in particular the pseudo regret bound, where previously established in  \cite{cassel2018general}, which is an antecedent to the present  paper. 
\cite{zimin2014generalized} consider criteria based on the mean and variance of distributions, and present and analyze the $\varphi-LCB$ algorithm. We note that these criteria correspond to a much narrower class of problems than the ones considered here.
\cite{maillard2013robust} presents and analyzes the RA-UCB algorithm which considers the measure of \textit{entropic risk} with a parameter $\lambda$. 

Slightly farther afield, other works consider the best arm identification or simple regret settings.
\cite{kagrecha2019distribution} propose distribution independent algorithms for a linear combination of the mean and $\CVAR$ measures.
\cite{icml2020_2156} consider the estimation of $\CVAR$ for both light and heavy tailed distributions, and provide an algorithm for best $\CVAR$ arm identification.
\cite{tran2014functional} consider a general functional of the arm distributions and demonstrate results on Mean-Variance, $\VAR$, and $\CVAR$ best arm and simple regret.
\cite{yu2017risk} consider a Mean-Variance best arm identification.
\cite{david2018pac} consider a quantile risk constrained setting for best arm identification.

Finally, we mention two alternative settings that consider quantile-based sequential performance measures such as $\VAR$ and $\CVAR$.
\cite{torossian19a} consider the convergence of approximate dynamic programming in Markov Decision Processes (MDP), and
\cite{jiang2018risk} consider a stochastic optimization setting with bandit feedback. The approach in our paper is quite distinct from these.

\textbf{Organization}.
Throughout,  all proofs are provided as a sketch that communicates their key ideas, with the full details deferred to the Appendix. In \cref{sec:EDPMMotivation} we give initial motivation for our suggested class of performance criteria. In \cref{sec:formulation} we formulate the problem setting, oracle rule, and regret metric under non-cumulative criteria. In \cref{sec:main} we provide the main results, and in \cref{sec:examples} we demonstrate them on well-known risk criteria. We also include some negative examples, which illustrate the implications of violation of the proposed conditions, indicating in some way the necessity of such conditions in achieving the unifying theme in our proposed framework. 

\section{A Motivating Example}
\label{sec:EDPMMotivation}

In the standard bandit setting, for a sequence of  integrable random variables we are interested in designing a policy $\policy$ that maximizes the average reward
$
    \RPiNamed{\text{ave}}
    =
    \EEBrk[1]{\sum_{t=1}^{T} \XtPi}
    ,
$
or, equivalently, minimizes regret compared to an oracle strategy, which is known to pull a single arm throughout the horizon.
It is well known that optimistic strategies achieve optimal performance in this setting.
Now, suppose we seek to design a policy $\policy$ that maximizes
$
    \RPiNamed{\CVAR}
    =
    \EEBrk[1]{
        \frac{1}{\ceil{t \alpha}} \sum_{s=1}^{\ceil{t \alpha}} \XtPiOrd{s}
    }
    ,
$
where $\XtPiOrd{s}$ is the $s^{th}$ order statistic of $(\XtPi[1], \ldots, \XtPi)$. This is known as Conditional Value at Risk ($\CVAR$) or Expected Shortfall, at percentile level $\alpha \in (0,1)$, and is a widely accepted performance measure from the risk literature.
\begin{align*}
    \text{
    Question: 
    Can we minimize regret using an optimistic strategy?
    }
\end{align*}
To answer this, we first need to understand what constitutes an optimistic strategy, which in turn requires that we further understand the oracle rule, which is aware of the true distributions of the arms.
The current formulation of $\RPiNamed{\CVAR}$ presents it as a direct function of the reward sequence. While this is very intuitive, it is in fact a sequence of mappings (one for each sample size) that do not naturally share a domain;  studying this sequence can be challenging. In lieu of that, we first observe that  $\RPiNamed{\CVAR}$ may be reformulated in terms of a single function that evaluates the sequence of empirical reward distributions $\FHatPi{t}$, where $\FHatPi{t}(x)$ is the fraction of rewards less than or equal to $x$. Formally,
$
    \RPiNamed{\CVAR}
    =
    \EE{\RHatFunc[\CVAR]{\FHatPi{t}}}
$
where
\begin{align*}
   \RHatFunc[\CVAR]{F}
    =
        \max_{z \in \RR} 
            z
            -
            \frac{1}{\alpha} \int_{-\infty}^{z} F(x) dx
	,
\end{align*}
is the accepted notion for measuring $\CVAR$ for a random variable $X$ with cumulative distribution function $F$.
In \cref{appendix:EDRMmotivation} we show that essentially any performance measure that is invariant to the order of the reward sequence may be reformulated as 
$
    \EE{\RHatFunc{\FHatPi{t}}}
$
for some function $\RHat$. Importantly, this captures many well-known performance measures such as Mean-Variance, Entropic Risk, Sharpe Ratio, and more.

Having restricted ourselves to the class of performance measures that can be expressed this way, we have obtained a consistent way to evaluate performance, which is independent of the time $t$. Our study now turns to investigating the properties of the function $\RHat$, in conjunction with the arm distributions, that allow for an optimistic strategy.
In the case of $\CVAR$, if we only assume sub-Gaussian arm distributions, we find that $\RHat[\CVAR]$ may be non-smooth, and our framework can only guarantee a regret of $\mathcal{O}(1/ \sqrt{T})$ using an optimistic strategy. However, if for example the arm distributions have positive density around their $\alpha$ percentile then we show that the same strategy only incurs
$
    \mathcal{O}(\log T / T)
$
regret. The remainder of this paper will flesh out these ideas and provide a set of easy to verify conditions that yield the results discussed in this section.

\section{Problem Formulation}
\label{sec:formulation}
\paragraph{Model and admissible policies.} Consider a standard MAB with 
$
    \actionSet
    =
    \brk[c]*{1, \ldots, K}
    ,
$
the set of arms. Arm $i \in \actionSet$ is associated with a sequence $\Xti{t}$ ($t \ge 1$) of $i.i.d$ random variables with distribution $\Fi \in \DistSet$, the set of all distributions on the real line. When pulling arm $i$ for the $t^{th}$ time, the decision maker receives reward $\Xti{t}$, which is independent of the remaining arms, i.e., the variables $\Xti{t}$ (for all $i \in \actionSet, t \ge 1$) are mutually independent.

We define the set of \textit{admissible} policies (strategies) of the decision maker in the following way. Let $\tauIAt{t}$ be the number of times arm $i$ was pulled up to time $t$. Let $V$ be a random variable over a probability space $\brk*{\mathbb{V}, \mathcal{V}, P_v}$ which is independent of the rewards. An \textit{admissible} policy $\policy = \brk*{\policyAt{1}, \policyAt{2}, \ldots}$ is a random process recursively defined by
\begin{align}
    &\policyAt{t} 
    :=
    \policyAt{t} \brk*{
        V, \policyAt{1}, \ldots, \policyAt{t-1}, \XtPi[1], \ldots, \XtPi[t-1]
    }
    \\
    &\tauIAt{t}
    =
    \sum_{s=1}^{t} \indEvent{\policyAt{s} = i} \label{eq:tauIDef}
    \\
    &\XtPi
    :=
    \Xti[i]{\tauIAt{t}} \text{, given the event } \brk[c]*{\policyAt{t} = i}
    .
\end{align}
We denote the set of \textit{admissible} policies by $\policiesSet$, and note that \textit{admissible} policies $\policy$ are non anticipating, i.e.,  depend only on the past history of actions and observations, and allow for randomized strategies via their dependence on $V$. Formally, let $\seqDef{\FiltrationAt{t}}{t=0}{\infty}$ be the filtration defined by $\FiltrationAt{t} = \sigma \brk*{V, \policyAt{1},  \XtPi[1], \ldots, \policyAt{t}, \XtPi[t]}$, then  $\policyAt{t}$ is $\FiltrationAt{t-1}$ measurable.

\paragraph{Empirical Distribution Performance Measures (\EDRMabbrv).} 
The classical bandit optimization criterion centers on the \textit{empirical mean}, i.e., $\frac{1}{t}\sum_{s=1}^{t}\XtPi[s]$. We generalize this by considering criteria that are based on the \textit{empirical distribution}. Formally, the \textit{empirical distribution} of a real number sequence $x_1, \ldots, x_t$ is obtained through the mapping ${\FHatt : \RR[t] \rightarrow \DistSet}$, given by
\begin{align} \label{eq:empDistMap}
    \FHattFunc{x_1,\ldots,x_t ; \cdot}
    =
    \frac{1}{t}\sum_{s=1}^{t}\indFunc{x_s} (\cdot),
\end{align}
where $\indFunc[b]{a}(\cdot)$ is the indicator function of the interval $\brk[s]*{a,b}$ defined on the extended real line, i.e.
\begin{align*}
    \indFuncAt[b]{a}{y}
    =
    \begin{cases}
        1 \qquad, y \in [a,b] \\
        0 \qquad, y \notin [a,b].
    \end{cases}
\end{align*}
Of particular interest to this work are the empirical distributions of the reward sequence under policy $\policy$, and of arm $i$. We denote these respectively by,
\begin{align}
    &\FHatPi{t}(\cdot) 
    ~:=
    \FHattFunc{\XtPi[1], \ldots, \XtPi[t]; \cdot}  \label{eq:FhatPiAbbrv}
    \\
    &\FHatPi[(i)]{t}(\cdot) 
    :=
    \FHattFunc{\Xti{1}, \ldots, \Xti{t}; \cdot}.  \label{eq:FhatiAbbrv}
\end{align}
The decision maker possesses a function $\RHat : \DistSet \to \RR$, which measures the ``quality'' of a distribution. The resulting criterion is called \EDRMabbrv, and the decision maker aims to maximize $\EE{\RHatFunc{\FHatPi{T}}}$. (Throughout it is assumed implicitly that the class of distributions and performance functions is such that this expectation exists and is finite valued.)

\paragraph{Oracle and regret.} For given horizon $T$, the oracle policy $\optPolicyT = (\optPolicyTAt{1}, \optPolicyTAt{2}, \ldots)$ is one that achieves optimal performance given full knowledge of the arm distributions $\Fi$ ($i \in \actionSet$). Formally, it satisfies
\begin{align} \label{eq:oracleDef}
    \optPolicyT
    \in
    \argmax_{\policy \in \policiesSet} \EEBrk{\RHatFunc{\FHatPi{T}}}.
\end{align}
Similar to the classic bandit setting, we define a notion of regret that compares the performance of policy $\policy$ to that of $\optPolicyT$. The expected (normalized) regret of policy $\policy \in \policiesSet$ at time $T$ is given by,
\begin{align} \label{eq:regretDef}
    \regret
    := 
    \EEBrk{\RHatFunc{\FHatPi[\optPolicyT]{T}} - \RHatFunc{\FHatPi{T}}}.
\end{align}
We note that this definition is normalized with respect to the horizon $T$, thus transforming familiar regret bounds such as $\mathcal{O}(\log T)$ into $\mathcal{O}(\frac{\log T}{T})$. With that convention we simply refer to the above as the ``regret'' without added qualifiers.  

\paragraph{Assumptions.}
 Beyond the existence and finiteness of expectations, flagged earlier, we make the following assumption throughout.
Let the simplex in $\RR[K]$ be
\begin{align*} \textstyle
    \simplex
    =
    \setDef[\Big]{
        p = \brk*{p_1, \ldots, p_K} \in \RR[K] 
        \given
        \sum_{i=1}^{K}p_i = 1,\;
        p_i \ge 0 \;\forall i \in \actionSet
    },
\end{align*}
and define the set of all convex combinations of the arms' reward distributions by
\begin{align} \textstyle
\label{eq:distSetDeltaDef}
	\DistSetDelta 
	=
	\setDef[\Big]{
	    \Fp = \sum_{i=1}^{K}{p_{i} \Fi} 
	    \given 
	    p \in \simplex
    }
	.
\end{align}
Let $i^* \in \argmax \RHatFunc{\Fi}$ be an ``optimal" arm. We assume that
\begin{align}
    \label{eq:singleOpt}
    \RHatFunc{F} \le \RHatFunc{\FiOpt}
    \quad
    ,
    \forall F \in \DistSetDelta.
\end{align}
\rev
{
While this assumption holds for many known performance measures, and seems crucial for obtaining the improved $\log T / T$ regret rate, we note that there is some loss of generality here. In \cref{sec:limitations}, we give a few examples of known performance measures where \cref{eq:singleOpt} does not hold, and discuss how one could potentially address them within our framework. Finally, we note that \cref{eq:singleOpt} alone does not immediately imply that playing $i^*$ is optimal or even near-optimal over a finite horizon $T$ (for an illustration of this, see  \cref{appendix:subsection:counterExamples}).
}
{
While there is a potential loss of generality here, we could not find any interesting performance measure that violates this inequality. In particular, it provably holds when $\RHat$ is quasiconvex, which will always be the case in our examples.
}

\section{Main Results}
\label{sec:main}

When defining an objective, it was sufficient to consider $\RHat$ as a mapping from $\DistSet$ (a \textit{set}) to $\RR$. Moving forward, our analysis relies on properties such as continuity and differentiability, which requires that we consider $\RHat$ as a mapping between seminormed spaces. To that end $\DistSet$ is a subset of an infinite dimensional vector space for which norm equivalence does not hold. This hints at the importance of using the ``correct'' (semi)norm for each $\RHat$. As a result, our analysis is done with respect to a general seminorm $\norm{\cdot}$ and its matching seminormed space $\funcSpace$.
We therefore consider \EDRMabbrv s as mappings $\RHat : \funcSpace \to \RR$.

The goal of this work is to provide a generic analysis of the regret, similar to that of the classical bandit setting, and which culminates in the following result. \vskip 1em 

\begin{theorem*}[\textbf{Informal meta-result}]
There exists an efficient algorithm such that:
\begin{enumerate}
    \item Under suitable regularity conditions obtains regret
    $
        \regret 
        =
        \mathcal{O}(\frac{\log T}{\sqrt{T}})
        ;
    $
    \item Under additional smoothness and gap conditions obtains regret
    $
    \regret
    =
    \mathcal{O}(\frac{\log T}{{T}})
    .
    $
\end{enumerate}
\end{theorem*}
\vskip 1em 

\noi
In what follows, we introduce the technical details required to make this statement rigorous. This culminates in \cref{sec:regretBound}, where \cref{thm:regretBound} gives the desired statement, and where we also explain how the standard bandit setting fits into our framework.
Unlike the classical  bandit setting, the oracle policy $\optPolicyT$, defined in \cref{eq:oracleDef}, need not choose a single arm. Since the typical learning algorithms are structured to emulate the oracle rule, we need to first understand the structure of the oracle policy before we can analyze $\regret$.

\subsection{Insights From the Infinite Horizon Oracle}
\label{sec:infOracle}

The oracle problem in \cref{eq:oracleDef} does not admit a tractable solution, in the absence of further structural assumptions. In this section we consider a \textit{relaxation} of the oracle problem which examines asymptotic behavior. We provide conditions under which this behavior is ``simple'' thus suggesting it as a proxy for the finite time performance. More concretely, let 
$
\RPi
=
\liminf_{t \to \infty} \RHatFunc{\FHatPi{t}}
$
be the \textit{worst case} asymptotic performance of policy $\policy$, then the infinite horizon oracle $\optPolicy = (\optPolicyAt{1}, \optPolicyAt{2}, \ldots)$ satisfies
\begin{align} \label{eq:infOracleDef}
	\optPolicy \in \argmax_{\policy \in \policiesSet} \EEBrk{\RPi}.
\end{align}
Note that $\RPi$ is well defined as the limit inferior of a sequence of random variables, however (as indicated earlier) we require that its expectation exist for \cref{eq:infOracleDef} to be well defined. 

\paragraph{\Simple\space oracle.} In the traditional multi-armed bandit problem, the oracle policy, which selects a single arm throughout the horizon, is clearly \simple.
It may seem intuitive that \EDRMabbrv s always admit such a \simple\space infinite horizon oracle policy. However, in (\ref{appendix:subsection:counterExamples}) we give counter examples, which arise from the ``bad behavior" that is still allowed by this objective. The following result gives sufficient conditions for \EDRMabbrv s to be ``well behaved."
\vskip 1em 

\begin{theorem}[\textbf{\EDRMabbrv\space admits a \simple\space oracle policy}]
\label{theorem:StableEDRMoptPolicy}
    Suppose an \EDRMabbrv, $\RHat : \funcSpace \to \RR$, is continuous on $\DistSetDelta$, and that 
    $
    \lim_{t \to \infty} \norm{\FHatPi[(i)]{t} - \Fi}
    = 
    0
    $
    almost surely $\mbox{for all }  i \in \actionSet$.
    Then the single arm policy that always chooses $i^*$ is an infinite horizon oracle policy, as defined in \cref{eq:infOracleDef}.
\end{theorem}

\noi\textbf{Proof sketch.}
(see full details in \cref{appendix:infHorizonOracle})
First, define the arm pulling ratio
$
    \piAt{t} 
    =
    {\tauIAt{t}}/{t}
    ,
$
and notice that the empirical distribution may be written as
$
    \FHatPi{t}
    =
    \sum_{i=1}^{K} \piAt{t} \FHatPi[(i)]{t}
    .
$
Since $p(t) \in \simplex$, which is closed and compact, we have that any subsequence of $t$ has a further subsquence, $t_l$, such that $p(t_l) \to p \in \simplex$.
Next, since we assumed that
$
\lim_{t \to \infty} \norm{\FHatPi[(i)]{t} - \Fi}
= 
0
$
almost surely, we conclude that
$
\FHatPi{t_l}
\to
\Fp
.
$
Applying the continuity assumption we conclude that
$
\RPi
=
\liminf_{t \to \infty} \RHatFunc{\FHatPi{t}}
\le
\lim_{l \to \infty} \RHatFunc{\FHatPi{t_l}}
=
\RHatFunc{\Fp}
\le
\RHatFunc{\FiOpt}
.
$
We conclude the proof by showing that similar arguments imply that the proposed \simple\space oracle policy achieves this upper bound.

\begin{remark*}
    \cref{theorem:StableEDRMoptPolicy} depends not only on $\RHat$ but also on the given distributions $\Fi$. Meaning, it may hold for a given $\RHat$ only for some distributions, and thus the choice of a seminorm is important in order to get sharp conditions on the viable reward distributions. For example, consider the supremum norm given by $\norm{f}_\infty = \sup_{x \in \RR}\abs{f(x)}$. By the Glivenko-Cantelli theorem \cite{van2000asymptotic}, it satisfies the convergence condition for any given distributions $\Fi$, $i \in \actionSet$. However, in most cases, continuity holds only if the distributions have bounded support.
\end{remark*}

\subsection{Regret Decomposition}

Having gained some understanding of the infinite horizon oracle, we consider a regret decomposition that uses the infinite horizon performance as a benchmark. 
Let
\begin{align}
\label{eq:FProxDef}
    \FProxPi{T}
    =
    \frac{1}{T}\sum_{t=1}^{T} \Fi[\policyAt{t}] 
    =
    \frac{1}{T}\sum_{i=1}^{K} \tauI \Fi,
\end{align}
be the pseudo empirical distribution, where we recall that $\Fi$ is the distribution associated with arm $i \in \actionSet$, and $i^*$ is such that
$
    \RHatFunc{F}
    \le
    \RHatFunc{\FiOpt}
    ,
$
for all $F \in \DistSetDelta$.
The regret may now be decomposed as
\begin{align} \label{eq:regretDecomp}
\begin{aligned}
    \regret
    &=
    \underbrace{
    \EEBrk{
        \RHatFunc{\FHatPi[\optPolicyT]{T}}
        -
        \RHatFunc{\FProxPi[\optPolicyT]{T}}
    }
    }_{J_1(T)}
    \\
    &+
    \underbrace{
    \EEBrk{
        \RHatFunc{\FProxPi[\optPolicyT]{T}}
        -
        \RHatFunc{\FiOpt}
    }
    }_{J_2(T)}
    \\
    &+
    \underbrace{
    \EEBrk{
        \RHatFunc{\FiOpt}
        -
        \RHatFunc{\FProxPi{T}}
    }
    }_{\pseudoRegret}
    \\
    &+
    \underbrace{
    \EEBrk{
        \RHatFunc{\FProxPi{T}}
        -
        \RHatFunc{\FHatPi{T}}
    }
    }_{J_3(T)}
    .
\end{aligned}
\end{align}
The term $\pseudoRegret$ represents what we believe to be the correct notion of pseudo regret in our setting.
Unlike the standard bandit pseudo regret
$
\frac{1}{T} \EEBrk[1]{
    \sum_{t=1}^{T} 
    \RHatFunc{\FiOpt}
    -
    \RHatFunc{\Fi[\policyAt{t}]}
}
,
$
which aggregates a policy's decisions in reward space, here aggregation occurs in distribution space and then evaluates to a reward via $\RHat$. We note that in the standard average reward setting $\RHat$ is linear and both notions coincide. However, in the general non-linear setting, the previous notion may underestimate the regret and is thus unsatisfactory.

The remaining terms in \cref{eq:regretDecomp} may be viewed as a decomposition into error terms that measure discrepancy between distributions via the criterion $\RHat$. As hinted at by our notation, these may be bounded essentially independently of the learning algorithm's policy $\policy$.
The term $J_2(T)$ measures the difference between the optimal infinite horizon arm pulling mixture, which is a single arm, and that of the finite horizon oracle. Notice that by definition of $i^*$ we always have that $J_2(T) \le 0$.
The terms $J_1(T)$ and $J_3(T)$, which we refer to as horizon gaps, measure the convergence of empirical distributions to their appropriate infinite horizon counterparts, which are given by pseudo empirical distributions. While bounding these proves to be the crux of our problem, we begin by proposing a learning algorithm that minimizes the pseudo regret.

\subsection{Learning Algorithm}
\label{sec:learningAlgorithm}
\cref{theorem:StableEDRMoptPolicy} presented conditions for understanding the asymptotic behavior of performance. As we now seek a finite time analysis (of the pseudo regret), it stands to reason to employ the following stronger conditions, which quantify the rate of convergence. \vskip 1em 
\begin{definition}[\textbf{Stable \EDRMabbrv}] \label{definition:StrongEDRM}
	We say that $\RHat$ is stable with respect to a seminorm $\norm{\cdot}$ if there exist
	$
	    \concentrationConst, \polyContModCoeff > 0
	    ,
	    \polyContModDeg \ge 1
    $
    such that:
	\begin{enumerate}
		\item \label{item:StrongEDRMCond1}
		$\RHat$ admits $\contModFunc{x} = \polyContModCoeff\brk*{x + x^\polyContModDeg}$ as a local modulus of continuity for all $F \in \DistSetDelta$, i.e.,
		\begin{align*}
    		\abs{\RHatFunc{F} - \RHatFunc{G}} 
    		\le
    		\contModFunc{\norm{F-G}},
    		\qquad \forall F \in \DistSetDelta, G \in \funcSpace
    		.
		\end{align*}
		\item \label{item:StrongEDRMCond2}
		Recalling $\FHatPi[(i)]{t}$ from \cref{eq:FhatiAbbrv}, we have that for all $i \in \actionSet$, $t \ge 1$
		\begin{align*}
		\PP{
		    \norm{\FHatPi[(i)]{t} - \Fi} \ge x
	    } 
	    \le
	    2 \exp \brk*{- \concentrationConst t x^2},
	    \qquad \forall x>0.
		\end{align*}
	\end{enumerate}
\end{definition}
\rev{We note that the constant $\concentrationConst$ is treated as a parameter in what follows recognizing   that its value depends on the specific choice of norm above, as well as  the assumptions made on the arm distributions.}{}

\paragraph{Pseudo regret decomposition.}
In the traditional bandit setting, which considers the average reward, the analysis of the regret is well understood. The same analysis extends to any linear \EDRMabbrv, i.e., when $\RHat$ is linear. This follows straightforwardly as such rewards can be formulated as the usual average criterion with augmented arm distributions. Linearity facilitates the regret analysis by providing a decomposition of contributions from each sub-optimal arm. Define the standard single arm sub-optimality gap
\begin{align*}
    \Di
    =
    \RHatFunc{\FiOpt} - \RHatFunc{\Fi}
    ,
\end{align*}
where we recall that $i^* \in \argmax \RHatFunc{\Fi}$ is the optimal arm. The regret of a linear \EDRMabbrv\space is given by, 
$
    \regret
    =
    \frac{1}{T}\sum_{i \neq i^*} \Di \EE{\tauI}
    .
$
Departing from the simple realm of linearity, we seek a similar decomposition of the pseudo regret. To that end, denote the diameter of $\DistSetDelta$ and the maximum gap ratio respectively as
\begin{align}
\label{eq:diameterAndGapRatio}
    \distSetDiameter
    =
    \max_{i,j \in \actionSet}\norm{\Fi - \Fi[j]}
    \qquad
    \gapRatio
    =
    \max_{i \neq {i^*}} \norm{\FiOpt - \Fi} / \Di,
\end{align}
where the latter essentially measures how well the chosen seminorm captures the sub-optimality gaps.
We provide the following result, which is proved in \cref{appendix:proxyRegret}. \vskip 1em 
\begin{lemma}[\textbf{Pseudo regret decomposition}] \label{lemma:pseudoRegretDecomp}
	Let $\RHat$ be a stable \EDRMabbrv, then $\RHat$ is 
	$\lipConst$-Lipschitz over $\DistSetDelta$ with 
	$
	    \lipConst
	    =
	    \polyContModCoeff \brk*{1 + \distSetDiameter^{\polyContModDeg - 1}}
	    ,
    $
    and we have that
	\begin{align*}
		\pseudoRegret \le \frac{\lipConst \gapRatio}{T} \sum_{i \neq i^*} \Di \EE{\tauI}.
	\end{align*}
\end{lemma}

\paragraph{Learning algorithm.} 
We present $\UUCB$, a natural adaptation of $\brk*{\alpha,\psi}-UCB$ (see \cite{bubeck2012regret}) to a stable \EDRMabbrv. Let,
\begin{align*}
    &\phiFuncAt{y}
    = 
    \min \brk[c]*{
        \concentrationConst \brk*{\frac{y}{2 \polyContModCoeff}}^2,
        \concentrationConst \brk*{\frac{y}{2 \polyContModCoeff}}^{2 / \polyContModDeg}
    } 
    \\
    &\phiInvFuncAt{x}
    = 
    \max \brk[c]*{
    2\polyContModCoeff \brk*{\frac{x}{\concentrationConst}}^{1/2},
    2\polyContModCoeff \brk*{\frac{x}{\concentrationConst}}^{\polyContModDeg / 2}
    }
    ,
\end{align*}
where $\concentrationConst, \polyContModCoeff, \polyContModDeg$ are the parameters of \cref{definition:StrongEDRM}. The $\UUCB$ policy is given by,
\begin{align}
    \policyAt[\UUCB]{t}
    \in
    \argmax_{i \in \actionSet}\brk[s]*{
        \RHatFunc{\FHatPi[(i)]{\tauIAt{t-1}}} 
        +
        \phiInvFuncAt{\frac{\gamma \log t}{\tauIAt{t-1}}}
    },
    \quad t \ge K+1,
\end{align}
where for $1 \le t \le K$, it samples each arm once as initialization. \vskip 1em 

\begin{theorem}[$\mathbf{\UUCB}$ \textbf{Pseudo Regret}]
\label{theorem:UUCB}
	Let $\RHat$ be a stable \EDRMabbrv. If $\Di > 0$ for all $i \neq i^*$, then for $\lipConst$ defined in \cref{lemma:pseudoRegretDecomp} and $\gamma > 2$ we have that
	\begin{align*}
    	\pseudoRegret[{\UUCB}]
    	\le
    	\frac{\lipConst \gapRatio}{T} 
    	\sum_{i \neq i^*} \brk*{
    	    \frac{\gamma \Di \log T}{\phiFuncAt{\Di / 2}}
    	    +
    	    \frac{\gamma + 6}{\gamma - 2} \Di
	    }
	    .
	\end{align*}
\end{theorem}
\paragraph{Proof sketch.}
(see full details in \cref{appendix:proxyRegret})
The proof uses standard techniques from the UCB literature and consists of the following steps. 
First, we show that if at round $t$, the algorithm chooses sub-optimal arm $i$ that was pulled more than order-$\log T$ times, then we have significantly overestimated this arm and underestimated the optimal arm. 
Second, we bound the probability of this estimation failure using standard concentration arguments that are deduced from stability (\cref{definition:StrongEDRM}). 
We conclude that after choosing sub-optimal arm $i$ for order-$\log T$ times, the probability of choosing it again is very small, and thus the expected number of sub-optimal arm pulls is order-$\log T$. 
Finally, the proof is concluded by plugging this result into the pseudo regret decomposition, given in \cref{lemma:pseudoRegretDecomp}.

\subsection{Bounding the Horizon Gaps}
\label{sec:horizonGap}

Recall that the horizon gap of a policy $\policy \in \policiesSet$ is given by
$
    \abs1{
        \EEBrk[1]{
            \RHatFunc{\FHatPi{T}}
            -
            \RHatFunc{\FProxPi{T}}
        }
    }
$.
At their core, our bounds on the horizon gaps follow from the convergence of the empirical to pseudo empirical distribution. This convergence is quantified by the following lemma.
\begin{lemma}[\textbf{Empirical Distribution Tail Bound}] 
\label{lemma:empDistConcentration}
	Suppose that Requirement~\ref{item:StrongEDRMCond2} of stability holds, then
	\begin{align*}
		\PP{\norm{\FHatPi{T} - \FProxPi{T}} > x}
		\le
		2KT \exp \brk*{-\concentrationConst \frac{Tx^2}{K^2}} 
		\qquad,
		\forall T \ge 1 , x \ge 0
		.
	\end{align*}
\end{lemma}
The proof decomposes the deviation into its contributions from each arm and uses Requirement~\ref{item:StrongEDRMCond2} of stability, single arm concentration of the empirical distribution, and several union bounds to conclude the desired result. See full details in \cref{sec:horizonGapSideLemmas}.
We now present our first bound on the horizon gap, which is uniform over all policies $\policy \in \policiesSet$. \vskip 1em 
\begin{proposition}[\textbf{Uniform horizon gap bound}]
\label{prop:stableHorizonGap}
	Suppose that $\RHat$ is a stable \EDRMabbrv. Then we have that for all
	$
	    T
	    \ge
	    \frac{4 \polyContModDeg K^2 \log KT}{\concentrationConst}
    $
	\begin{align*}
	    \abs*{
	        \EEBrk{
	            \RHatFunc{\FHatPi{T}}
	            -
	            \RHatFunc{\FProxPi{T}}
	        }
	    }
	    \le
	    4 \polyContModCoeff \brk*{
        \frac{K^2 \log KT}{\concentrationConst T}
        }^{1/2}
        \qquad
        ,
        \forall \policy \in \policiesSet
        .
	\end{align*}
\end{proposition}
\noi The proof uses Requirement~\ref{item:StrongEDRMCond1} of stability, local modulus of continuity, to get that
\begin{align} \label{eq:horizonGapProblem}
\begin{aligned}
	\abs*{
	    \EEBrk{\RHatFunc{\FHatPi{T}} 
	    -
	    \RHatFunc{\FProxPi{T}}}
    }
	\le
	\EEBrk{
	    \contModFunc{\norm{\FHatPi{T} - \FProxPi{T}}}
    }
    =
    \polyContModCoeff
    \EEBrk{
        \norm{\FHatPi{T} - \FProxPi{T}}
        +
        \norm{\FHatPi{T} - \FProxPi{T}}^{\polyContModDeg}
    }
    ,
\end{aligned}
\end{align}
and then uses the tail sum formula together with the tail bound in \cref{lemma:empDistConcentration} to obtain the final bound.
See full details in \cref{appendix:horizonGapProofs}.

The result of \cref{prop:stableHorizonGap} exhibits a dependence on the time horizon $T$ which may be quite loose. To see this, consider a linear $\RHat$. It is relatively easy verify that the left hand side of \cref{eq:horizonGapProblem} is zero, while its right hand side behaves as $1/\sqrt{T}$ even when $K=1$.

In order to obtain improved bounds, we require a notion of smoothness. Formally, let $\linFunctionals$ be the space of bounded linear functionals on $\funcSpace$. Assuming $\RHat$ is differentiable on $F \in \DistSetDelta$, then its differential $\DRHatAt{F} \in \linFunctionals$ is well defined, and we denote its (linear) operation on $G \in \funcSpace$ by $\DRHatAtOn{F}{G}$.
\begin{definition}[\textbf{Smooth \EDRMabbrv}] \label{definition:smoothEDRM}
	An \EDRMabbrv, 
	$
	    \RHat : \funcSpace \to \RR
	    ,
    $
	is smooth if it is differentiable on $\DistSetDelta$, and there exist $\pSmooth \ge 0, M_0 > 0$ such that for any $F \in \DistSetDelta, G \in \funcSpace$ satisfying $\norm{G - F} \le M_0$ we have that
	\begin{align*}
	    \abs{
	        \RHatFunc{G}
	        -
	        \RHatFunc{F}
	        -
	        \DRHatAtOn{F}{(G - F)}
	    }
	    \le
	    \frac12 \pSmooth \norm{G - F}^2
	\end{align*}
\end{definition}
\noi This definition is a standard notion from optimization, stated for our infinite dimensional function space. 
The following result shows that ``reasonable'' policies enjoy a smaller horizon gap under the smoothness assumption.  \vskip 1em 

\begin{theorem}[\textbf{Improved horizon gap bound}]
\label{thm:smoothHorizonGap}
	Suppose that $\RHat$ is a stable and smooth \EDRMabbrv\space with $M_0 = \infty$.
    Letting
    $
        \MT^2
        =
        \frac{4 K^2 \log KT}{\concentrationConst T}
        ,
    $
    we have that for any policy $\policy \in \policiesSet$ and fixed $F \in \DistSetDelta$
	\begin{align*}
    	\abs*{
	        \EEBrk{
	            \RHatFunc{\FHatPi{T}}
	            -
	            \RHatFunc{\FProxPi{T}}
	        }
	    }
    	\le 
    	\pSmooth \MT^2
    	+
    	\pSmooth \MT \EE{\norm{\FProxPi{T} - F}}
    	,
	\end{align*}
	for all $T \ge T_0$, which is polynomial in problem parameters.
\end{theorem}
\vskip 1em 

\noi The proof may be found in \cref{appendix:horizonGapProofs}. It explicitly  identifies the parameter $T_0$, and also addresses the case of $M_0 < \infty$, which gives an additional low order term.
Notice that $\MT^2 = \mathcal{O}(\log T / T)$ and thus any policy whose arm pull frequencies converge in expectation at a rate of $\mathcal{O}(\sqrt{\log T / T})$ has horizon gap of $\mathcal{O}(\log T / T)$. In particular, this clearly holds for $\UUCB$. \vskip 1em 

\begin{proof}[(sketch)]
First, a simple calculation shows that
$
    \EE{\FHatPi{t}}
    =
    \EE{\FProxPi{t}}
    .
$
Next, since $\DRHat$ is a linear operator, we conclude that
$
    \EEBrk[1]{\DRHatAtOn
        {F}
        {(\FHatPi{T} - \FProxPi{T})}
    }
    =
    0
$
for all $F \in \DistSetDelta$.
We thus obtain the following decomposition
\begin{align*}
    \abs*{\EEBrk{
        \RHatFunc{\FHatPi{T}} 
	    -
	    \RHatFunc{\FProxPi{T}}
    }}
    &\le
    \EE{\abs2{
    \underbrace{
        \RHatFunc{\FHatPi{T}} 
	    -
	    \RHatFunc{{\FProxPi{T}}}
	    -
	    \DRHatAtOn
        {\FProxPi{T}}
        {(\FHatPi{T} - \FProxPi{T})}
    }_{\delta_1}
    }}
    \\
    &+
    \EE{\abs2{
    \underbrace{
        \brk1{
            \DRHatAt{\FProxPi{T}}
            -
            \DRHatAt{F}
        }
        \cdot
        \brk{\FHatPi{t} - \FProxPi{T}}
    }_{\delta_2}
    }}
    ,
\end{align*}
To bound $\EE{\abs{\delta_1}}$ we use smoothness to get that
\begin{align*}
    \EE{
        \abs{\delta_1} 
    }
    \le
    \frac12 \pSmooth \EEBrk{\norm{\FHatPi{T} - \FProxPi{T}}^2}
    ,
\end{align*}
and using the tail sum formula together with the tail bound in \cref{lemma:empDistConcentration} bounds $\EE{\abs{\delta_1}}$.

Finally, to bound $\EE{\abs{\delta_2}}$ we first use the Cauchy–Schwarz inequality together with smoothness to get that
\begin{align*}
    \EE{\abs{\delta_2}}
    \le
    \EEBrk{
        \norm{
            \DRHatAt{\FProxPi{T}}
            -
            \DRHatAt{F}
        }
        \norm{\FHatPi{T} - \FProxPi{T}}
    }
    \le
    \pSmooth
    \EEBrk{
        \norm{\FProxPi{T} - F}
        \norm{\FHatPi{T} - \FProxPi{T}}
    }
    .
\end{align*}
Now, for small deviations we have that
\begin{align*}
    \EEBrk{
        \abs{\delta_2}
        \indEvent[1]{\norm{\FHatPi{T} - \FProxPi{T}} \le \MT}
    }
    \le
    \pSmooth \MT \EE{\norm{\FProxPi{T} - F}}
    .
\end{align*}
On the other hand, recalling that $\distSetDiameter$ is the diameter of $\DistSetDelta$, we get that
\begin{align*}
    \EEBrk{
        \abs{\delta_2}
        \indEvent[1]{\norm{\FHatPi{T} - \FProxPi{T}} > \MT}
    }
    \le
    \pSmooth \distSetDiameter
    \EEBrk{
        \norm{\FHatPi{T} - \FProxPi{T}}
        \indEvent[1]{\norm{\FHatPi{T} - \FProxPi{T}} > \MT}
    }
    ,
\end{align*}
and using the tail sum formula together with \cref{lemma:empDistConcentration}, and summing the two inequalities bounds $\EE{\abs{\delta_2}}$, and concludes the proof.
\end{proof}

\subsection{Regret Bound}
\label{sec:regretBound}

In order to conclude our regret bounds, we require the following definition. \vskip 1em 
\begin{definition}[\textbf{Linear gap}]
We say $\RHat$ has a linear gap if there exists $\pLinGap > 0$ such that
\begin{align*}
    \RHatFunc{F}
    \le
    \RHatFunc{\FiOpt}
    -
    \frac{1}{\pLinGap}\norm{F - \FiOpt}
    \qquad
    ,
    \forall F \in \DistSetDelta.
\end{align*}
\end{definition}
\noi This assumption will be seen to hold for our examples,  in particular, the following result gives mild sufficient conditions.   See proof in \cref{appendix:proofsOfregretBound}. \vskip 1em

\begin{proposition}
\label{prop:linearGap}
    Suppose $\RHat$ is convex over $\DistSetDelta$ and $\Di > 0$ for all $i \neq i^*$, then $\RHat$ has a linear gap with $\pLinGap = \gapRatio$, where $\gapRatio$ is defined in \cref{eq:diameterAndGapRatio}.
\end{proposition}
\vskip 1em

Summarizing our observations thus far, the informal statement of our main findings is made concrete by the following result. \vskip 1em 

\begin{theorem}[$\mathbf{\UUCB}$ \textbf{regret}]
\label{thm:regretBound}
    Suppose an \EDRMabbrv, $\RHat$, is stable, smooth with ${M_0 = \infty}$,
    and has a linear gap.
    Letting
    $
	    \lipConst
	    =
	    \polyContModCoeff \brk{1 + \distSetDiameter^{\polyContModDeg - 1}}
	    ,
    $
	the regret of running $\UUCB$ is bounded as
	\begin{align*}
	    \regret[{\UUCB}]
    	\le
    	\frac{\lipConst \gapRatio}{T} 
    	\sum_{i \neq i^*} \brk*{
    	    \frac{\gamma \Di \log T}{\phiFuncAt{\Di / 2}}
    	    +
    	    \frac{\gamma + 6}{\gamma - 2} \Di
	    }
	    +
	    \frac{12\pSmooth K^2 \log KT}{\concentrationConst T}
	    ,
	\end{align*}
	for all $T \ge T_1$, which is polynomial in problem parameters.
\end{theorem}
\vskip 1em 
\noi For a detailed proof, which includes the exact dependence of $T_1$ on the problem parameters, and an additional low order term when $M_0 < \infty$, see \cref{appendix:proofsOfregretBound}. \vskip 1em 
\begin{proof}[(sketch)]
First, using \cref{thm:smoothHorizonGap} with $F = \FiOpt$ and large enough $T$ we get that
\begin{align*}
    &J_1(T)
    =
    \EEBrk{
        \RHatFunc{\FHatPi[\optPolicyT]{T}}
        -
        \RHatFunc{\FProxPi[\optPolicyT]{T}}
    }
    \le
    \frac{4\pSmooth K^2 \log KT}{\concentrationConst T}
	+
	\frac{1}{\pLinGap}\EE{\norm{\FProxPi[\optPolicyT]{T} - \FiOpt}}
	,
	\\
    &J_3(T)
    =
    \EEBrk{
        \RHatFunc{\FProxPi[\UUCB]{T}}
        -
        \RHatFunc{\FHatPi[\UUCB]{T}}
    }
    \le
    \frac{6\pSmooth K^2 \log KT}{\concentrationConst T}
	,
\end{align*}
where for $J_3(T)$ we bounded the second term of \cref{thm:smoothHorizonGap} using the fact that \cref{theorem:UUCB} actually bounds
$
    \lipConst \EE{\norm{
        \FProxPi[\UUCB]{T}
        -
        \FiOpt
    }}
    .
$
Second, using the linear gap assumption we get that
\begin{align*}
    J_2(T)
    =
    \EEBrk{
        \RHatFunc{\FProxPi[\optPolicyT]{T}}
        -
        \RHatFunc{\FiOpt}
    }
    \le
    -\frac{1}{\pLinGap}\EE{\norm{\FProxPi[\optPolicyT]{T} - \FiOpt}}
    .
\end{align*}
Finally, recall that in \cref{eq:regretDecomp} we decompose the regret as
$
    \regret
    =
    \pseudoRegret
    +
    J_1(T)
    +
    J_2(T)
    +
    J_3(T)
    .
$
Combining the above and using \cref{theorem:UUCB} to bound $\pseudoRegret[\UUCB]$ concludes the proof.
\end{proof}

\begin{remark}
\label{remark:sqrtRegret}
    Notice that even in the absence of the smoothness and linear gap assumptions, we may still apply \cref{prop:stableHorizonGap} to obtain a weaker regret bound in which the last two terms of \cref{thm:regretBound} are replaced by
    $
        8 \polyContModCoeff \sqrt{
        {K^2 \log KT}/{\concentrationConst T}
        }
        .
    $
    In \cref{sec:examples} we will show that typical examples satisfy \cref{thm:regretBound}, however, we also give two cases where this weaker bound is the best that can be achieved within  our framework.
\end{remark}

\paragraph{Example: Average Reward}
We summarize our approach for the familiar bandit average reward setting that, in our \EDRMabbrv\space formulation, is given by
$
    \RHatFunc[\text{ave}]{F}
    =
    \int_{\RR} x dF(x)
    .
$
Notice that $\RHat[\text{ave}]$ is linear and so the regret decomposition in \cref{eq:regretDecomp} becomes trivial, i.e., $J_1=J_2=J_3=0$ and $\regret = \pseudoRegret$.
For simplicity, suppose that the rewards are constrained to the interval $\brk[s]{0, 1}$, and consider the seminorm
$
    \norm{F}
    =
    \abs{\RHatFunc[\text{ave}]{F}}
.
$
Notice that
$
    \Di
    =
    \norm{\FiOpt - \Fi}
    ,
$
and thus $\gapRatio = 1$.
Using Hoeffding's inequality we get that $\RHat[\text{ave}]$ is stable with 
$
    \concentrationConst = 2, 
    \polyContModCoeff = 1/2,
    \polyContModDeg = 1
    ,
$
and thus $\lipConst = 1$.
Since $\RHat[\text{ave}]$ is linear, it is clearly smooth with $\pSmooth=0$ and $M_0 = \infty$. Plugging all parameters into \cref{thm:regretBound} we recover the standard regret bound for average reward bandits
\begin{align}
\label{eq:linearRegretBound}
    \regret[{\UUCB}]
    \le
    \frac{1}{T} 
	\sum_{i \neq i^*} \brk*{
	    \frac{2 \gamma \log T}{\Di} + \Di \frac{\gamma + 6}{\gamma - 2}
    }
    .
\end{align}

\paragraph{Discussion}
Our main result demonstrates that the \EDRMabbrv\space formulation allows us to convert difficult questions in learning under sequential performance criteria to, essentially, simple questions in functional analysis. Roughly speaking, any quasiconvex function, $\RHat$, that, for an appropriate seminorm, is twice differentiable and grows at most polynomially, can be accommodated by our framework (at least for some arm distributions), i.e., may be learned efficiently.
We note that our results focused solely on the time horizon parameter $T$, and we suspect that the dependence on the number of arms $K$ can be improved. The main issue there is the squared dependence on $K$ in the tail bound of the empirical distribution (\cref{lemma:empDistConcentration}), and subsequently in the horizon gap (\cref{thm:smoothHorizonGap}). It is not clear whether this could be improved uniformly over all policies $\policy \in \policiesSet$, or whether this should be done only for near optimal policies. As a motivating example, it is clear that a single arm policy has horizon gap that does not depend on $K$. As the optimal policy is close to a single arm policy, we expect that its dependence on $K$ should be weak, perhaps even sub-linear. This would make the horizon gap a low order term compared to the pseudo regret and establish an equivalence between the two notions of regret. We leave this as an open question for future work.
So far, we demonstrated how the standard average reward setting fits into our framework. In the following section we use our framework to analyze various well-known performance measures from the risk literature.

\section{Illustrative Examples} \label{sec:examples}
The purpose of this section is, first and foremost, to show the relative ease with which various performance criteria can be analyzed within the framework developed in the previous sections. To make the exposition more accessible, we forego detailed introductions of the various criteria as well as various other technical details. Our main focus is to show the use of \cref{thm:regretBound}, after which we give some edge cases that demonstrate the subtleties and limitations of our framework. We refer the interested reader to \cref{appendix:examplesDetails} for the complete details\rev{, and to \cref{table:1} for a summary of the parameter settings of each criterion}{}. We make the following assumption throughout this section. \vskip 1em 

\begin{assumption}
\label{assumption:boundedReward}
    The rewards are restricted to the interval $\brk[s]{0,1}$, i.e., the support of $\Fi$ is in $\brk[s]{0,1}$ for all $i \in \actionSet$.
\end{assumption}
\vskip 1em 
This assumption is intended to simplify the exposition and can always be replaced by an appropriate ``light tailed"  condition.

\begin{table}[t]
\begin{tabular}{l|lllllll}
              & $b$ & $q$ & $\concentrationConst$ & $\lipConst$ & $\gapRatio$ & $\pLinGap$ & $\pSmooth$ \\
\hline
Average       & $1/2$                                                                                      & $1$    & $2$          & $2b$                & 1                                               & $\gapRatio$    & 0                                         \\
TSV           & $1/2$                                                                                      & $1$    & $2/r^4$      & $2b$                & 1                                               & $\gapRatio$    & 0                                         \\
Entropic Risk & $\frac{1}{\theta}\exp^{\theta}$                                                            & $1$    & $2$          & $2b$                & $\theta \exp^{-\theta \RHat[ent]_{\min}}$       & $\gapRatio$    & $\frac{1}{\theta} \exp^{2\theta}$         \\
Variance      & $\sqrt{5}$                                                                                 & $2$    & $1/2$        & $2b$                & $\sqrt{2}/\Delta_{\min}$                        & $\gapRatio$    & 2                                         \\
Mean-Variance & $2(1 + \rho)$                                                                              & $2$    & $1/2$        & $\frac52 b$          & $\sqrt{2}/\Delta_{\min}$                        & $\gapRatio$    & $2\rho$                                   \\
Sortino Ratio & $\frac{\abs{r}+2}{4 \epsilon_{0}^{3/2}}$                                                   & $2$    & $1/2$      & $2b$                      & $\sqrt{1+r^4}/\Delta_{\min}$                    & $\gapRatio\sqrt{1+\frac{r^2}{\epsilon_0}}$    & $\frac{2\epsilon_0 M_0 + \abs{r} + 1}{4 \epsilon_0^{5/2}}$                                     \\
$\CVAR$       & $\frac{4}{\alpha \min\brk[c]*{\alpha, 1-\alpha}}$                                          & $2$    & $2/3$      & $2b$                      & $1/\Delta_{\min}$                               & $\gapRatio$                                   & $\frac{2\varBa}{\alpha}$                                     \\
$\VAR$        & $\max\brk[c]*{\varBa, \frac{\varMa + 2}{\varMa \min\brk[c]{\alpha,1-\alpha}}}$        & $1$    & $2/3$      & $2b$                      & $1/\Delta_{\min}$                               & ------                                        & ------      
\end{tabular}
\caption{\label{table:1}\rev{Worst case parameters for \cref{prop:stableHorizonGap,theorem:UUCB,thm:regretBound} under the assumption that arm distributions are supported on $[0,1]$. $\Delta_{\min}$ is the minimum over the gaps $\Delta_i$, and $\RHat[ent]_{\min}$ is the minimal value for Entropic risk, which is simply $0$ in the worst case.}{}}
\end{table}

\subsection{Linear \EDRMabbrv s}
We begin with with a few examples of linear \EDRMabbrv s, which are essentially standard stochastic multi-armed bandit settings with augmented arm distributions.

\paragraph{Average reward}
is the classic bandit performance criterion, which is given by
$
    \RHatFunc[\text{ave}]{F}
    =
    \int_{\RR} x dF(x)
    .
$
For a given reward sequence, it is explicitly stated as
\begin{align*}
    \RHatFunc[\text{ave}]{\FHatPi{T}}
    =
    \frac{1}{T} \sum_{t=1}^{T} \XtPi
    .
\end{align*}

\paragraph{Squared reward}
is a less typical performance measure on its own but will serve us in what follows. It is given by
$
    \RHatFunc[\text{sqr}]{F}
    =
    \int_{\RR} x^2 dF(x)
    ,
$
and in terms of the reward sequence as
\begin{align*}
    \RHatFunc[\text{sqr}]{\FHatPi{T}}
    =
    \frac{1}{T} \sum_{t=1}^{T} (\XtPi)^2
    .
\end{align*}

\paragraph{Below Target Semi-Variance (TSV)}
measures the negative variation from a threshold parameter $r \in \RR$. The goal here is to minimize the variation and since our setting is expressed in terms of maximization, we state its negation
$
    \RHatFunc[\text{TSV}]{F}
    =
    -\int_{\RR} (x-r)^2 \indEvent{x \le r} dF(x)
    .
$
In terms of the reward sequence this stated as
\begin{align*}
    \RHatFunc[\text{TSV}]{\FHatPi{T}}
    =
    - \frac{1}{T} \sum_{t=1}^{T} (\XtPi - r)^2 \indEvent{\XtPi \le r}
    .
\end{align*}

\paragraph{The Analysis}
for all linear \EDRMabbrv s follows in a similar fashion to the average reward demonstrated in \cref{sec:regretBound}, with the only potential change being the value of $\concentrationConst$. More formally, let $\RHat[\text{lin}]$ be any linear \EDRMabbrv. We define the seminorm 
$
    \norm{F}
    = 
    \abs{\RHatFunc[\text{lin}]{F}}
    ,
$
which implies that
$
    \Di
    =
    \norm{\FiOpt - \Fi}
    ,
$
and consequently $\gapRatio = 1$. Requirement~\ref{item:StrongEDRMCond1} of stability clearly holds with 
$
    \polyContModCoeff = 1/2,
    \polyContModDeg = 1
    ,
$
and thus $\lipConst = 1$.
Recalling the empirical distribution and indicator functions from \cref{eq:empDistMap}, Hoeffding's inequality implies Requirement~\ref{item:StrongEDRMCond2} of stability holds with $\concentrationConst = 2 / \vartheta_{\text{lin}}$ where
\begin{align*}
    \vartheta_{\text{lin}}
    =
    \max_{x,y \in \brk[s]{0,1}}
    \brk[s]*{
        \RHatFunc[\text{lin}]{\indFunc{x}}
        -
        \RHatFunc[\text{lin}]{\indFunc{y}}
    }^2
    ,
\end{align*}
is the squared length of the reward interval under $\RHat[\text{lin}]$, which in the examples above is at most $1$.
Since $\RHat[\text{lin}]$ is linear, it is clearly smooth with $\pSmooth=0$ and $M_0 = \infty$. Plugging all parameters into \cref{thm:regretBound} we recover the standard bandit regret bound given in \cref{eq:linearRegretBound}.

\subsection{Composite \EDRMabbrv s}
Moving on to more complex performance criteria, we consider compositions of linear \EDRMabbrv s. Such criteria are often used to state a trade-off between multiple objectives. A partial list of widely used risk metrics which we consider here consists of: Entropic Risk, Variance, Mean-Variance (Markowitz), Sharpe ratio, and Sortino ratio. 
Formally, we say an \EDRMabbrv\space $\RHat[\hComp]$ is composite if there exist linear \EDRMabbrv s 
$
    \RHat[(1)], \ldots, \RHat[(n)]
$
and
$\hComp : \RR[n] \to \RR$ such that
\begin{align*}
    \RHatFunc[\hComp]{F}
    =
    \hCompFunc{
        \RHatFunc[(1)]{F}, \ldots, \RHatFunc[(n)]{F}
    }
    .
\end{align*}
Considering this class under the seminorm
$
    \norm{F}
    =
    \norm{\brk{
        \RHatFunc[(1)]{F}, \ldots, \RHatFunc[(n)]{F}
    }}_2
    ,
$
where $\norm{\cdot}_2$ is the $\ell^2$ norm in $\RR[n]$, 
the verification of our framework becomes easy, as seen in the following result.\vskip 1em 
 
\begin{lemma}[\textbf{Informal}] \label{lemma:compositeInformal}
    Suppose $\RHat[(1)], \ldots, \RHat[(n)]$ are linear, and stable, then:
	\begin{enumerate}
		\item If $h$ admits a polynomial local modulus of continuity,
		then $\RHat[\hComp]$ is stable;
		\item If $\hComp$ is locally smooth,
		then $\RHat[\hComp]$ is smooth;
		\item If $\hComp$ is convex then so is $\RHat[\hComp]$.
	\end{enumerate}
\end{lemma}
\vskip 1em 
\noi The formal statement along with its proof may be found in \cref{appendix:composite}.
Verifying \cref{lemma:compositeInformal} is typically very easy, often amounting to bounding the gradient and Hessian of $\hComp$, and yields all the needed properties to invoke \cref{thm:regretBound} and obtain an $\mathcal{O}(\log T / T)$ regret bound for $\UUCB$.

\paragraph{Entropic risk}
is a risk assessment measure that uses an exponential utility function with risk aversion parameter $\theta > 0$.
It is given by
\begin{align*}
    \RHatFunc[\text{ent}]{F}
    =
    -\frac{1}{\theta} \log \brk*{
        \int_{\RR}\exp\brk*{-\theta x} dF(x)
    }
    =
    -\frac{1}{\theta} \log \brk*{\RHatFunc[\text{exp}]{F}}
    ,
\end{align*}
where 
$
    \RHatFunc[\text{exp}]{F}
    =
    \int_{\RR}\exp\brk*{-\theta x} dF(x)
$
is a linear \EDRMabbrv\space and thus $\RHat[\text{ent}]$ is composite with $\hCompFunc{x} = -\frac{1}{\theta} \log x$, which is convex. Since the rewards are in $\brk[s]{0,1}$, we can bound the derivatives of $\hComp$ to conclude that it satisfies \cref{lemma:compositeEDRM} with
$
    \polyContModCoeff
    =
    \frac{1}{2\theta}\exp(\theta)
    ,
    \polyContModDeg
    =
    1
    ,
    \concentrationConst
    =
    2
    ,
    \pSmooth
    =
    \frac{1}{\theta}\exp(2\theta)
    ,
$
and $M_0 = \infty$.

\paragraph{Variance}
measures the empirical squared deviation from the mean reward. As we seek to minimize this deviation, it is given by
\begin{align*}
    \RHatFunc[\text{var}]{F} 
    =
    -\brk[s]*{
        \RHatFunc[\text{sqr}]{F}
        -
        \brk[s]*{\RHatFunc[\text{ave}]{F}}^2
    }
    ,
\end{align*}
and thus $\hCompFunc{x_1, x_2} = x_1^2 - x_2$, which is convex. It is then easy to verify that \cref{lemma:compositeEDRM} holds with 
$
    \polyContModCoeff
    =
    \sqrt{5}
    ,
    \polyContModDeg
    =
    2
    ,
    \concentrationConst
    =
    1/2
    ,
    \pSmooth
    =
    2
    ,
$
and $M_0 = \infty$.

\paragraph{Mean-variance (Markowitz)}
measures performance as an additive trade-off between the empirical mean and variance. For $\rho \ge 0$ it is given by
\begin{align*}
    \RHatFunc[\text{MV}]{F}
    =
    \RHatFunc[\text{ave}]{F}
    +
    \rho \RHatFunc[\text{var}]{F}
    ,
\end{align*}
and thus $\hCompFunc{x_1, x_2} = x_1 + \rho (x_1^2 - x_2)$, which is convex. A simple calculation shows that \cref{lemma:compositeEDRM} holds with
$
    \polyContModCoeff
    =
    2(1 + \rho)
    ,
    \polyContModDeg
    =
    2
    ,
    \concentrationConst
    =
    1/2
    ,
    \pSmooth
    =
    2 \rho
    ,
$
and $M_0 = \infty$.

\paragraph{Sharpe ratio}
measures performance as a ratio between the empirical mean and variance. For $r \in \RR$ and $\varepsilon_0 > 0$ it is given by 
\begin{align*}
    \RHatFunc[\text{Sh}]{F}
    =
    \frac{
        \RHatFunc[\text{ave}]{F} 
        -
        r
    }{\sqrt{{\varepsilon_0}-\RHatFunc[\text{var}]{F}}}
    ,
\end{align*}
and thus 
$
    \hCompFunc{x_1, x_2}
    =
    (x_1 - r) / \sqrt{\varepsilon_0 - x_1^2 + x_2}
    .
$
The parameter $r$ is essentially a threshold for the average reward, while $\varepsilon_0$ is a regularization parameter. Unlike the previous examples, where $\hComp$ was convex, here it is only quasiconvex. While a quasiconvex function could potentially have no linear gap, we show that Sharpe ratio has a linear gap with a slightly worse constant. The calculation is technical and deferred to \cref{appendix:examplesDetails}.

\paragraph{Sortino ratio}
is Sharpe ratio with variance replaced by below target semi-variance. As such, for $r \in \RR$ and $\varepsilon_0 > 0$ it is given by 
\begin{align*}
    \RHatFunc[\text{So}]{F}
    =
    \frac{
        \RHatFunc[\text{ave}]{F}
        -
        r
    }{\sqrt{{\varepsilon_0}-\RHatFunc[\text{TSV}]{F}}}
    ,
\end{align*}
and thus
$
    \hCompFunc{x_1, x_2}
    =
    (x_1 - r) / \sqrt{\varepsilon_0 - x_2}
    .
$
Similar to Sharpe ratio, here $\hComp$ is also quasiconvex. The resulting analysis is thus similar, if perhaps a bit simpler, and we defer it to \cref{appendix:examplesDetails}.

\subsection{Non-composite \EDRMabbrv s}

We now consider two examples of non-composite criteria. The first, $\CVAR$, is found to be smooth and stable under appropriate conditions. The second, $\VAR$, is stable but appears to be non-smooth. In both cases the resulting conditions possess a more particular nature than those presented for composite \EDRMabbrv s. 
\rev
{
(In \cref{appendix:examplesDetails} we also include some results for Utility-Based Shortfall Risk (UBSR)).
}{}

\paragraph{Conditional Value at Risk ($\mathbf{\CVAR}$)}
is the average reward below percentile level $\alpha \in \brk{0, 1}$, which is given by
\begin{align}
\label{eq:cvarDef}
   \RHatFunc[\CVAR]{F}
    =
        \max_{z \in \RR} 
            z
            -
            \frac{1}{\alpha} \int_{-\infty}^{z} F(x) dx
	.
\end{align}
We note that a more explicit expression can be obtained by plugging in the maximizer $z^* = \RHatFunc[\VAR]{F}$, which is the Value at Risk of $F$, defined in \cref{eq:varDef}.
Now, in order to invoke \cref{thm:regretBound} we need to show that $\CVAR$ is convex, stable and smooth.

Convexity is immediate since the expression in \cref{eq:cvarDef} is a maximum over linear functions, which is convex. For stability, we use the norm
\begin{align}
\label{eq:cvarNorm}
    \norm{F}
    =
    \max\brk[c]*{
        \norm{F}_\infty
        ,
        \abs2{\int_{-\infty}^{0} x dF(x)}
        ,
        \abs2{\int_{0}^{\infty} x dF(x)}
    }
    ,
\end{align}
where
$
    \norm{F}_\infty
    =
    \max_{x \in \RR} \abs{F(x)}
$
is the $\ell^{\infty}$ norm. The two additional terms may be foregone when the rewards are constrained to $\brk[s]{0,1}$.
The concentration of $\norm{F}_\infty$ follows from the Dvoretzky-Kiefer-Wolfowitz inequality \citep{massart1990tight}, while the other two follow from Hoeffding's inequality. A further technical calculation yields the desired modulus of continuity and thus stability is concluded with parameters
$
    \polyContModCoeff
    =
    {4}/{\alpha \min\brk[c]*{\alpha,1-\alpha}}
    ,
    \polyContModDeg
    =
    2
    ,
    \concentrationConst
    =
    2/3
    .
$
Recalling \cref{remark:sqrtRegret}, we can now conclude that $\UUCB$ obtains regret $\mathcal{O}(\sqrt{\log T / T})$ for any bounded reward distributions.

We would like to show that $\CVAR$ is smooth and thus conclude the requirements for \cref{thm:regretBound}. However, we find that even for bounded distributions, $\CVAR$ may be non-smooth (see \cref{fig:1}). To overcome this limitation, we require that the arm distributions have a positive density around their $\alpha$ percentile.  Formally, we require that there exist $\varBa >0, \varMa \ge \distSetDiameter$ such that
\begin{align}
\label{eq:positivePDF}
    \abs{F \brk*{\RHatFunc[\VAR]{F} + \varBa y} - \alpha}
    \ge
    \abs{y}
    \qquad
    ,
    \forall
    F \in \DistSetDelta
    ,
    y \in \brk[s]*{- \varMa, \varMa}
    ,
\end{align}
where $\RHatFunc[\VAR]{F}$ is the Value at Risk of $F$, defined in \cref{eq:varDef}. This condition is one of the subtleties that arise from our framework. With it, we show that $\CVAR$ is smooth with $\pSmooth = 2\varBa / \alpha$ and $M_0 = \varMa$ and thus obtain the desired $\mathcal{O}(\log T / T)$ regret bound. Without it, we provide a numerical experiment (see \cref{fig:1}), suggesting that the horizon gap truly behaves as $\Omega(\sqrt{1 / T})$ as suggested by our framework.

\paragraph{Value at Risk ($\mathbf{\VAR}$)}
is the reward at percentile $\alpha \in \brk{0,1}$, which is given by
\begin{align} 
\label{eq:varDef}
	\RHatFunc[\VAR]{F}
	=
	\inf_{x \in \RR}\brk[c]*{x ~\big|~ F(x) \ge \alpha}
	.
\end{align}
We show that $\VAR$ is quasiconvex, however, as previously mentioned, we could not find any set of conditions that ensure the smoothness of $\VAR$, and thus we cannot  invoke \cref{thm:regretBound}. 
On a positive note, we show that for distributions satisfying \cref{eq:positivePDF}, stability holds under the semi-norm in \cref{eq:cvarNorm} and with parameters
$
    \polyContModCoeff
    =
    \max\brk[c]*{
        \varBa
        ,
        \brk{
            \varMa
            +
            2
        }/{\min\brk[c]*{\alpha,1-\alpha}\varMa}}
    ,
    \polyContModDeg
    =
    1
    ,
    \concentrationConst
    =
    2/3
    .
$
We conclude that $\mathcal{O}(\sqrt{\log T / T})$ regret is obtainable by our framework (by means of \cref{remark:sqrtRegret}). We conjecture that improved regret is possible for $\VAR$ when the arm distributions are also twice differentiable, but it is not clear whether this could be achieved through the smoothness condition.

Notice that the stability issues of $\VAR$ are such that even \cref{theorem:StableEDRMoptPolicy} (single arm infinite horizon oracle) is not satisfied without further assumptions on the arm distributions. Concretely, denote the $\alpha$ level set of a function $F \in \DistSet$ by
$
    \levelSet{F}
    =
    \brk[c]{
        x \in \RR \big| F(x) = \alpha
    }
    ,
$
then \cref{theorem:StableEDRMoptPolicy} holds if 
$
    \abs{\levelSet{\alpha}}
    \le
    1
$
for all $F \in \DistSetDelta$. Intuitively, this condition ensures that the arm distributions do not have a flat region at their $\alpha$ percentile. If such a flat region exists, then arbitrarily small perturbations to the distribution may change the percentile by a constant, thus causing the instability issue. Interestingly, even in the presence of this instability, we have the following result. \vskip 1em 

\begin{proposition}[$\mathbf{\VAR}$ \textbf{oracle policy}]
\label{proposition:VAROptPolicy}
		For $\alpha \in \brk*{0,1}$, $\VAR$ always admits a \simple\space oracle policy $\optPolicy$, i.e., choosing a single arm throughout the horizon is asymptotically optimal.
\end{proposition}

\subsection{A numerical illustration of an edge case}
\begin{figure}[t!]
	\centering
	\includegraphics[width=0.32\linewidth]{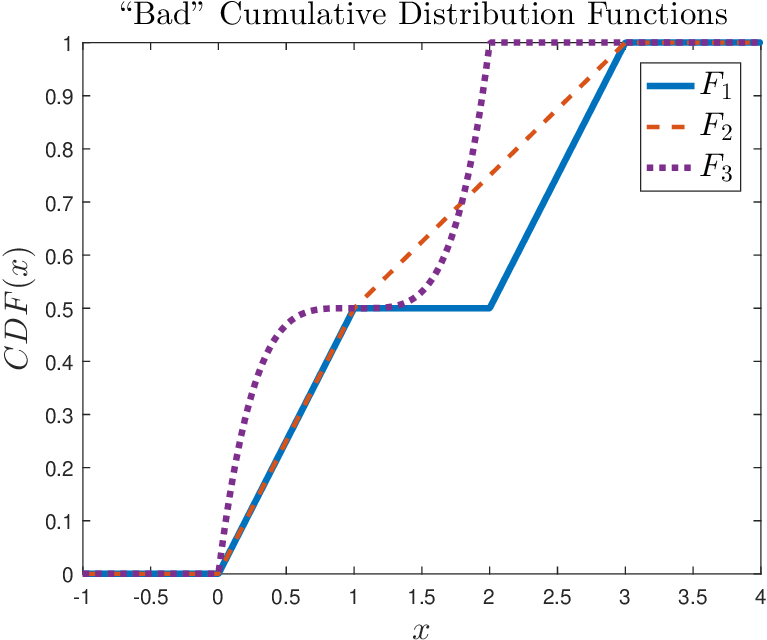}~~
	\includegraphics[width=0.32\linewidth]{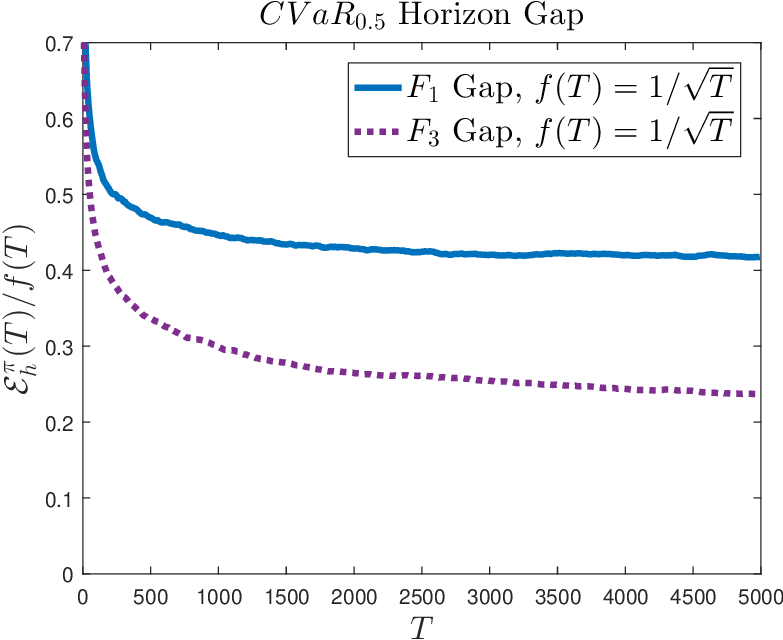}~~
	\includegraphics[width=0.32\linewidth]{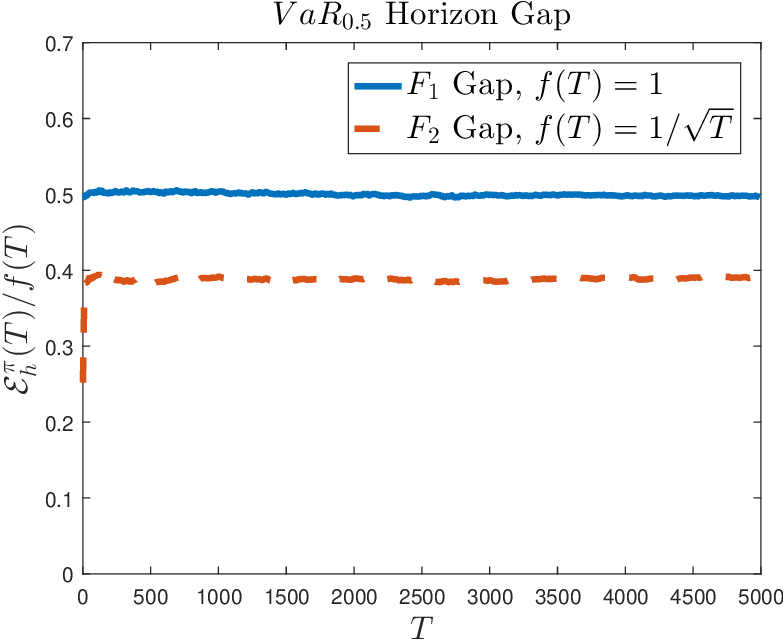}
	\caption{\label{fig:1}
    	$\VAR$ and $\CVAR$ horizon gap for ``bad'' distributions. ($F_1$) has 
    	$
    	    \left| \levelSet[0.5]{F} \right|
    	    >
    	    1
	    $
    	and has no density to the right of the percentile; ($F_2$) has no density around percentile $\alpha = 0.5$;  and ($F_3$) is not differentiable. The figures essentially show that $\lim_{T \to \infty} J_1(T) / f(T) = c > 0$ thus claiming that $J_1(T)$ behaves as $f(T)$, which is slower than the desired $\mathcal{O}(\frac{\log T}{T})$.
	}
\end{figure}

When considering the existence of \simple\space oracle policies, \cref{proposition:VAROptPolicy} essentially means that stability, while a sufficient condition, is not necessary. However, for the purpose of regret analysis, we highlight the importance of stability by means of a simulation. Note that \cref{thm:regretBound} relies on a suitably fast  diminishing horizon gap (see $J_1(T), J_3(T)$ in \cref{eq:regretDecomp}). We calculate this gap in a simple simulation with $K=1$ arms. This is done for three different distributions, each not satisfying a different subset of the previously discussed conditions for stability and or smoothness. \cref{fig:1} displays the simulation results, which show that the obtained rate is slower than the desired $\frac{\log T}{T}$ which is achieved in \cref{thm:smoothHorizonGap}.

\rev{
\section{Limitations}
\label{sec:limitations}
While our framework encompasses many known performance measures, it has limitations that make it inapplicable to certain potentially interesting examples. The following outlines some of these limitations and indicates how they may be overcome.
\paragraph{No ``single optimal arm''.} While all of the examples in \cref{sec:examples} satisfy this assumption, formally defined in \cref{eq:singleOpt}, this is not always the case. For example, consider the performance measure
\begin{align}
\label{eq:RhatwDef}
    \RHatFunc[w]{F}
    =
    \int_0^\infty w(1 - F(x)) dx
    ,
\end{align}
which is a type of ``distorted risk measure'' with distortion function $w: [0,1] \to [0,1]$. If we take $w(p) = p(1-p)$ we get that $\RHat[w]$ is concave and thus typically obtains its maximum in an interior point. More concretely, suppose we have two arms with Bernoulli distributions $F_1, F_2$ with parameters $p_1 \le p_2$,  respectively. Then $\DistSetDelta$, the set of convex combinations, is exactly the Bernoulli distributions with parameter $p \in [p_1, p_2]$. Next, it is easy to see that for a Bernoulli distribution with parameter $p$, we have that $\RHatFunc[w]{F} = p (1-p)$, which is strictly concave and obtains its maximum at $p=1/2$. We conclude that if $1/2 \in (p_1, p_2)$ then for $\lambda^*$ such that $\lambda^* p_1 + (1-\lambda^*) p_2 = 1/2$
\begin{align*}
    \RHatFunc[w]{\lambda^* F_1 + (1-\lambda^*) F_2}
    >
    \max \brk[c]*{\RHatFunc[w]{F_1}, \RHatFunc[w]{F_2}}
    ,
\end{align*}
which contradicts \cref{eq:singleOpt}. Overall, the above implies that there is no single-arm infinite horizon oracle policy, i.e., one that always chooses a single arm. An example of an infinite horizon oracle in this setting is one that chooses arm $1$ with probability $\lambda^*$ and arm $2$ otherwise.
\\
The above performance measure is admittedly somewhat contrived: it prefers a Bernoulli distribution with $p=1/2$ over one with $p=1$, even though the latter has stochastic dominance over the former. Regardless, it is worth emphasizing here that  \cref{prop:stableHorizonGap} holds regardless of \cref{eq:singleOpt} and thus it only remains to design a pseudo regret minimization algorithm to obtain $1 / \sqrt{T}$ regret. We suspect this could be achieved by a similar UCB-type algorithm that optimizes over all convex combinations of the arms. We note that the resulting optimization problem might be difficult to solve and thus may require additional assumptions to make it computationally tractable.
\paragraph{Unstable performance measures.} As seen in the example of $\VAR$ in \cref{sec:examples}, some performance measures could be unstable without additional assumptions on the arm distributions. 
The following is another such example. Consider $\RHat[w]$ from \cref{eq:RhatwDef} with $w(p) = \sqrt{p}$ with $F_1, F_2$ Bernoulli as before. If $p_1 = 0$ then we get that
\begin{align*}
    \RHatFunc[w]{F_2} - \RHatFunc[w]{F_1}
    =
    \sqrt{p_2}
    .
\end{align*}
On the other hand, by the homogeneity of the semi-norm we have that
\begin{align*}
    \norm{F_1 - F_2}
    =
    \norm{p_2 \mathds{1}_{[0,1)}}
    =
    p_2 \norm{\mathds{1}_{[0,1)}}
    ,
\end{align*}
and thus for any norm, and $\polyContModCoeff, \polyContModDeg$ there exists $p_2$ small enough such that
\begin{align*}
    \RHatFunc[w]{F_2} - \RHatFunc[w]{F_1}
    =
    \sqrt{p_2}
    >
    \polyContModDeg\brk*{
    p_2 \norm{\mathds{1}_{[0,1)}} + p_2^{\polyContModDeg} \norm{\mathds{1}_{[0,1)}}^{\polyContModDeg}}
    =
    \polyContModDeg\brk*{
    \norm{F_1 - F_2} + \norm{F_1 - F_2}^\polyContModDeg}
    ,
\end{align*}
i.e., $\RHat[w]$ does not satisfy \cref{definition:StrongEDRM}. This issue could potentially be overcome by weakening the Lipschitz condition in \cref{definition:StrongEDRM} to a H\"older type condition. This type of change will clearly impact  the overall performance result.
\paragraph{Heavy-tailed distributions.} This subject is well understood in the classic bandit setting \cite{bubeck2012regret}, and has also been studied for general learning settings under a $\CVAR$ performance measure \cite{holland2020learning}. Our results hold for light-tailed sub-Gaussian arm distributions. We suspect it is  possible to extend them to sub-Exponential distributions without significant degradation, but this will not carry over to  heavier tails that will likely necessitate a different approach. 
}
{}

	\section{Open Problems and Future Directions} \label{sec:summary}
	One main question that we leave open is the dependence of the regret on the number of arms $K$. We conjecture that a finer analysis of the horizon gap may reduce it from our $K^2 \log K$ to either $K$ or $K \log K$. The subject of lower bounds remains open as well. Future directions may include a more complete taxonomy of performance criteria, or an extension of this framework to different settings (e.g., adversarial or contextual). Additionally, we note that the majority of our proof techniques also apply to non-quasiconvex criteria. If such criteria are found to be of interest then extending the framework to this case may be appealing.
	
	A future direction of great interest is to consider a Markov decision model for the dynamics. The same criteria of interest are still relevant, but now it is unclear whether a simple (Markov) policy could approximate the oracle and if so, at what rate.

	\acks{A preliminary version of this work appeared at the Conference on Learning Theory, 2018. We thank Ron Amit, Guy Tennenholtz, Nir Baram and Nadav Merlis for helpful discussions of this work, and the anonymous reviewers for their helpful comments.
	This work was partially funded by the Israel Science Foundation under contract 1380/16 and by the European Community's Seventh Framework Programme (FP7/2007-2013) under grant agreement 306638 (SUPREL).}

	\bibliography{bibliography} %

\begin{APPENDICES}
\crefalias{section}{appendix}

\section{\EDRMabbrv s as Permutation Invariant Performance Measures} \label{appendix:EDRMmotivation}
The following continues the discussion in \cref{sec:EDPMMotivation} on the motivation behind \EDRMabbrv s and their relation to permutation invariant performance measures. Let $\seqDef[1]{\Rt}{t=1}{\infty}$, where $\Rt : \RR[t] \to \RR$ is a function that measures the quality of a given reward sequence of length $t$. A decision maker may then wish to maximize the expected performance, i.e., $\EE{\RtFunc{\XtPi[1],\ldots,\XtPi[t]}}$. It makes sense that the preferences of the decision maker remain fixed over time. This means $\Rt$ ($t \ge 1$) should, in some sense, be time invariant. However, such an invariance is hard to grasp when the functions $\Rt$ do not share a domain. One way of addressing this issue is to assume that $\Rt$ is permutation invariant, i.e., it maps all the permutations of its reward sequence to the same value. We provide a formal definition in the proof of the following (known) result.

\begin{lemma}[Permutation invariant function representation] \label{lemma:PermutationInvarianceRepresentation}
	$\Rt$ is permutation invariant if and only if, there exists $\RHat_t : \DistSet \to \RR$ such that, $\RtFunc{x_1,\ldots,x_t} = \RHat_t \brk*{\FHattFunc[t]{x_1, \ldots, x_t}}$, where $\FHattFunc{\cdot}$ is the empirical distribution mapping defined in \cref{eq:empDistMap}.
\end{lemma}
The representation given in \cref{lemma:PermutationInvarianceRepresentation} suggests $\DistSet$ as a shared domain thus making it simple to define time invariance. We conclude that \EDRMabbrv s describe the objectives that are time and permutation invariant.

\begin{proof}[of \cref{lemma:PermutationInvarianceRepresentation}]
We start with a few definitions. Let $\Sigma_t$ denote the set of $t \times t$ permutation matrices (binary and doubly stochastic). $\Rt$ is said to be permutation invariant if $\RtFunc{\sigma x_{1:t}} = \RtFunc{x_{1:t}}$ for all $x_{1:t} \in \RR[t]$ and $\sigma \in \Sigma_t$. Let, $\EmpDistSetT = \setDef[\big]{\FHattFunc{x_{1:t}} \given x_{1:t} \in \RR[t]}$, be the set of empirical distributions created from $t$ elements (the image of $\FHatt$). Let,
\begin{align*}
\FHatt^{-1}\brk*{\hat{F}} = \setDef[\Big]{x_{1:t} \in \RR[t] \given \FHattFunc{x_{1:t}} = \hat{F}},
\end{align*}
be the inverse image of $\FHatt$ at $\hat{F} \in \EmpDistSetT$. Let,
\begin{align*}
\Sigma(x_{1:t}) = \setDef[\Big]{\sigma x_{1:t} \given \sigma \in \Sigma_t},
\end{align*}
be the set of all permutations of $x_{1:t}$. We can now begin the proof.

\textbf{First direction:} 
Suppose $\RtFunc{x_1,\ldots,x_t} = \RHat_t \brk*{\FHattFunc[t]{x_1, \ldots, x_t}}$. Notice that $\FHatt$ is indeed permutation invariant as permuting its input simply reorders its finite sum thus not changing the value. This clearly implies that $\Rt$ is permutation invariant.

\textbf{Second direction:}
Suppose that $\Rt$ is permutation invariant. Below, we prove that for any $x_{1:t} \in \RR[t]$
\begin{align}
\label{eq:FhatPermInvariance}
    \FHatt^{-1}\brk*{\FHattFunc{x_{1:t}})} 
    =
    \Sigma(x_{1:t})
    .
\end{align}
Assuming \cref{eq:FhatPermInvariance} holds true, define $g : \EmpDistSetT \to \RR[t]$ in the following way. For any $\hat{F} \in \EmpDistSetT$ choose arbitrarily $g\brk*{\hat{F}} \in \FHatt^{-1}(\hat{F}))$. Further define $\RHat_t : \DistSet \to \RR$ by,
\begin{align*}
\RHat_t(F) = 
\begin{cases}
\RtFunc{g(F)} & F \in \EmpDistSetT \\
0 & \mbox{otherwise.}
\end{cases}
\end{align*}
Then by \cref{eq:FhatPermInvariance} we have that, $g\brk*{\FHattFunc{x_{1:t}}} \in \FHatt^{-1}\brk*{\FHattFunc{x_{1:t}}} = \Sigma(x_{1:t})$, and thus there exists $\sigma_{g(x)} \in \Sigma_t$, such that $g\brk*{\FHattFunc{x_{1:t}}} = \sigma_{g(x)} x_{1:t}$. We conclude that,
\begin{align*}
\RHat_t \brk*{\FHatPi[]{t}(x_{1:t})}
= \RtFunc{g(\FHatPi[]{t}(x_{1:t}))}
= \RtFunc{\sigma_{g(x)} x_{1:t}}
= \RtFunc{x_{1:t}},
\end{align*}
where the last step uses the permutation invariance of $\Rt$. We now show that \cref{eq:FhatPermInvariance} holds, thus concluding the proof.

\textbf{Proof of \cref{eq:FhatPermInvariance}:}
Let $y_{1:t} \in \Sigma(x_{1:t})$ then there exists $\sigma \in \Sigma_t$ such that $y_{1:t} = \sigma x_{1:t}$. Since $\FHatt$ is permutation invariant then,
\begin{align*}
\FHattFunc{y_{1:t}}
= \FHattFunc{\sigma x_{1:t}}
= \FHattFunc{x_{1:t}}
\implies y_{1:t} \in \FHatt^{-1}\brk*{\FHattFunc{x_{1:t}}},
\end{align*}
and so $\Sigma(x_{1:t}) \subseteq \FHatt^{-1}\brk*{\FHattFunc{x_{1:t}}}$. On the other hand, let $y_{1:t} \in \FHatt^{-1}\brk*{\FHattFunc{x_{1:t}}}$, then we have that, $\FHattFunc{y_{1:t}} = \FHattFunc{x_{1:t}}$. Take $\sigma_x^*, \sigma_y^* \in \Sigma_t$ such that, $x_{1:t}^* = \sigma_x^* x_{1:t}$, $y_{1:t}^* = \sigma_y^* y_{1:t}$ are sorted in ascending order. Suppose in contradiction that $x_{1:t}^* \neq y_{1:t}^*$ and let,
\begin{align*}
s_0 = \min \setDef[\Big]{s \in \brk[c]*{1,\ldots,t} \given x_{s_0}^* \neq y_{s_0}^*}
\end{align*}
be the first index where $x_{1:t}^*$ and $y_{1:t}^*$ differ. Without loss of generality assume that $x_{s_0}^* < y_{s_0}^*$, then we have that,
\begin{align*}
\FHattFunc{y_{1:t}^*}(x_{s_0}^*)
= \frac{1}{t}\sum_{s=1}^{t}\indFuncAt{y_s^*}{x_{s_0}^*}
= \frac{1}{t}\sum_{s=1}^{s_0 - 1}\indFuncAt{y_s^*}{x_{s_0}^*}
= \frac{1}{t}\sum_{s=1}^{s_0 - 1}\indFuncAt{x_s^*}{x_{s_0}^*}
< \FHattFunc{x_{1:t}^*}\brk*{x_{s_0}^*},
\end{align*}
where the strict inequality follows since $\indFuncAt{x_{s_0}^*}{x_{s_0}^*} = 1$, and  if $s_0 = 1$, then the empty sum is in fact zero. This contradicts $\FHattFunc{y_{1:t}} = \FHattFunc{x_{1:t}}$ and so, $x_{1:t}^* = y_{1:t}^*$. Since, permutation matrices are invertible then, $y_{1:t} = {\sigma_y^*}^{-1} \sigma_x^* x_{1:t}$. It is well known that ${\sigma_y^*}^{-1}\sigma_x^*$ is always a permutation matrix. So, $y_{1:t} \in \Sigma(x_{1:t})$ and we conclude that $\FHatt^{-1}\brk*{\FHattFunc{x_{1:t}}} = \Sigma(x_{1:t})$, as desired. 
\end{proof}

\section{Proofs of Section~\ref{sec:infOracle}} 
\label{appendix:infHorizonOracle}
	
Denote the fraction of time at which arm $i$ was pulled by
\begin{align} \label{eq:ActionDistribution}
	\piT = \frac{\tauI}{T},
\end{align}
where $\tauI$ is defined in \cref{eq:tauIDef}.
Recall the definitions of $\FHatPi{T}$ and $\FHatPi[(i)]{T}$ given in \cref{eq:FhatPiAbbrv,eq:FhatiAbbrv}. The following Lemma is the main argument of the proof of \cref{theorem:StableEDRMoptPolicy}.

\begin{lemma}[$\FHatPi{T}$ sub-convergence] \label{lemma:FhatSubConvergence}
	Suppose
	$
        \lim_{t \to \infty} \norm{\FHatPi[(i)]{t} - \Fi}
        = 
        0
    $
    almost surely $\mbox{for all }  i \in \actionSet$.
    Let
    $p = \brk*{p_1,\ldots,p_K} \in \simplex$ 
    and
    $\seqDef{t_l}{l=1}{\infty}$
    be a random vector and subsequence. If
	\begin{align*}
		\lim_{l \to \infty} \norm{\piAt[]{t_l} - p} = 0 \qquad \textit{Almost Surely},
	\end{align*}
	Then
	\begin{align*}
	\lim_{l \to \infty} \norm{\FHatPi{t_l} - \Fp} = 0 \qquad \textit{Almost Surely},
	\end{align*}
	where $\Fp$ is defined in \cref{eq:distSetDeltaDef}.
\end{lemma}
\begin{proof}
	We rearrange the expression of $\FHatPi{T}$ such that the sum is over actions and instead of time:
	\begin{align*}
		\FHatPi{T}
		= \frac{1}{T}\sum_{t=1}^{T}\indFunc{\XtPi}
		= \sum_{i=1}^{K} \frac{\tauI}{T}\brk[s]*{\frac{1}{\tauI}\sum_{t=1}^{\tauI}\indFunc{\XtPi}}
		= \sum_{i=1}^{K} \piT \FHatPi[(i)]{\tauI}.
\end{align*}
Then we have that
\begin{align*}
	\norm{\FHatPi{t_l} - \Fp}& 
	= \norm{\sum_{i=1}^{K} \piAt{t_l} \FHatPi[(i)]{\tauIAt{t_l}} - p_{i}\Fi} \\
	&= \norm{\brk[s]*{\sum_{i=1}^{K} \piAt{t_l} \FHatPi[(i)]{\tauIAt{t_l}} - \piAt{t_l} \Fi} + \brk[s]*{\sum_{i=1}^{K} \piAt{t_l} \Fi - p_{i}\Fi}} \\
	&\le \norm{\sum_{i=1}^{K} \piAt{t_l} \brk*{\FHatPi[(i)]{\tauIAt{t_l}} - \Fi}} + \norm{\sum_{i=1}^{K} \Fi\brk*{\piAt{t_l} - p_{i}}} \\
	&\le \sum_{i=1}^{K} \piAt{t_l} \norm{\FHatPi[(i)]{\tauIAt{t_l}} - \Fi} + \sum_{i=1}^{K} \norm{\Fi}\abs{\piAt{t_l} - p_{i}} \\
	&\le \sum_{i=1}^{K} \underbrace{\piAt{t_l} \norm{\FHatPi[(i)]{\tauIAt{t_l}} - \Fi}}_{(*)} + \underbrace{K \max_{1 \le i \le K}{\norm{\Fi}}\norm{\piAt[]{t_l} - p}_\infty}_{(**) \to 0}.
\end{align*}
The first and second inequalities follow by the triangle inequality and homogeneity of norms. The third follows by H\"{o}lder's inequality.
By the Lemma's assumption $(**) \to 0$. We show that the same holds for (*). It is enough to show the convergence of the summands in order to conclude the overall convergence of this finite sum. By the Lemma's assumption we have that
\begin{align*}
\lim_{l \to \infty} \norm{\FHatPi[(i)]{\tauIAt{t_l}} - \Fi} =
\begin{cases}
	0 \quad&, \lim_{l \to \infty}\tauIAt{t_l} = \infty \\
	\norm{\FHatPi[(i)]{\tau} - \Fi} \quad&, \lim_{l \to \infty}\tauIAt{t_l} = \tau < \infty
\end{cases}
\qquad \textit{Almost Surely},
\end{align*}
where we used the fact that $\tauIAt{t}$ is non-decreasing and thus always converges. Now since both parts of (*) converge then we have that,
\begin{align*}
	\lim_{l \to \infty} \piAt{t_l} \norm{\FHatPi[(i)]{\tauIAt{t_l}} - \Fi}
	&= \lim_{l \to \infty} \piAt{t_l} \lim_{l \to \infty} \norm{\FHatPi[(i)]{\tauIAt{t_l}} - \Fi} \\
	&= \begin{cases}
	0 \quad&, \lim_{l \to \infty}\tauIAt{t_l} = \infty \\
	p_i \norm{\FHatPi[(i)]{\tau} - \Fi} \quad&, \lim_{l \to \infty}\tauIAt{t_l} = \tau < \infty
	\end{cases}
	\qquad \textit{Almost Surely}.
\end{align*}
Noticing that 
\begin{align*}
\lim_{l \to \infty}\tauIAt{t_l} = \tau < \infty \implies p_i = \lim_{l \to \infty} \piAt{t_l} = \lim_{l \to \infty}\frac{\tauIAt{t_l}}{t_l} = 0,
\end{align*}
the proof is concluded. 
\end{proof}

\begin{proof}[of \cref{theorem:StableEDRMoptPolicy}]
	The remainder of the proof consists of applying \cref{lemma:FhatSubConvergence}.
	\rev{Let $\statPolicy$ be a simple randomized policy that at each turn draws an arm $i.i.d$ with distribution $p \in \simplex$, i.e., $\PP{\statPolicyAt{t} = i} = p_i$.}{}
	We begin by proving $\EE{\RPi[\statPolicy]} = \RHatFunc{\Fp}$. Let $p \in \simplex$ define the \simple\space policy $\statPolicy$. Using the strong law of large numbers (\cite{Simonnet1996}) on each coordinate of $\piAt[]{t}$, we conclude that
	\begin{align*}
		\lim_{t \to \infty} \norm{\piAt[]{t} - p}_\infty = 0 \qquad \textit{Almost Surely}.
	\end{align*}
	Applying \cref{lemma:FhatSubConvergence} we get that
	\begin{align*}
		\lim_{t \to \infty} \norm{\FHatPi[\statPolicy]{t} - \Fp} = 0 \qquad \textit{Almost Surely}.
	\end{align*}
	Since $\RHat$ is assumed to be continuous, we have that,
	\begin{align*}
		\RPi[\statPolicy]
		= \liminf_{t \to \infty} \RHatFunc{\FHatPi[\statPolicy]{t}}
		\overset{a.s}{=} \lim_{l \to \infty}
		\RHatFunc{\FHatPi[\statPolicy]{t_l}}
		\overset{a.s}{=} \RHatFunc{\Fp},
	\end{align*}
	where $\seqDef{t_l}{l=1}{\infty}$ is the (random) subsequence that achieves the limit inferior. Taking expectation, we conclude that $\EE{\RPi[\statPolicy]}=\RHatFunc{\Fp}$. Now since $\simplex$ is compact and $\RHatFunc{\Fp}$ is continuous, then by the Weierstrass theorem we have that there exists $p^* \in \simplex$ such that,
	\begin{align} \label{eq:simpleOracleProof01}
		\RHatFunc{\FpOpt} = \max_{p \in \simplex} \RHatFunc{\Fp}.
	\end{align}
	We now show that $\statPolicy[p^*]$ is optimal thus concluding the first part of the proof. Let $\seqDef{t_m}{m=1}{\infty}$ be a (random) subsequence satisfying the limit inferior. Then we have that
	\begin{align*}
	\RPi
	= \liminf_{t \to \infty} \RHatFunc{\FHatPi{t}}
	\overset{a.s}{=} \lim_{m \to \infty} \RHatFunc{\FHatPi{t_m}}
	= (**).
	\end{align*}
	Noticing again that $\simplex$ is compact, we have that for any policy $\policy \in \policiesSet$, there exist $p \in \simplex$ and $\seqDef{t_l}{l=1}{\infty} \subseteq \seqDef{t_m}{m=1}{\infty}$ (both random) satisfying $\lim_{l \to \infty} \norm{\piAt[]{t_l} - p}_\infty = 0$ almost surely. Using \cref{lemma:FhatSubConvergence}, \cref{eq:simpleOracleProof01}, and the continuity of $\RHat$ we get,
	\begin{align*}
		(**)
		\overset{a.s}{=} \lim_{l \to \infty} \RHatFunc{\FHatPi{t_l}}
		\overset{a.s}{=} \RHatFunc{\Fp}
		\le \RHatFunc{\FpOpt}
		= \EE{\RPi[{\statPolicy[p^*]}]},
	\end{align*}
	and taking expectation we have $\EE{\RPi} \le \EE{\RPi[{\statPolicy[p^*]}]}$ for all $\policy \in \policiesSet$, i.e., $\statPolicy[p^*] = \optPolicy$.
	
	Moving on to the second part of the Theorem, notice that $\simplex$ is convex, compact and its set of extreme points is also compact (discrete). So, returning to \cref{eq:simpleOracleProof01} and using the quasiconvexity of $\RHat$, we notice that a maximizer is attained at an extreme point of $\simplex$. Formally, there exists $i^* \in \actionSet$ such that
	\begin{align*}
		\RHatFunc{\Fp[e_{i^*}]} = \max_{p \in \simplex} \RHatFunc{\Fp},
	\end{align*}
	where $\seqDef{e_i}{i=1}{K}$ are the standard unit vectors in $\RR[K]$. Continuing as before we conclude that $\statPolicy[e_{i^*}] = \optPolicy$ as desired. 
\end{proof}

\section{Proofs of Section~\ref{sec:learningAlgorithm}} \label{appendix:proxyRegret}
	
\begin{proof}[of \cref{lemma:pseudoRegretDecomp}]
We begin by proving the Lipschitz property.
Using the local modulus of continuity assumed by stability we get that for any $F_1, F_2 \in \DistSetDelta$
\begin{align*}
    \abs{\RHatFunc{F_1} - \RHatFunc{F_2}}
    &\le 
    \polyContModCoeff \brk*{\norm{F_1 - F_2}
    +
    \norm{F_1 - F_2}^{\polyContModDeg}}
    \\
    &=
    \polyContModCoeff \brk*{1 + \norm{F_1 - F_2}^{\polyContModDeg - 1}} \norm{F_1 - F_2}
    \\
    &\le 
    \polyContModCoeff \brk*{1 + \distSetDiameter^{\polyContModDeg - 1}} \norm{F_1 - F_2}
    \\
    &= 
    L \norm{F_1 - F_2},
\end{align*}
as desired.
Next, we show the pseudo regret decomposition. Using quasiconvexity as in the second part of \cref{theorem:StableEDRMoptPolicy}, there exists $i^* \in \actionSet$ such that $\FpOpt = \FiOpt$.
Using the Lipschitz constant $\lipConst$, and the triangle inequality we thus have that,
\begin{align*}
    \pseudoRegret
    &=
    \EEBrk{\RHatFunc{\FiOpt} - \RHatFunc{\FProxPi{T}}}
    \\
    &\le
    \lipConst \EE{\norm{\FiOpt - \FProxPi{T}}} \\
    &= 
    \lipConst \EE{\norm{\frac{1}{T}\sum_{i=1}^{K} \tauI\brk*{\FiOpt - \Fi}}}
    \\
    &\le
    \frac{\lipConst}{T} \EEBrk{\sum_{i=1}^{K} \tauI \norm{\FiOpt - \Fi}}
    \\
    &\le
    \frac{\lipConst \gapRatio}{T} \sum_{i \neq i^*} \Di \EE{\tauI}
    .
\end{align*}
\end{proof}

\begin{proof}[of \cref{theorem:UUCB}]
\rev{The proof uses standard techniques from the UCB literature, and \cite{bubeck2012regret} in particular.}{}
We begin with the following concentration result due to Requirement~\ref{item:StrongEDRMCond2} of stability.
\begin{align} \label{eq:UUCBproof4}
\begin{aligned}
	\PP{\abs{\RHatFunc{\FHatPi[(i)]{t}} - \RHatFunc{\Fi}} \ge x}
	&\le \PP{\polyContModCoeff \brk*{\norm{\FHatPi[(i)]{t} - \Fi} + \norm{\FHatPi[(i)]{t} - \Fi}^\polyContModDeg} \ge x} \\
	&\le \PP{\norm{\FHatPi[(i)]{t} - \Fi} \ge \frac{x}{2\polyContModCoeff}} + \PP{\norm{\FHatPi[(i)]{t} - \Fi} \ge \brk*{\frac{x}{2\polyContModCoeff}}^{1/\polyContModDeg}} \\
	&\le 2\exp \brk*{- \concentrationConst t \brk*{\frac{x}{2\polyContModCoeff}}^2} + 2\exp \brk*{- \concentrationConst t \brk*{\frac{x}{2\polyContModCoeff}}^{2/\polyContModDeg}} \\
	&\le 4 \exp \brk*{ -t \phiFuncAt{x}}.
\end{aligned}
\end{align}
Now, for all $i \in \actionSet$ and $1 \le t \le T$ denote the events
\begin{align*}
	&\ucbEvent{i}{t} = \brk[c]*{\RHatFunc{\FHatPi[(i)]{\tauIAt{t-1}}} > \RHatFunc{\Fi} + \phiInvFuncAt{\frac{\gamma \log t}{\tauIAt{t-1}}}} ,\\
	&\ucbEvent{*}{t} = \brk[c]*{\RHatFunc{\FHatPi[(i^*)]{\tauIAt[i^*]{t-1}}} + \phiInvFuncAt{\frac{\gamma \log t}{\tauIAt[i^*]{t-1}}} \le \RHatFunc{\FiOpt}},
\end{align*}
and their complements by $\ucbEventInv{i}{t}$, $\ucbEventInv{*}{t}$ respectively. Using the union bound and \cref{eq:UUCBproof4} we have that
\begin{align*}
\PP{\ucbEvent{i}{t}}
&= \PP{\RHatFunc{\FHatPi[(i)]{\tauIAt{t-1}}} - \phiInvFuncAt{\frac{\gamma \log t}{\tauIAt{t-1}}} > \RHatFunc{\Fi}} \\
&\le \PP{\max_{1 \le s \le t}\brk[c]*{\RHatFunc{\FHatPi[(i)]{s}} - \phiInvFuncAt{\frac{\gamma \log t}{s}}} > \RHatFunc{\Fi}} \\
&\le \sum_{s=1}^{t}\PP{\RHatFunc{\FHatPi[(i)]{s}} - \phiInvFuncAt{\frac{\gamma \log t}{s}} > \RHatFunc{\Fi}} \\
&\le \sum_{s=1}^{t}\PP{\abs{\RHatFunc{\FHatPi[(i)]{s}} - \RHatFunc{\Fi}} \ge \phiInvFuncAt{\frac{\gamma \log t}{s}}} \\
&\le \sum_{s=1}^{t} 4 \exp \brk*{-s \phiFuncAt{\phiInvFuncAt{\frac{\gamma \log t}{s}}}} \\
&\le \sum_{s=1}^{t} \frac{4}{t^\gamma} = \frac{4}{t^{\gamma - 1}}.
\end{align*}
The same holds for $\PP{\ucbEvent{*}{t}}$, and so we obtain
\begin{align} \label{eq:ucbEventBound}
	\PP{\ucbEvent{i}{t} \cup \ucbEvent{*}{t}} \le \PP{\ucbEvent{i}{t}} + \PP{\ucbEvent{*}{t}} \le \frac{8}{t^{\gamma - 1}}.
\end{align}
Next, we denote $u = {\frac{\gamma \log T}{\phiFuncAt{\Di / 2}}}$, and show that
\begin{align} \label{eq:ucbEventSubset}
	\brk[c]*{\policyAt[\UUCB]{t} = i} \cap \brk[c]*{\tauIAt{t-1} \ge u} \subseteq \ucbEvent{i}{t} \cup \ucbEvent{*}{t}.
\end{align}
Indeed, assume in contradiction that $\brk[c]*{\policyAt[\UUCB]{t} = i} \cap \brk[c]*{\tauIAt{t-1} \ge u} \cap \ucbEventInv{i}{t} \cap \ucbEventInv{*}{t} \neq \emptyset$, then noticing that $\brk[c]*{\tauIAt{t-1} \ge u}$ implies $\brk[c]*{\Di \ge 2\phiInvFuncAt{\frac{\gamma \log t}{\tauIAt{t-1}}}}$ we have:
\begin{align*}
\RHatFunc{\FHatPi[(i^*)]{\tauIAt[i^*]{t-1}}} + \phiInvFuncAt{\frac{\gamma \log t}{\tauIAt[i^*]{t-1}}}
&> \RHatFunc{\FiOpt} \\
&= \RHatFunc{\Fi} + \Di \\
&\ge \RHatFunc{\Fi} + 2\phiInvFuncAt{\frac{\gamma \log t}{\tauIAt{t-1}}} \\
&\ge \RHatFunc{\FHatPi[(i)]{\tauIAt{t-1}}} + \phiInvFuncAt{\frac{\gamma \log t}{\tauIAt{t-1}}},
\end{align*}
which implies that $\policyAt[\UUCB]{T} \neq i$, thus contradicting our assumption.
Finally, denoting
\begin{align*}
	t_0 = \max_{1 \le t \le T}\brk[c]*{t ~\Big|~ \tauIAt{t-1} \le \max\brk[c]*{u,1}},
\end{align*}
and using \cref{eq:ucbEventBound,eq:ucbEventSubset} we have that
\begin{align*}
	\EE{\tauI}
	&= \EEBrk{\sum_{t=1}^{T}\indEvent{\policyAt[\UUCB]{t} = i}} \\
	&= \EEBrk{\sum_{t=1}^{t_0}\indEvent{\policyAt[\UUCB]{t} = i} + \sum_{t=t_0+1}^{T}\indEvent{\policyAt[\UUCB]{t} = i}} \\
	&= \EEBrk{\tauIAt{t_0} + \sum_{t=t_0+1}^{T}\indEvent{\policyAt[\UUCB]{t} = i \bigcap \tauIAt{t-1} \ge u}}  \\
	&\le \EEBrk{u+1 + \sum_{t=K+1}^{T}\indEvent{\policyAt[\UUCB]{t} = i \bigcap \tauIAt{t-1} \ge u}}  \\
	&= u+1 + \sum_{t=K+1}^{T}\PP{\policyAt[\UUCB]{t} = i \bigcap \tauIAt{t-1} \ge u}  \\
	&\le u+1 + \sum_{t=K+1}^{T}\PP{\ucbEvent{i}{t} \cup \ucbEvent{*}{t}}  \\
	&\le u+1 + \sum_{t=K+1}^{T}\frac{8}{t^{\gamma - 1}}
	\le u+1 + \int_{K}^{\infty} \frac{8}{t^{\gamma - 1}} dt
	\le u+1 + \frac{8}{\brk*{\gamma - 2}K^{\gamma - 2}}
	\le u+ \frac{\gamma + 6}{\gamma - 2}.
\end{align*}
Combining this with the expression for the pseudo regret given in \cref{lemma:pseudoRegretDecomp} we obtain the desired.
\end{proof}

\section{Proofs of Section~\ref{sec:horizonGap}} \label{appendix:horizonGapProofs}

We first need the following technical lemma whose proof may be found in \cref{sec:horizonGapSideLemmas}.
\begin{lemma}
\label{lemma:EmpiricalDistExpectation}
	Suppose that Requirement~\ref{item:StrongEDRMCond2} of stability holds. Then for any integer $\varDeg \ge 1$, policy $\policy \in \policiesSet$, and $K, T$ such that $\log KT \ge 3$, we have that:
	\begin{enumerate}
	    \item
	    $
	        \EE{\norm{\FHatPi{T} - \FProxPi{T}}^{\varDeg}}
	        \le
	        \brk[s]{
	            1 + \varDeg m!
	        }
	        \brk*{
	            \frac{K^2 \log KT}{\concentrationConst T}
	        }^{\varDeg/2}
	        ,
	    $
	    where 
	    $m = \ceil{\frac{\varDeg}{2} - 1}$;
	    \item 
	    $
	        \EE{\norm{\FHatPi{T} - \FProxPi{T}}^{\varDeg}}
	        \le
	        2 \brk*{
	            \frac{K^2 \log KT}{\concentrationConst T}
	        }^{1/2}
	    $
	    for all $T \ge \frac{4 \varDeg K^2 \log KT}{\concentrationConst}$;
	    \item 
	    $
	        \EEBrk{
	            \norm{\FHatPi{T} - \FProxPi{T}}^{\varDeg}
	            \indEvent[1]{\norm{\FHatPi{T} - \FProxPi{T}} > M}
            }
	        \le
	        \frac{1}{T^2}
	    $
	    for all 
	    $
	        T
	        \ge
	        \frac{4 \varDeg K^2 \log KT}{\concentrationConst}
	        ,
	        M^2
	        \ge
	        \frac{4 \varDeg K^2 \log KT}{\concentrationConst T}
	        ;
        $
	\end{enumerate}
\end{lemma}

\begin{proof}[of \cref{prop:stableHorizonGap}]
We use Requirement~\ref{item:StrongEDRMCond1} of stability together with the second part of \cref{lemma:EmpiricalDistExpectation} to get that
\begin{align*}
    \abs2{\EEBrk{\RHatFunc{\FHatPi{T}} - \RHatFunc{\FProxPi{T}}}}
    &\le
    \EE{\abs{\RHatFunc{\FHatPi{T}} - \RHatFunc{\FProxPi{T}}}}
    \\
    &\le
    \polyContModCoeff \brk[s]*{
        \EE{\norm{\FHatPi{T} - \FProxPi{T}}}
        +
        \EE{\norm{\FHatPi{T} - \FProxPi{T}}^\polyContModDeg}
    }
    \\
    &\le
    4 \polyContModCoeff \brk*{
        \frac{K^2 \log KT}{\concentrationConst T}
    }^{1/2}
    .
\end{align*}
\end{proof}

\begin{proof}[of \cref{thm:smoothHorizonGap}]
First, notice that 
$
    \EE{\FHatPi{T}}
    =
    \EE{\FProxPi{T}}
$
for any $\policy \in \policiesSet$. This is easily seen as,
\begin{align*}
    \EE{\FHatPi{T}} 
    = 
    \EEBrk{\frac{1}{T}\sum_{t=1}^{T} \indFunc{\XtPi}}
    =
    \EEBrk{\frac{1}{T}\sum_{t=1}^{T} \EEBrk{\indFunc{\XtPi} \big| \policyAt{t}}}
    =
    \EEBrk{\frac{1}{T}\sum_{t=1}^{T} \Fi[\policyAt{t}]}
    =
    \EE{\FProxPi{T}}.
\end{align*}
Since $\DRHat$ is a linear operator, we conclude that
\begin{align*}
    \EEBrk{\DRHatAtOn
        {F}
        {(\FHatPi{T} - \FProxPi{T})}
    }
    =
    0
    ,
    \qquad
    \forall F \in \DistSetDelta.
\end{align*}
With this in mind, we have the following decomposition
\begin{align*}
    \abs*{\EEBrk{
        \RHatFunc{\FHatPi{T}} 
	    -
	    \RHatFunc{\FProxPi{T}}
    }}
    &\le
    \EE{\abs2{
    \underbrace{
        \RHatFunc{\FHatPi{T}} 
	    -
	    \RHatFunc{{\FProxPi{T}}}
	    -
	    \DRHatAtOn
        {\FProxPi{T}}
        {(\FHatPi{T} - \FProxPi{T})}
    }_{\delta_1}
    }}
    \\
    &+
    \EE{\abs2{
    \underbrace{
        \brk1{
            \DRHatAt{\FProxPi{T}}
            -
            \DRHatAt{F}
        }
        \cdot
        \brk{\FHatPi{t} - \FProxPi{T}}
    }_{\delta_2}
    }}
    ,
\end{align*}
We bound $\EE{\abs{\delta_1}}, \EE{\abs{\delta_2}}$ to conclude the proof. 
For $\EE{\abs{\delta_1}}$, recalling the parameter $M_0$ in \cref{definition:smoothEDRM} (smoothness), we use smoothness together with the first part of \cref{lemma:EmpiricalDistExpectation} to get that
\begin{align*}
    \EEBrk{
        \abs{\delta_1} 
        \indEvent[1]{\norm{\FHatPi{T} - \FProxPi{T}} \le M_0}
    }
    \le
    \frac12 \pSmooth \EEBrk{\norm{\FHatPi{T} - \FProxPi{T}}^2}
    \le
    \frac{3\pSmooth K^2 \log KT}{2 \concentrationConst T}
    .
\end{align*}
Next, we use stability (modulus of continuity) together with part 3 of \cref{lemma:EmpiricalDistExpectation}, which holds due to our assumption on $M_0$, to get that
\begin{align*}
    \EEBrk{
        \abs{\delta_1} 
        \indEvent[1]{\norm{\FHatPi{T} - \FProxPi{T}} > M_0}
    }
    &\le
    \EEBrk{
        \indEvent[1]{\norm{\FHatPi{T} - \FProxPi{T}} > M_0}
        \brk*{
        \contModFunc{\norm{\FHatPi{T} - \FProxPi{T}}}
        +
        2\polyContModCoeff \norm{\FHatPi{T} - \FProxPi{T}}
        }
    }
    \\
    &=
    \polyContModCoeff \EEBrk{
        \indEvent[1]{\norm{\FHatPi{T} - \FProxPi{T}} > M_0}
        \brk*{
        \norm{\FHatPi{T} - \FProxPi{T}}^{\polyContModDeg}
        +
        3\norm{\FHatPi{T} - \FProxPi{T}}
        }
    }
    \\
    &\le
    \frac{4 \polyContModCoeff \indEvent{M_0 < \infty}}{T^2}
    ,
\end{align*}
where the first step also used the modulus of continuity to bound the Gateaux derivative of $\RHat$. Summing the two inequalities bounds $\EE{\abs{\delta_1}}$.

Finally, to bound $\EE{\abs{\delta_2}}$ we first use the Cauchy–Schwarz inequality together with smoothness (\cref{definition:smoothEDRM}) to get that
\begin{align*}
    \EE{\abs{\delta_2}}
    \le
    \EEBrk{
        \norm{
            \DRHatAt{\FProxPi{T}}
            -
            \DRHatAt{F}
        }
        \norm{\FHatPi{T} - \FProxPi{T}}
    }
    \le
    \pSmooth
    \EEBrk{
        \norm{\FProxPi{T} - F}
        \norm{\FHatPi{T} - \FProxPi{T}}
    }
    .
\end{align*}
Next, let
$
    \MT^2
    =
    \frac{4 K^2 \log KT}{\concentrationConst T}
$
to get that
\begin{align*}
    \EEBrk{
        \abs{\delta_2}
        \indEvent[1]{\norm{\FHatPi{T} - \FProxPi{T}} \le \MT}
    }
    \le
    \pSmooth \MT \EE{\norm{\FProxPi{T} - F}}
    .
\end{align*}
On the other hand, recalling that $\distSetDiameter$ is the diameter of $\DistSetDelta$, we use part 3 of \cref{lemma:EmpiricalDistExpectation} to get that
\begin{align*}
    \EEBrk{
        \abs{\delta_2}
        \indEvent[1]{\norm{\FHatPi{T} - \FProxPi{T}} > \MT}
    }
    \le
    \pSmooth \distSetDiameter
    \EEBrk{
        \norm{\FHatPi{T} - \FProxPi{T}}
        \indEvent[1]{\norm{\FHatPi{T} - \FProxPi{T}} > \MT}
    }
    \le
    \frac{\pSmooth \distSetDiameter}{T^2}
    .
\end{align*}
Summing the two inequalities bounds $\EE{\abs{\delta_2}}$, and concludes the proof.
\end{proof}

\subsection{Technical Side Lemmas}
\label{sec:horizonGapSideLemmas}
\begin{proof}[of \cref{lemma:empDistConcentration}]
We start by using the triangle inequality and the union bound to get,
\begin{align*}
    \PP{\norm{\FHatPi{T} - \FProxPi{T}} > x}
    &=
    \PP{\norm{\frac{1}{T}\sum_{i=1}^{K} \tauI \brk*{\FHatPi[(i)]{\tauI} - \Fi}} > x} 
    \\
    &\le 
    \PP{\frac{1}{T}\sum_{i=1}^{K} \tauI \norm{\FHatPi[(i)]{\tauI} - \Fi} > x} 
    \\
    &\le
    \sum_{i=1}^{K}
    \PP{\tauI \norm{\FHatPi[(i)]{\tauI} - \Fi} > \frac{T}{K}x}.
\end{align*}
Now notice that $0 \le \tauI \le T$. So we have that,
\begin{align*}
    \tauI \norm{\FHatPi[(i)]{\tauI} - \Fi} 
    \le
    \max_{1 \le s \le T} s \norm{\FHatPi[(i)]{s} - \Fi}
    ,
\end{align*}
where the case of $\tauI = s = 0$ is dropped as it is clearly not the maximizer. Using this expression together with the union bound we get that,
\begin{align*}
    \PP{\norm{\FHatPi{T} - \FProxPi{T}} > x}
    &\le 
    \sum_{i=1}^{K} \PP{\max_{1 \le s \le T} s  \norm{\FHatPi[(i)]{s} - \Fi} > \frac{T}{K}x}
    \\
    &\le
    \sum_{i=1}^{K}\sum_{s=1}^{T} \PP{\norm{\FHatPi[(i)]{s} - \Fi} > \frac{Tx}{sK}}.
\end{align*}
Applying Requirement~\ref{item:StrongEDRMCond2} of stability (concentration), we have that
\begin{align*}
\begin{aligned}
    \PP{\norm{\FHatPi{T} - \FProxPi{T}} > x}
    &\le
    \sum_{i=1}^{K}\sum_{s=1}^{T} 2 \exp \brk*{-\concentrationConst s \brk*{\frac{Tx}{sK}}^2 } 
    \\
    &=
    2K \sum_{s=1}^{T} \exp \brk*{-\concentrationConst \frac{T^2x^2}{sK^2}} 
    \\
    &\le
    2KT \exp \brk*{-\concentrationConst \frac{Tx^2}{K^2}}
    ,
\end{aligned}
\end{align*}
where in the last step we use the fact that $s=T$ maximizes the summands.
\end{proof}

\begin{proof}[of \cref{lemma:EmpiricalDistExpectation}]
For the first claim, recall that
$
    m
    =
    \ceil{\frac{\varDeg}{2} - 1}
    ,
$
and let $x_0 \ge 0$ be a constant to be determined later. We begin by using the tail sum formula and exchanging variables to get that
\begin{align*}
    \EE{
        \norm{\FHatPi{T} - \FProxPi{T}}^\varDeg
    }
    &= 
    \int_{0}^{\infty} 
        \PP{\norm{\FHatPi{T} - \FProxPi{T}}^\varDeg > x} 
    dx 
    \\
    &\le
    x_0^\varDeg 
    +
    \int_{x_0^\varDeg}^{\infty}
        \PP{\norm{\FHatPi{T} - \FProxPi{T}}^\varDeg > x} 
    dx
    \\
    \tag{$x = (x_0 x')^\varDeg$}
    &=
    x_0^\varDeg
    +
    \varDeg x_0^{\varDeg} \int_{1}^{\infty}
        x^{\varDeg - 1} \PP{\norm{\FHatPi{T} - \FProxPi{T}} > x_0 x}
    dx
    \\
    &\le
    x_0^\varDeg \brk[s]*{
        1
        +
        \varDeg \int_{1}^{\infty}
            x^{2m+1} \PP{\norm{\FHatPi{T} - \FProxPi{T}} > x_0 x}
        dx
    }
    ,
\end{align*}
where the last transition used the fact that
$
    2m+1
    \ge
    \varDeg - 1
    .
$
Next, we use the tail bound in \cref{lemma:empDistConcentration} to get that
\begin{align*}
    \EE{
        \norm{\FHatPi{T} - \FProxPi{T}}^\varDeg
    }
    \le
    x_0^\varDeg \brk[s]*{
        1
        +
        2 \varDeg K T \int_{1}^{\infty}
            x^{2m+1} \exp\brk*{-\frac{\concentrationConst T x_0^2 x^2}{K^2}}
        dx
        .
    }
\end{align*}
Now, choose 
$
    x_0^2
    =
    \frac{K^2 \log KT}{\concentrationConst T}
    ,
$
and using \cref{lemma:knownIntegral} with $a = \log KT \ge 3$ to solve this known integral, we get that
\begin{align*}
    \EE{
        \norm{\FHatPi{T} - \FProxPi{T}}^\varDeg
    }
    \le
    \brk[s]*{
        1
        +
        \varDeg m!
    }
    \brk*{\frac{K^2 \log KT}{\concentrationConst T}}^{\varDeg/2}
    .
\end{align*}
This holds for all $\policy \in \policiesSet$ thus concluding the proof of the first claim.

Next, we prove the second claim by showing that, under the assumption on $T$, the first claim may be bounded by the desired term. To see this notice that
\begin{align*}
    \frac{
    \brk[s]*{
        1
        +
        \varDeg m!
    }}
    {2}
    \brk*{\frac{K^2 \log KT}{\concentrationConst T}}^{\frac{\varDeg-1}{2}}
    &\le
    \varDeg^{\varDeg/2}
    \brk*{\frac{K^2 \log KT}{\concentrationConst T}}^{\frac{\varDeg-1}{2}}
    \\
    \tag{$T \ge \frac{4 \varDeg K^2 \log KT}{\concentrationConst}$}
    &\le
    \frac{d^{1/2}}{4^{(d-1)/2}}
    \\
    &=
    \exp{\brk*{
        \frac12 \log \varDeg
        -
        \frac{\varDeg - 1}{2} \log 4
    }}
    \\
    \tag{$\log x \le x - 1$}
    &\le
    \exp{\brk*{
        \frac{\varDeg - 1}{2}
        -
        \frac{\varDeg - 1}{2} \log 4
    }}
    \le
    1
    .
\end{align*}
Changing sides gives the desired bound, and concludes the proof of the second claim.

Finally, for the third claim, we begin by repeating the first steps of the first claim to get that
\begin{align*}
    \EEBrk{
        \indEvent[1]{\norm{\FHatPi{T} - \FProxPi{T}} > M}
        \norm{\FHatPi{T} - \FProxPi{T}}^\varDeg
    }
    &\le
    \EEBrk{
        \indEvent[1]{\norm{\FHatPi{T} - \FProxPi{T}} > x_1}
        \norm{\FHatPi{T} - \FProxPi{T}}^\varDeg
    }
    \\
    &\le
    2 \varDeg x_1^\varDeg K T \int_{1}^{\infty}
        x^{2m+1} \exp\brk*{-\frac{\concentrationConst T x_1^2 x^2}{K^2}}
    dx
    ,
\end{align*}
where we choose
$
    x_1^2
    =
    \frac{4 \varDeg K^2 \log KT}{\concentrationConst T}
    \le
    M
    .
$
Notice that our choice of $T$ ensures that $x_1 \le 1$ and thus applying \cref{lemma:knownIntegral} with 
$a = 4 \varDeg \log KT \ge 3,$
we get that
\begin{align*}
    \EEBrk{
        \indEvent[1]{\norm{\FHatPi{T} - \FProxPi{T}} > M}
        \norm{\FHatPi{T} - \FProxPi{T}}^\varDeg
    }
    &\le
    \frac{\varDeg m!}{(K T)^{4\varDeg-1}}
    \le
    \frac{1}{T^2}
    ,
\end{align*}
where the last step also used the fact that $T \ge d$.
\end{proof}

\begin{lemma}
\label{lemma:knownIntegral}
    For any real $a > 0$ and integer $m \ge 0$ we have that
    \begin{align*}
        \int_{1}^{\infty}
            x^{2m+1} \exp{(-ax^2)}
        dx
        =
        \frac{\exp{(-a)}}{2 a^{m+1}} \sum_{j=0}^{m} \frac{m!}{j!} a^j
        .
    \end{align*}
    If $a \ge 3$ then we also have that
    \begin{align*}
        \int_{1}^{\infty}
            x^{2m+1} \exp{(-ax^2)}
        dx
        \le
        m! \exp{(-a)} / 2
        .
    \end{align*}
\end{lemma}
\begin{proof}
    Denote
    $
        f(m)
        =
        \int_{1}^{\infty}
            x^{2m+1} \exp{(-ax^2)}
        dx
        ,
    $
    and use integration by parts to get that
    \begin{align*}
        f(m)
        =
        \frac{m}{a} f(m-1)
        +
        \frac{\exp{(-a)}}{2a}
        .
    \end{align*}
    Now, plugging $m = 0$ we get that
    $
        f(0)
        =
        \frac{\exp{(-a)}}{2a}
        .
    $
    Finally, it is trivial to verify that the suggested solution satisfies the difference equation as well as the initial condition, thus concluding the first part of the proof.
    For the second part we use the assumption on $a \ge 3$ to upper bound the expression as
    \begin{align*}
        f(m)
        \le
        \frac{ m! \exp{(-a)}}{2 a} \sum_{j=0}^{m} \frac{1}{j!}
        \le
        \frac{ m! e \exp{(-a)}}{2 a}
        \le
        \frac{ m! \exp{(-a)}}{2}
    \end{align*}
\end{proof}

\section{Proofs of Section~\ref{sec:regretBound}}
\label{appendix:proofsOfregretBound}
\begin{proof}[of \cref{prop:linearGap}]
Recall that for any $F \in \DistSetDelta$ there exists $p \in \simplex$ such that $F = \sum_{i=1}^{K} p_i \Fi$. Now, we use convexity to conclude that
\begin{align*}
    \RHatFunc{\sum_{i=1}^{K} p_i \Fi}
    -
    \RHatFunc{\FiOpt}
    &\le
    \sum_{i=1}^{K} p_i \brk*{
        \RHatFunc{\Fi}
        -
        \RHatFunc{\FiOpt}
    }
    \\
    &=
    - \sum_{i=1}^{K} p_i \Di
    \le
    -\frac{1}{\gapRatio} \sum_{i=1}^{K} p_i \norm{\Fi - \FiOpt}
    =
    -\frac{1}{\gapRatio} \norm{F - \FiOpt}
    ,
\end{align*}
where $\gapRatio$, which is defined in \cref{eq:diameterAndGapRatio}, is finite since $\Di > 0$ for all $i \neq i^*$.
\end{proof}

\begin{proof}[of \cref{thm:regretBound}]
We begin by stating the explicit condition on the time horizon $T$. Letting
$
    \MT^2
    =
    \frac{4 K^2 \log KT}{\concentrationConst T}
    ,
$
we require that $T$ is large enough such that
\begin{align}
\label{eq:Tconditions}
    \frac{\gapRatio}{T} 
	\sum_{i \neq i^*} \brk*{
	    \frac{\gamma \Di \log T}{\phiFuncAt{\Di / 2}}
	    +
	    \frac{\gamma + 6}{\gamma - 2} \Di
    }
    \le
    \MT
    \le
    \min\brk[c]*{
    \frac{1}{\sqrt{\polyContModDeg}}
    ,
    \frac{M_0}{\sqrt{\polyContModDeg}}
    ,
    \frac{1}{\pLinGap \pSmooth}
    }
    ,
\end{align}
which is indeed polynomial in the problem parameters. The first two terms in the minimum are the basic requirements of \cref{thm:smoothHorizonGap}, and the third term was chosen such that applying \cref{thm:smoothHorizonGap} with $F = \FiOpt$, we get that
\begin{align*}
    J_1(T)
    =
    \EEBrk{
        \RHatFunc{\FHatPi[\optPolicyT]{T}}
        -
        \RHatFunc{\FProxPi[\optPolicyT]{T}}
    }
    \le
    \frac{2\pSmooth K^2 \log KT}{\concentrationConst T}
	+
	\frac{1}{\pLinGap}\EE{\norm{\FProxPi[\optPolicyT]{T} - \FiOpt}}
	+
	\frac{
	\pSmooth \distSetDiameter
	+
	4\polyContModCoeff \indEvent{M_0 < \infty}}{T^2}
	.
\end{align*}
Next, notice that the first step in the decomposition of the pseudo regret, which is given in \cref{lemma:pseudoRegretDecomp}, is
$
    \pseudoRegret
    \le
    \lipConst \EE{\norm{
        \FProxPi[\UUCB]{T}
        -
        \FiOpt
    }}
    ,
$
and thus the bound in \cref{theorem:UUCB} together with the left hand side of \cref{eq:Tconditions} imply that
$
    \EE{\norm{
        \FProxPi[\UUCB]{T}
        -
        \FiOpt
    }}
    \le
    \MT
    .
$
Applying \cref{thm:smoothHorizonGap} with $F = \FiOpt$ we obtain that
\begin{align*}
    J_3(T)
    =
    \EEBrk{
        \RHatFunc{\FProxPi[\UUCB]{T}}
        -
        \RHatFunc{\FHatPi[\UUCB]{T}}
    }
    \le
    \underbrace{
        \frac{6\pSmooth K^2 \log KT}{\concentrationConst T}
    }_{3\pSmooth\MT^2 / 2}
	+
	\frac{
	\pSmooth \distSetDiameter
	+
	4\polyContModCoeff \indEvent{M_0 < \infty}}{T^2}
	.
\end{align*}
Next, using the linear gap assumption we get that
\begin{align*}
    J_2(T)
    =
    \EEBrk{
        \RHatFunc{\FProxPi[\optPolicyT]{T}}
        -
        \RHatFunc{\FiOpt}
    }
    \le
    -\frac{1}{\pLinGap}\EE{\norm{\FProxPi[\optPolicyT]{T} - \FiOpt}}
    .
\end{align*}
Finally, recall that in \cref{eq:regretDecomp} we decompose the regret as
$
    \regret
    =
    \pseudoRegret
    +
    J_1(T)
    +
    J_2(T)
    +
    J_3(T)
    .
$
Combining the above and using \cref{theorem:UUCB} to bound $\pseudoRegret[\UUCB]$ concludes the proof.
\end{proof}

\ifdefined\morStyle
    \newpage
    \section*{Online Companion}
\fi

\section{Details of Section~\ref{sec:examples}} 
\label{appendix:examplesDetails}

In this section we provide the missing details from \cref{sec:examples}. For the most part, our goal is to verify stability and smoothness, which are the conditions for \cref{thm:regretBound}. We note that \cref{theorem:StableEDRMoptPolicy} will typically hold as long as $\abs{\RHatFunc{\Fi}} < \infty$. The exception to this rule is $\VAR$, for which we will require an additional assumption for this to hold. However, we also show that a single arm infinite horizon oracle always exists, even without this assumption.

In what follow we continue to operate under Assumption~\ref{assumption:boundedReward}, which bounds the rewards. This is mostly to make the exposition more concise, and we state explicitly the places where it is indeed necessary.

\subsection{Linear \EDRMabbrv s}
\label{appendix:linear}
We expand on the application of Hoeffding's inequality for the stability of linear \EDRMabbrv s, and note that this could easily be replaced by a sub-Gaussian type assumption. Recall that for a linear \EDRMabbrv\space $\RHat[\text{lin}]$, we use the seminorm 
$\norm{F} = \abs{\RHatFunc[\text{lin}]{F}}$. We thus have that
\begin{align*}
    \norm{\FHatPi[(i)]{t} - \Fi}
    =
    \abs{\RHatFunc[\text{lin}]{\FHatPi[(i)]{t} - \Fi}}
    =
    \abs*{\frac{1}{t} \sum_{s=1}^{t}
        \RHatFunc[\text{lin}]{\indFunc{\Xti{s}}}
        -
        \RHatFunc[\text{lin}]{\Fi}
    }
    ,
\end{align*}
where the last transition used the definition of the empirical distribution in \cref{eq:FhatiAbbrv}, and the linearity of $\RHat[\text{lin}]$. Notice that the linearity of $\RHat[\text{lin}]$ also implies that
$
    \RHatFunc[\text{lin}]{\Fi}
    =
    \EE{\RHatFunc[\text{lin}]{\indFunc{\Xti{s}}}}
    .
$
We now have a sum of zero mean $i.i.d$ random variables that take values in an interval of squared length
\begin{align*}
    \vartheta_{\text{lin}}
    =
    \max_{x,y \in \brk[s]{0,1}}
    \brk[s]*{
        \RHatFunc[\text{lin}]{\indFunc{x}}
        -
        \RHatFunc[\text{lin}]{\indFunc{y}}
    }^2
    ,
\end{align*}
and thus invoking Hoeffding's inequality we get that $\RHat[\text{lin}]$ satisfies Requirement~\ref{item:StrongEDRMCond2} of stability with $\concentrationConst = 2 / \vartheta_{\text{lin}}$.

\subsection{Composite \EDRMabbrv s}
\label{appendix:composite}

Recall that an \EDRMabbrv\space is composite if there exist $\RHat[(1)], \ldots \RHat[(n)]$ and $\hComp : \RR[n] \to \RR$ such that
\begin{align*}
    \RHatFunc[\hComp]{F}
    =
    \hCompFunc{\RHatFunc[(1)]{F}, \ldots, \RHatFunc[(n)]{F}}
    .
\end{align*}
For a set $S \subseteq \DistSet$ let
\begin{align*}
    \RHatFunc[\hComp]{S}
    =
    \brk[c]*{
        \brk1{
            \RHatFunc[(1)]{F}, \ldots, \RHatFunc[(n)]{F}
        } \in \RR[n]
        \Big|\;
        \forall F \in S
    }
    .
\end{align*}
be its image under the linear mappings that compose $\RHat[\hComp]$. \cref{lemma:compositeInformal} is made formal in the following result.

\begin{lemma}[\textbf{Composite \EDRMabbrv}] \label{lemma:compositeEDRM}
    Suppose $\RHat[(1)], \ldots, \RHat[(n)]$ are linear, and stable with parameter $\concentrationConst_0$. Then:
	\begin{enumerate}
		\item If $h$ admits a polynomial local modulus of continuity, i.e., there exist $\polyContModCoeff > 0, \polyContModDeg \ge 1$ such that
		\begin{align*}
			\abs{\hCompFunc{x} - \hCompFunc{y}}
			\le
			\polyContModCoeff \brk*{
			    \norm{x - y}_2 
			    +
			    \norm{x - y}_2^{\polyContModDeg}
		    }
		    \qquad
		    ,
		    \forall x \in \RHatFunc[\hComp]{\DistSetDelta}
		    ,
		    y \in \RHatFunc[\hComp]{\funcSpace}
		    ,
		\end{align*}
		then $\RHat[\hComp]$ is stable with the same $\polyContModCoeff, \polyContModDeg$ and $\concentrationConst = \frac{\log 2}{n \log 2n} \concentrationConst_0$;
		\item If $\hComp$ is locally smooth, i.e., there exist $\pSmooth \ge 0, M_0 > 0$ such that for any 
		$
		    x 
		    \in
		    \RHatFunc[\hComp]{\DistSetDelta}
		    ,
		    y 
		    \in
		    \RHatFunc[\hComp]{\funcSpace}
	    $
	    satisfying $\norm{x-y}_2 \le M_0$ we have that
		\begin{align*}
		    \abs{
		        \hCompFunc{y}
		        -
		        \hCompFunc{x}
		        -
		        \nabla \hCompFunc{x}^T (y - x)
	        }
	        \le
	        \frac{\pSmooth}{2} \norm{x - y}_2^2
	        ,
		\end{align*}
		then $\RHat[\hComp]$ is smooth with the same parameters;
		\item If $\hComp$ is convex then so is $\RHat[\hComp]$.
	\end{enumerate}
\end{lemma}

\begin{proof}
Recall that we consider $\RHat[\hComp]$ under the norm
\begin{align*}
    \norm{F}
    =
    \norm{\RHatFunc[(1)]{F}, \ldots, \RHatFunc[(n)]{F}}_2
    ,
\end{align*}
where $\norm{\cdot}_2$ is the $\ell^2$ norm on $\RR[n]$.
Starting with Requirement~\ref{item:StrongEDRMCond1} of stability, we use the modulus of continuity assumption on $\hComp$ to get that for all $F \in \DistSetDelta$ and $G \in \funcSpace$
\begin{align*}
    \abs{\RHatFunc[\hComp]{F} - \RHatFunc[\hComp]{G}}
    &=
    \abs{
        \hCompFunc{\RHatFunc[(1)]{F}, \ldots, \RHatFunc[(n)]{F}}
        -
        \hCompFunc{\RHatFunc[(1)]{G}, \ldots, \RHatFunc[(n)]{G}}
    }
    \\
    &\le
    \polyContModCoeff \brk{
        \norm{\RHatFunc[(1)]{F-G}, \ldots, \RHatFunc[(n)]{F-G}}_2
        +
        \norm{\RHatFunc[(1)]{F-G}, \ldots, \RHatFunc[(n)]{F-G}}_2^{\polyContModDeg}
    }
    \\
    &=
    \polyContModCoeff \brk{
        \norm{F-G} + \norm{F-G}^{\polyContModDeg}
    }
    .
\end{align*}
Next, for Requirement~\ref{item:StrongEDRMCond2} we use the stability of the linear \EDRMabbrv s to get that
\begin{align*}
    \PP{\norm{\FHatPi[(i)]{t} - \Fi} > x}
    &=
    \PP{\norm{
        \RHatFunc[(1)]{\FHatPi[(i)]{t} - \Fi}
        ,
        \ldots
        ,
        \RHatFunc[(n)]{\FHatPi[(i)]{t} - \Fi}
    }_2 > x}
    \\
    \tag{union bound}
    &\le
    \min\brk[c]*{
    1,
    \sum_{j=1}^{n} \PP{\abs{\RHatFunc[(j)]{\FHatPi[(i)]{t} - \Fi}} > \frac{x}{\sqrt{n}}}
    }
    \\
    &\le
    \min\brk[c]*{
    1,
    2n \exp\brk{-\frac{\concentrationConst_0 t x^2}{n}}
    }
    \\
    &\le
    2 \exp\brk{-\frac{\log 2 \concentrationConst_0}{n \log 2n} t x^2},
\end{align*}
thus concluding stability of $\RHat[\hComp]$, which is the first claim.

Now, moving on to smoothness, we use to chain rule to get that for any $F \in \DistSet$ and $G \in \funcSpace$
\begin{align*}
    \DRHatAtOn{F}{G}
    =
    \partial \hCompFunc{\RHatFunc[(1)]{F}, \ldots, \RHatFunc[(n)]{F}} \cdot G
    =
    \nabla \hCompFunc{\RHatFunc[(1)]{F}, \ldots, \RHatFunc[(n)]{F}}^T
    \brk{\RHatFunc[(1)]{G}, \ldots, \RHatFunc[(n)]{G}}
    ,
\end{align*}
and applying the assumed smoothness of $\hComp$ we get that if
$
    \norm{F-G} \le M_0
$
then
\begin{align*}
    &\abs{
        \RHatFunc[\hComp]{G}
        -
        \RHatFunc[\hComp]{F}
        -
        \DRHatAtOn{F}{(G - F)}
    }
    \\
    &\quad=
    \left|
        \hCompFunc{\RHatFunc[(1)]{G}, \ldots, \RHatFunc[(n)]{G}}
        -
        \hCompFunc{\RHatFunc[(1)]{F}, \ldots, \RHatFunc[(n)]{F}}
    \right.
    \\
    &\qquad
    \left.
        -
        \nabla \hCompFunc{\RHatFunc[(1)]{F}, \ldots, \RHatFunc[(n)]{F}}^T
        \brk{\RHatFunc[(1)]{G-F}, \ldots, \RHatFunc[(n)]{G-F}}
    \right|
    \\
    &\quad\le
    \frac12 \pSmooth \norm{\RHatFunc[(1)]{G-F}, \ldots, \RHatFunc[(n)]{G-F}}_2^2
    \\
    &\quad=
    \frac12 \pSmooth \norm{F-G}^2,
\end{align*}
thus concluding the smoothness of $\RHat[\hComp]$, which is the second claim.

Finally, if $\hComp$ is convex the $\RHat[\hComp]$ is a linear variable on a convex function and as such convex.
\end{proof}

\subsubsection{Entropic risk}
This is the only example where Assumption~\ref{assumption:boundedReward} is indeed necessary for our framework. We note that this could be removed in the future by expanding our analysis to an exponential family of moduli of continuity.
Recall that in terms of \cref{lemma:compositeEDRM}, we have that $\hComp(x) = -\frac{1}{\theta} \log x$, which is convex, and $x \in \brk[s]{\exp(-\theta), 1}$. Bounding the first derivative, we get that
\begin{align*}
    \frac{d \hComp}{dx}(x)
    =
    -\frac{1}{\theta x}
    \implies
    \abs*{\frac{d \hComp}{dx}(x)}
    \le
    \frac{1}{\theta}\exp(\theta),
\end{align*}
and thus $\hComp$ is Lipschitz with this constant and has a modulus of continuity with parameters 
$
    \polyContModCoeff
    =
    \frac{1}{2\theta}\exp(\theta)
    ,
    \polyContModDeg
    =
    1
    .
$
Next, recalling the second order charachterization of smoothness, we bound the second derivative, to get that
\begin{align*}
    \pSmooth
    \le
    \max_{x \in \brk[s]{\exp(-\theta), 1}}
        \abs*{\frac{d^2 \hComp}{dx^2}(x)}
    =
    \max_{x \in \brk[s]{\exp(-\theta), 1}} \frac{1}{\theta x^2}
    =
    \frac{1}{\theta}\exp(2\theta),
\end{align*}
thus proving the desired properties for \cref{lemma:compositeEDRM}.

\subsubsection{Variance}
Here $\hCompFunc{x_1, x_2} = x_1^2 - x_2$, which is convex. 
Next, for the modulus of continuity we have that
\begin{align*}
    \abs{\hCompFunc{x_1, x_2} - \hCompFunc{y_1, y_2}}
    &=
    \abs{(x_1 - y_1)(x_1 + y_1) + (y_2 - x_2)}
    \\
    &\le
    (x_1 - y_1)^2 
    +
    \abs{2x_1(x_1 - y_1) + (y_2 - x_2)}
    \\
    &\le
    \sqrt{1 + 4 x_1^2} \brk{
        \norm{x-y}_2^2
        +
        \norm{x-y}_2
    }
    .
\end{align*}
Since $(x_1, x_2) \in \RHatFunc[\hComp]{\DistSetDelta}$, we can bound
$
    \abs{x_1}
    \le
    \max_{i \in \actionSet} \abs{\RHatFunc[\text{ave}]{\Fi}}
    .
$
Since the reward is also bounded in $\brk[s]{0,1}$ we further have that $\abs{x_1} \le 1$, giving us the constant $\polyContModCoeff = \sqrt{5}$.
Finally, for smoothness we may bound the hessian as,
\begin{align*}
    \pSmooth
    =
    \max_{x_1,x_2 \in \RR} \norm{\nabla^2\hCompFunc{x_1,x_2}}
    =
    \norm*{
        \begin{pmatrix}
    		2 & 0 \\
    		0 & 0
		\end{pmatrix}
    }
    =
    2
    .
\end{align*}

\subsubsection{Mean-variance (Markowitz)}
Here we have that for $\rho \ge 0$ $\hComp(x, y) = x + \rho (x^2 - y)$, which is convex.
Next, for the modulus of continuity we have that
\begin{align*}
    \abs{\hCompFunc{x_1, x_2} - \hCompFunc{y_1, y_2}}
    &=
    \abs{\rho(x_1 - y_1)(x_1 + y_1) + \rho(y_2 - x_2) + (x_1 - y_1)}
    \\
    &\le
    \rho(x_1 - y_1)^2 
    +
    \abs{(2\rho x_1 + 1)(x_1 - y_1) + (y_2 - x_2)}
    \\
    &\le
    \sqrt{1 + (2\rho\abs{x_1} + 1)^2} \brk{
        \norm{x-y}_2^2
        +
        \norm{x-y}_2
    }
    .
\end{align*}
Since $(x_1, x_2) \in \RHatFunc[\hComp]{\DistSetDelta}$, we can bound
$
    \abs{x_1}
    \le
    \max_{i \in \actionSet} \abs{\RHatFunc[\text{ave}]{\Fi}}
    .
$
Since the reward is also bounded in $\brk[s]{0,1}$ we further have that $\abs{x_1} \le 1$, giving us the constant $\polyContModCoeff = 2(1 + \rho)$.
Finally, for smoothness we may bound the hessian as,
\begin{align*}
    \pSmooth
    =
    \max_{x_1,x_2 \in \RR} \norm{\nabla^2\hCompFunc{x_1,x_2}}
    =
    \norm*{
        \begin{pmatrix}
    		2\rho & 0 \\
    		0 & 0
		\end{pmatrix}
    }
    =
    2\rho
    .
\end{align*}

\subsubsection{Sortino ratio}
Here we have that for $r \in \RR$ and $\varepsilon_0 > 0$
\begin{align*}
    \hCompFunc{x_1, x_2}
    =
    (x_1 - r) / \sqrt{\varepsilon_0 - x_2}
    .
\end{align*}
\textbf{Linear gap:}
Let $\lambda \in \simplex$ and $F = \sum_{i=1}^{K} \lambda_i \Fi$. We need to show that
\begin{align*}
    \RHatFunc[\text{So}]{\FiOpt}
    -
    \RHatFunc[\text{So}]{F}
    \ge
    \frac{1}{\pLinGap} \norm{\FiOpt - F}.
\end{align*}
Denote 
$x_i = \RHatFunc[\text{ave}]{\Fi}$,
$y_i = \RHatFunc[\text{TSV}]{\Fi}$,
and
$z_i = \sqrt{\varepsilon_0 - y_i}$
.
Notice that $z_i$ is concave in $x_i, y_i$ and so we have that
\begin{align*}
    \sqrt{
        \varepsilon_0
        -
        \sum_{i=1}^{K} \lambda_i y_i
    }
    \ge
    \sum_{i=1}^{K} \lambda_i z_i
    .
\end{align*}
We conclude that
\begin{align*}
    \RHatFunc[\text{So}]{\FiOpt}
    -
    \RHatFunc[\text{So}]{F}
    &=
    \hCompFunc{x_{i^*}, y_{i^*}}
    -
    \hCompFunc{\sum_{i=1}^{K} \lambda_i x_i, \sum_{i=1}^{K} \lambda_i y_i}
    \\
    &\ge
    \frac{x_{i^*} - r}{z_{i^*}}
    -
    \frac{\sum_{i=1}^{K} \lambda_i x_i - r}{\sum_{i=1}^{K} \lambda_i z_i}
    \\
    &=
    \frac{\sum_{i=1}^{K} \lambda_i \brk{
        z_i (x_{i^*} - r)
        -
        z_{i^*} (x_i - r)
    }}{z_{i^*} \sum_{j=1}^{K} \lambda_j z_j}
    \\
    &=
    \sum_{i=1}^{K} \lambda_i \frac{z_i}{\sum_{j=1}^{K} \lambda_i z_i} \brk{
        \frac{x_{i^*} - r}{z_{i^*}}
        -
        \frac{x_i - r}{z_i}
    }
    \ge
    \frac{z_{\text{min}}}{z_{\text{max}}} \sum_{i=1}^{K} \lambda_i \Di
    ,
\end{align*}
where
$
    z_{\text{min}} 
    = 
    \min_{i \neq i^*} z_i
$
and
$
    z_{\text{max}} 
    = 
    \max_{i \in \actionSet} z_i
    .
$
Using the gap ratio defined in \cref{eq:diameterAndGapRatio} we get a linear gap with $\pLinGap = \gapRatio z_{\text{max}} / z_{\text{min}}$.

\textbf{Stability:}
For $x \in \RHatFunc[\text{So}]{\DistSetDelta}$ and $y \in \RHatFunc[\text{So}]{\funcSpace}$ we have that
\begin{align*}
    \abs{
        \hCompFunc{x_1, x_2}
        -
        \hCompFunc{y_1, y_2}
    }
    &=
    \abs*{
        \frac{x_1 - r}{\sqrt{\varepsilon_0 - x_2}}
        -
        \frac{y_1 - r}{\sqrt{\varepsilon_0 - y_2}}
    }
    \\
    &\le
    \abs*{
        \frac{x_1 - r}{\sqrt{\varepsilon_0 - y_2}}
        -
        \frac{y_1 - r}{\sqrt{\varepsilon_0 - y_2}}
    }
    +
    \abs*{
        \frac{x_1 - r}{\sqrt{\varepsilon_0 - x_2}}
        -
        \frac{x_1 - r}{\sqrt{\varepsilon_0 - y_2}}
    }
    \\
    &\le
    \frac{\abs{x_1 - y_1}}{\varepsilon_0}
    +
    \abs{x_1 - r}
    \abs*{
        \frac{
            \sqrt{\varepsilon_0 - y_2}
            -
            \sqrt{\varepsilon_0 - x_2}
        }{\sqrt{\varepsilon_0 - x_2}\sqrt{\varepsilon_0 - y_2}}
    }
    \\
    &\le
    \frac{\abs{x_1 - y_1}}{\varepsilon_0}
    +
    \frac{\abs{x_1 - r}}{2\varepsilon_0^{3/2}}
    \abs*{y_2 - x_2}
    \\
    &\le
    \sqrt{
        \frac{1}{\varepsilon_0^2}
        +
        \frac{(x_1 - r)^2}{4\varepsilon_0^3}
    }\norm{x-y}_2
    \\
    &\le
    \frac{\abs{x_1 - r} + 2}{2\min\brk[c]{\varepsilon_0, \varepsilon_0^{3/2}}}\norm{x-y}_2
    .
\end{align*}
Since $(x_1, x_2) \in \RHatFunc[\text{So}]{\DistSetDelta}$, we can bound
$
    \abs{x_1}
    \le
    \max_{i \in \actionSet} \abs{\RHatFunc[\text{ave}]{\Fi}}
    .
$
Since the reward is also bounded in $\brk[s]{0,1}$ we further have that $\abs{x_1} \le 1$, giving us the constants
$
    \polyContModCoeff 
    =
    (\abs{r} + 2) 
    /
    4 \min\brk[c]{\varepsilon_0, \varepsilon_0^{3/2}}
$
and
$\polyContModDeg = 1$.

\textbf{Smoothness:}
First, we calculate the hessian to get that
\begin{align*}
    \nabla^2 \hCompFunc{x_1, x_2}
    =
	\begin{pmatrix}
    	0 
    	&
    	\frac{1}{2\brk*{\varepsilon_0 - x_2}^{3/2}}
    	\\
    	\frac{1}{2\brk*{\varepsilon_0 - x_2}^{3/2}}
    	&
    	\frac{3\brk*{x_1 - r}}{4\brk*{\varepsilon_0 - x_2}^{5/2}}
	\end{pmatrix}
	.
\end{align*}
Next, we upper bound its spectral norm to get that. Let $w \in \RR[2]$ be such that $\norm{w}_2 \le 1$. Then we have that
\begin{align*}
    \abs{w^T \nabla \hCompFunc{x_1, x_2} w}
    &=
    \abs*{
        \frac{w_1 w_2}{(\varepsilon_0 - x_2)^{3/2}}
        +
        w_2^2 \frac{3(x_1 - r)}{4(\varepsilon_0 - x_2)^{5/2}}
    }
    \\
    &\le
    \frac{1}{2(\varepsilon_0 - x_2)^{3/2}}
    +
    \frac{3\abs{x_1 - r}}{4(\varepsilon_0 - x_2)^{5/2}}
    \\
    &\le
    \frac{2\varepsilon_0 + \abs{x_1 - r}}{4 \varepsilon_0^{5/2}}
\end{align*}
Here we cannot take $M_0 = \infty$. However, for any $M_0 < \infty$ we can use the above bound to conclude that the smoothness assumption of \cref{lemma:compositeEDRM} holds with
\begin{align*}
    \pSmooth
    =
    \frac{2\varepsilon_0 + \distSetDiameter + M_0 + \abs{r}}{4 \varepsilon_0^{5/2}}
    ,
\end{align*}
where under the assumption that the reward is in $\brk[s]{0,1}$ we further have that $\distSetDiameter = 1$.

\subsubsection{Sharpe ratio}
Here we have that for $r \in \RR$ and $\varepsilon_0 > 0$
\begin{align*}
    \hCompFunc{x_1, x_2}
    =
    (x_1 - r) / \sqrt{\varepsilon_0 - x_1^2 + x_2}
    .
\end{align*}

\textbf{Linear gap:}
Let $\lambda \in \simplex$ and $F = \sum_{i=1}^{K} \lambda_i \Fi$. We need to show that
\begin{align*}
    \RHatFunc[\text{Sh}]{\FiOpt}
    -
    \RHatFunc[\text{Sh}]{F}
    \ge
    \frac{1}{\pLinGap} \norm{\FiOpt - F}.
\end{align*}
Denote 
$x_i = \RHatFunc[\text{ave}]{\Fi}$,
$y_i = \RHatFunc[\text{sqr}]{\Fi}$,
and
$z_i = \sqrt{\varepsilon_0 - x_i^2 + y_i}$
.
Notice that $z_i$ is concave in $x_i, y_i$ and so we have that
\begin{align*}
    \sqrt{
        \varepsilon_0
        -
        \brk{\sum_{i=1}^{K} \lambda_i x_i}^2
        +
        \sum_{i=1}^{K} \lambda_i y_i
    }
    \ge
    \sum_{i=1}^{K} \lambda_i z_i
    .
\end{align*}
We conclude that
\begin{align*}
    \RHatFunc[\text{Sh}]{\FiOpt}
    -
    \RHatFunc[\text{Sh}]{F}
    &=
    \hCompFunc{x_{i^*}, y_{i^*}}
    -
    \hCompFunc{\sum_{i=1}^{K} \lambda_i x_i, \sum_{i=1}^{K} \lambda_i y_i}
    \\
    &\ge
    \frac{x_{i^*} - r}{z_{i^*}}
    -
    \frac{\sum_{i=1}^{K} \lambda_i x_i - r}{\sum_{i=1}^{K} \lambda_i z_i}
    \\
    &=
    \frac{\sum_{i=1}^{K} \lambda_i \brk{
        z_i (x_{i^*} - r)
        -
        z_{i^*} (x_i - r)
    }}{z_{i^*} \sum_{j=1}^{K} \lambda_j z_j}
    \\
    &=
    \sum_{i=1}^{K} \lambda_i \frac{z_i}{\sum_{j=1}^{K} \lambda_i z_i} \brk{
        \frac{x_{i^*} - r}{z_{i^*}}
        -
        \frac{x_i - r}{z_i}
    }
    \ge
    \frac{z_{\text{min}}}{z_{\text{max}}} \sum_{i=1}^{K} \lambda_i \Di
    ,
\end{align*}
where
$
    z_{\text{min}} 
    = 
    \min_{i \neq i^*} z_i
$
and
$
    z_{\text{max}} 
    = 
    \max_{i \in \actionSet} z_i
    .
$
Using the gap ratio defined in \cref{eq:diameterAndGapRatio} we get a linear gap with $\pLinGap = \gapRatio z_{\text{max}} / z_{\text{min}}$.

\textbf{Stability:}
For $x \in \RHatFunc[\text{Sh}]{\DistSetDelta}$ and $y \in \RHatFunc[\text{Sh}]{\funcSpace}$ we have that
\begin{align*}
    \abs{
        \hCompFunc{x_1, x_2}
        -
        \hCompFunc{y_1, y_2}
    }
    &=
    \abs*{
        \frac{x_1 - r}{\sqrt{\varepsilon_0 - x_1^2 + x_2}}
        -
        \frac{y_1 - r}{\sqrt{\varepsilon_0 - y_1^2 + y_2}}
    }
    \\
    &\le
    \abs*{
        \frac{x_1 - r}{\sqrt{\varepsilon_0 - y_1^2 + y_2}}
        -
        \frac{y_1 - r}{\sqrt{\varepsilon_0 - y_1^2 + y_2}}
    }
    +
    \abs*{
        \frac{x_1 - r}{\sqrt{\varepsilon_0 - x_1^2 + x_2}}
        -
        \frac{x_1 - r}{\sqrt{\varepsilon_0 - y_1^2 + y_2}}
    }
    \\
    &\le
    \frac{\abs{x_1 - y_1}}{\varepsilon_0}
    +
    \abs{x_1 - r}
    \abs*{
        \frac{
            \sqrt{\varepsilon_0 - y_1^2 + y_2}
            -
            \sqrt{\varepsilon_0 - x_1^2 + x_2}
        }{\sqrt{\varepsilon_0 - x_1^2 + x_2}\sqrt{\varepsilon_0 - y_1^2 + y_2}}
    }
    \\
    &\le
    \frac{\abs{x_1 - y_1}}{\varepsilon_0}
    +
    \frac{\abs{x_1 - r}}{2\varepsilon_0^{3/2}}
    \abs*{
        (x_1^2 - y_1^2) + (y_2 - x_2)
    }
    \\
    &\le
    \frac{\abs{x_1 - y_1}}{\varepsilon_0}
    +
    \abs{x_1 - r}
    \abs*{
        \frac{
            \sqrt{\varepsilon_0 - y_1^2 + y_2}
            -
            \sqrt{\varepsilon_0 - x_1^2 + x_2}
        }{\sqrt{\varepsilon_0 - x_1^2 + x_2}\sqrt{\varepsilon_0 - y_1^2 + y_2}}
    }
    \\
    &\le
    \frac{\abs{x_1 - y_1}}{\varepsilon_0}
    +
    \frac{\abs{x_1 - r}}{2\varepsilon_0^{3/2}}
    \brk{
        (x_1 - y_1)^2
        +
        2\abs{x_1}\abs{x_1 - y_1} 
        +
        \abs{y_2 - x_2}
    }
    \\
    &\le
    \max\brk[c]*{
        \varepsilon_0^{-1}
        ,
        2\varepsilon_0^{-3/2} \abs{x_1 - r}
    }
    \brk*{
        (x_1 - y_1)^2
        +
        (2\abs{x_1} + 1)\abs{x_1 - y_1} 
        +
        \abs{y_2 - x_2}
    }
    \\
    &\le
    2(1 + \abs{x_1}) \max\brk[c]*{
        \varepsilon_0^{-1}
        ,
        2\varepsilon_0^{-3/2} \abs{x_1 - r}
    }
    \brk*{
        \norm{x-y}_2^2
        +
        \norm{x-y}_2
    }
    .
\end{align*}
Since $(x_1, x_2) \in \RHatFunc[\text{Sh}]{\DistSetDelta}$, we can bound
$
    \abs{x_1}
    \le
    \max_{i \in \actionSet} \abs{\RHatFunc[\text{ave}]{\Fi}}
    .
$
Since the reward is also bounded in $\brk[s]{0,1}$ we further have that $\abs{x_1} \le 1$, giving us the constants
$
    \polyContModCoeff 
    =
    4 \max\brk[c]{
        \varepsilon_0^{-1}
        ,
        2\varepsilon_0^{-3/2} (\abs{r}+1)
    }
$
and
$\polyContModDeg = 2$.

\textbf{Smoothness:}
The idea here is to bound the spectral norm of the hessian. This is very similar to previous examples and in particular to Sortino ratio.

\subsection{Non-composite \EDRMabbrv s.} 
\label{appendix:nonDiffEDRM}

In this section we show the properties of $\VAR$ and $\CVAR$ required by our framework. Unless stated otherwise, we use the norm defined in \cref{eq:cvarNorm}. We recall the definitions of $\CVAR$ and $\VAR$ from \cref{eq:varDef,eq:cvarDef}, and specifically that we have that
\begin{align*}
   \RHatFunc[\CVAR]{F}
    =
    z^*
    -
    \frac{1}{\alpha} \int_{-\infty}^{z^*} F(x) dx
	,
\end{align*}
where $z^* = \RHatFunc[\VAR]{F}$. Before starting, we require the following technical lemma, proved in \cref{appendix:atRiskBoundsProof}.

\begin{lemma}[$\CVAR$ \textbf{and} $\mathbf{\VAR}$ \textbf{bounds}] 
\label{lemma:atRiskBounds}
	For any $F,G \in \funcSpace$ we have that
	\begin{align*}
    	\abs{\RHatFunc[\VAR]{F} - \RHatFunc[\VAR]{G}}
    	\le 
    	\frac{2\norm{F} + \norm{F-G}}{\min\brk[c]*{\alpha,1-\alpha}}
    	,
	\end{align*}
	and
	\begin{align*}
		0 
		\le
		\RHatFunc[\CVAR]{G}
		-
		\RHatFunc[\CVAR]{F} 
		&+
		\frac{1}{\alpha}\int_{-\infty}^{\RHatFunc[\VAR]{F}}\brk*{G(x) - F(x)} dx
		\\
		&\le
		\frac{1}{\alpha}\norm{F-G}\abs{\RHatFunc[\VAR]{F} - \RHatFunc[\VAR]{G}}
		.
	\end{align*}
	If additionally \cref{eq:positivePDF} holds, $F \in \DistSetDelta$, and $\norm{F-G} < \varMa$ then we also have that
	\begin{align*}
        \abs{\RHatFunc[\VAR]{F} - \RHatFunc[\VAR]{G}} 
        \le
        \varBa \norm{F-G}_\infty
        .
	\end{align*}
\end{lemma}

\subsubsection{Conditional Value at Risk (CVaR)} 
\label{appendix:subsection:cvar}
The following summarizes the properties of $\CVAR$ required for applying \cref{thm:regretBound}.
\begin{proposition}[$\mathbf{\CVAR}$ \textbf{properties}]
\label{prop:cvarProperties}
    We have that:
    \begin{enumerate}
        \item $\CVAR$ is convex;
        \item $\CVAR$ is stable with parameters
        $
            \polyContModCoeff
            =
            {4}/{\alpha \min\brk[c]*{\alpha,1-\alpha}}
            ,
            \polyContModDeg
            =
            2
            ,
            \concentrationConst
            =
            2/3
            ;
        $
        \item If \cref{eq:positivePDF} holds then $\CVAR$ is smooth with parameters
        $\pSmooth = 2\varBa / \alpha$ 
        and
        $M_0 = \varMa$.
    \end{enumerate}
\end{proposition}

\begin{proof}
    As previously mentioned, convexity follows from \cref{eq:cvarDef}, which expresses $\CVAR$ as a maximum over linear functions.
    
    \textbf{Stability.} Starting with the easier Requirement~\ref{item:StrongEDRMCond2} of stability, the concentration of $\norm{\FHatPi[(i)]{t} - \Fi}_\infty$ follows from the Dvoretzky-Kiefer-Wolfowitz inequality \citep{massart1990tight} with $\concentrationConst_0 = 2$, and since the other two terms are linear, the same holds for them by Hoeffding's inequality. As in \cref{lemma:compositeEDRM} (but for \emph{max} norm), we conclude that Requirement~\ref{item:StrongEDRMCond2} holds with $\concentrationConst = 2 \log 2 / \log 6 \ge 2/3$.
    Next, for Requirement~\ref{item:StrongEDRMCond1}, first notice that
    \begin{align*}
        \abs*{
            \frac{1}{\alpha}\int_{-\infty}^{\RHatFunc[\VAR]{F}}
            \brk*{F(x) - G(x)}
            dx
        }
        &\le
        \abs*{
            \frac{1}{\alpha} \int_{-\infty}^{0} \brk*{F(x) - G(x)} dx
        }
        +
        \abs*{
            \frac{1}{\alpha} \int_{0}^{\RHatFunc[\VAR]{F}} 
            \brk*{F(x) - G(x)}
            dx
        }
        \\
		&\le
		\frac{1}{\alpha}\brk[s]*{\norm{F-G} 
		+
		\abs{\RHatFunc[\VAR]{F}}\norm{F-G}} 
		\\
		&\le
		\frac{\norm{F-G}}{\alpha}\brk[s]*{1 + \abs{\RHatFunc[\VAR]{F}}}
		,
    \end{align*}
	and apply this to the second claim of \cref{lemma:atRiskBounds} to get that
	\begin{align*}
		\abs{\RHatFunc[\CVAR]{G} - \RHatFunc[\CVAR]{F}}
		\le 
		\frac{\norm{F-G}}{\alpha}\brk[s]*{
		    1
		    +
		    \abs{\RHatFunc[\VAR]{F}}
		    +
		    \abs{\RHatFunc[\VAR]{F} - \RHatFunc[\VAR]{G}}}.
	\end{align*}
	Finally, using the first part of \cref{lemma:atRiskBounds} we get that
	\begin{align*}
		\abs{\RHatFunc[\CVAR]{G} - \RHatFunc[\CVAR]{F}}
		&\le
		\frac{\norm{F-G}}{\alpha}\brk[s]*{1+\abs{\RHatFunc[\VAR]{F}} + \frac{2\norm{F} + \norm{F-G}}{\min\brk[c]*{\alpha,1-\alpha}}}
		\\
		&\le
		\frac{1}{\alpha}\brk*{
		    1
		    +
		    \abs{\RHatFunc[\VAR]{F}}
		    +
		    \frac{\max\brk[c]*{1, 2\norm{F}}}{\min\brk[c]*{\alpha,1-\alpha}}
	    }\brk[s]*{\norm{F-G} + \norm{F-G}^2}
		,
	\end{align*}
	which is Requirement~\ref{item:StrongEDRMCond1} of stability. Further using Assumption~\ref{assumption:boundedReward}, we have that
	$
	    \norm{F}, \abs{\RHatFunc[\VAR]{F}} \le 1,
	$
	and plugging this into the above gives the desired value for $\polyContModCoeff$.

	\textbf{Smoothness.} Assume that
	$
	    \DRHatAtOn[\CVAR]{F}{G}
	    =
	    -\frac{1}{\alpha} \int_{-\infty}^{\RHatFunc[\VAR]{F}} G(x) dx
	    .
	$
	We show that the smoothness condition holds under this assumption, which in turn implies that this assumption must be true (see definition of Frechet derivative). To that end, we use the second and third parts of \cref{lemma:atRiskBounds} to get that for any $F \in \DistSetDelta$ and $G \in \funcSpace$ satisfying $\norm{F-G} \le \varMa$ we have that
	\begin{align*}
	    &\abs*{
		    \RHatFunc[\CVAR]{G}
		    -
		    \RHatFunc[\CVAR]{F} 
		    -
		    \frac{-1}{\alpha}\int_{-\infty}^{\RHatFunc[\VAR]{F}}
		    \brk*{G(x) - F(x)}
		    dx
	    }
		\\
		&\qquad\le
		\frac{1}{\alpha}\norm{F-G}\abs{\RHatFunc[\VAR]{F} - \RHatFunc[\VAR]{G}}
		\\
		&\qquad\le
		\frac12 \frac{2\varBa}{\alpha}\norm{F-G}^2
		,
	\end{align*}
	as desired.
\end{proof}

\subsubsection{Value at Risk (VaR)} 
\label{appendix:subsection:var}

We begin with the following proposition, which proves the needed properties of $\VAR$ to get $\mathcal{O}(1/\sqrt{T})$ regret, as described in \cref{remark:sqrtRegret}.
\begin{proposition}[$\mathbf{\VAR}$ \textbf{properties}] \label{prop:varProperties}
    We have that:
    \begin{enumerate}
        \item $\VAR$ is quasiconvex;
        \item If \cref{eq:positivePDF} holds then $\VAR$ is stable with parameters
        $
            \concentrationConst
            =
            2/3
            ,
            \polyContModDeg
            =
            1
            ,
        $
        and
        \begin{align*}
            \polyContModCoeff
            =
            \max\brk[c]*{
                \varBa
                ,
                \frac{
                    \varMa
                    +
                    2
                }{\min\brk[c]*{\alpha,1-\alpha}\varMa}}
            .
        \end{align*}
    \end{enumerate}
\end{proposition}

\begin{proof}
    Starting with quasiconvexity, let $F_1, F_2 \in \DistSet$ and $\lambda \in \brk[s]*{0,1}$. Denote $F_\lambda = \lambda F_1 + \brk*{1-\lambda}F_2$, then by the definition of $\RHat[\VAR]$ we have that
	\begin{align*}
		F_\lambda \brk*{\max\brk[c]*{
		    \RHatFunc[\VAR]{F_1}
		    ,
		    \RHatFunc[\VAR]{F_2}
	    }}
		&=
		\lambda F_1 \brk*{\max\brk[c]*{
		    \RHatFunc[\VAR]{F_1}
		    ,
		    \RHatFunc[\VAR]{F_2}
	    }}
	    \\
		&\quad+
		\brk*{1-\lambda} F_2 \brk*{\max\brk[c]*{
		    \RHatFunc[\VAR]{F_1}
		    ,
		    \RHatFunc[\VAR]{F_2}
	    }}
		\ge
		\alpha
		.
	\end{align*}
	Using the definition of $\RHat[\VAR]$ another time we conclude that
	\begin{align*}
		\RHatFunc[\VAR]{F_\lambda}
		\le
		\max\brk[c]*{\RHatFunc[\VAR]{F_1}, \RHatFunc[\VAR]{F_2}}
		,
	\end{align*}
	which is one of the characterizations of quasiconvexity.

	\textbf{Stability.}
	Since we use the same norm as $\CVAR$, Requirement~\ref{item:StrongEDRMCond2} is proven by \cref{prop:cvarProperties}. As for Requirement~\ref{item:StrongEDRMCond1}, if $\norm{F-G} < \varMa$ then using the third part of \cref{lemma:atRiskBounds} we have that
	\begin{align*}
		\abs{\RHatFunc[\VAR]{F} - \RHatFunc[\VAR]{G}}
		\le
		\varBa \norm{F-G}
		.
	\end{align*}
	On the other hand, if $\norm{F-G} \ge \varMa$ then using the first part of \cref{lemma:atRiskBounds} we have that
	\begin{align*}
		\abs{\RHatFunc[\VAR]{F} - \RHatFunc[\VAR]{G}}
		&\le 
		\frac{2\norm{F} + \norm{F-G}}{\min\brk[c]*{\alpha,1-\alpha}}
		\\
		&=
		\frac{\norm{F-G}\brk*{1 + \frac{2\norm{F}}{\norm{F-G}}}}{\min\brk[c]*{\alpha,1-\alpha}}
		\\
		&\le 
		\frac{1 + \frac{2\norm{F}}{\varMa}}{\min\brk[c]*{\alpha,1-\alpha}} \norm{F-G}
		,
	\end{align*}
	and combining both results we obtain get that stability holds with $\polyContModDeg = 1$ and
	\begin{align*}
        \polyContModCoeff
        =
        \max\brk[c]*{
            \varBa
            ,
            \frac{
                \varMa
                +
                2\norm{F}
            }{\min\brk[c]*{\alpha,1-\alpha}\varMa}}
        .
    \end{align*}
    Further using Assumption~\ref{assumption:boundedReward}, we have that $\norm{F} \le 1$, which gives the desired value of $\polyContModCoeff$.
\end{proof}

The following result shows when $\VAR$ satisfies the conditions of \cref{theorem:StableEDRMoptPolicy}, and thus has a single arm infinite horizon oracle policy. We note that this holds regardless as shown \cref{proposition:VAROptPolicy}, which uses a different approach that is specific to $\VAR$. Denote the $\alpha$ level set of a function $F \in \DistSet$ by
$
    \levelSet{F}
    =
    \brk[c]{
        x \in \RR \big| F(x) = \alpha
    }
    .
$

\begin{proposition}[$\mathbf{\VAR}$ \textbf{\cref{theorem:StableEDRMoptPolicy} conditions}]
\label{proposition:varStability}
	If 
	$
        \abs{\levelSet{\alpha}}
        \le
        1
    $
    for all $F \in \DistSetDelta$.
    Then $\RHat[\VAR]$ satisfies then conditions of \cref{theorem:StableEDRMoptPolicy}, and thus has a single arm infinite horizon oracle policy.
\end{proposition}

\begin{proof}
	Unlike the remainder of this section, here we use the norm $\norm{F} = \norm{F}_\infty$. By the Glivenko-Cantelli theorem \cite{van2000asymptotic}, the convergence of the empirical distribution required in \cref{theorem:StableEDRMoptPolicy} is established. It thus remains to show that $\VAR$ is continuous on $\DistSetDelta$.
	Our condition on the level set can be interpreted in the following way. For any fixed $F \in \DistSetDelta$
	\begin{align}
		\label{eq:VaRCont1}
		&y 
		>
		\RHatFunc[\VAR]{F}
		\implies
		\exists c_y>0,~ s.t,~ F(y) \ge \alpha + c_y 
		\\
		\label{eq:VaRCont2}
		&y
		< 
		\RHatFunc[\VAR]{F}
		\implies 
		\exists c_y>0,~ s.t,~ F(y) \le \alpha - c_y
		.
	\end{align}
	Let $g : [ -\frac{\alpha}{2}, \frac{1 - \alpha}{2}] \to \RR$ be given by, 
	$
	    g(\delta)
	    =
	    \RHatFunc[{\VAR[\alpha + \delta]}]{F}
	    .
    $
    We show that $g$ is continuous at $0$. $g$ is monotone non decreasing and so has left and right limits at $0$. Let $\seqDef{\delta_n}{n=1}{\infty} \searrow 0$ and denote,
	\begin{align*}
		\lim_{n \to \infty} g(\delta_n) 
		=
		a^+
		.
	\end{align*}
	By the monotonicity of $g$ we have that $a^+ \ge g(0)$. Using \cref{eq:VaRCont1} we have that, for any $\varepsilon > 0$,
	\begin{align*}
		F(g(0) + \varepsilon) 
		\ge
		\alpha + c_\varepsilon
		,
	\end{align*}
	where $c_\varepsilon > 0$. So, by the expression of $\RHat[\VAR]$, we have that,
	\begin{align*}
		g(0) + \varepsilon \ge g(c_\varepsilon) 
		\ge
		a^+
		,
	\end{align*}
	where the second inequality follows by the monotonicity of $g$. So, $g(0) \le a^+ \le g(0) + \varepsilon$ for all $\varepsilon > 0$ and so $a^+ = g(0)$.
	Now take $\seqDef{\bar{\delta}_n}{n=1}{\infty} \nearrow 0$ and denote,
	\begin{align*}
		\lim_{n \to \infty} g(\bar{\delta}_n) = a^-.
	\end{align*}
	A similar set of arguments shows that $a^- = g(0)$, and so $g$ is continuous at $0$.

	By the continuity of $g$, for any $\varepsilon > 0$, there exists $\delta_\varepsilon >0$ such that for all $\abs{\beta} \le \delta_\varepsilon$ we have that,
	\begin{align*}
		\abs{g(0) - g(\beta)}
		\le
		\varepsilon.
	\end{align*}
	For any $G \in \funcSpace$ satisfying $\norm{F - G}_\infty \le \delta_\varepsilon$ we have that,
	\begin{align*}
		\RHatFunc[\VAR]{F} - \RHatFunc[\VAR]{G}
		&= 
		g(0) - \min\setDef[\Big]{y \given G(y) \ge \alpha}
		\\
		&\le 
		g(0) - \min\setDef[\Big]{y \given F(y) \ge \alpha - \delta_\varepsilon}
		= 
		g(0) - g(-\delta_\varepsilon) 
		\le
		\varepsilon
		.
	\end{align*}
	We also have,
	\begin{align*}
		\RHatFunc[\VAR]{G} - \RHatFunc[\VAR]{F}
		&= 
		\min\setDef[\Big]{y \given G(y) \ge \alpha} - g(0)
		\\
		&\le 
		\min\setDef[\Big]{y \given F(y) \ge \alpha + \delta_\varepsilon} - g(0)
		= 
		g(\delta_\varepsilon) - g(0)
		\le
		\varepsilon
		.
	\end{align*}
	We conclude that, $\abs{\RHatFunc[\VAR]{F} - \RHatFunc[\VAR]{G}} \le \varepsilon$ thus concluding the continuity of $\RHat[\VAR]$ on $\DistSetDelta$.
\end{proof}

Finally, we prove \cref{proposition:VAROptPolicy}, showing that $\VAR$ always has a single arm infinite horizon oracle policy.	
	
\begin{proof}[of \cref{proposition:VAROptPolicy}]
	We begin calculating the performance of a single arm policy. We then proceed to show that the performance of any policy is upper bounded by that of the best single arm policy.
	
	\noi\textbf{Performance of \simple\space policies.} Let $\policy[i]$ be the policy that always plays arm $i$. We claim that, 
	$
	    \EE{\RPiNamed[{\policy[i]}]{\VAR}} 
	    =
	    \RHatFunc[\VAR]{\Fi}
	    .
    $
	If $\Fi$ represents a degenerate random variable then the expression holds trivially. Otherwise, let $y < \RHatFunc[\VAR]{\Fi} = a_i$, then there exists $\delta_y > 0$, such that, $\Fi(y) \le \alpha - \delta_y$. Using the strong law of large numbers \cite{Simonnet1996} we have that
	\begin{align*}
		\lim_{t \to \infty} \FHatPi[{\policy[i]}]{t}(y)
		\overset{a.s}{=}
		\Fi(y) \le \alpha - \delta_y.
	\end{align*}
	Let $E$ be the event on which the convergence occurs. Then $\forall \omega \in E$ there exists $T(\omega)$ such that $\forall t > T(\omega)$, we have that,
	\begin{align*}
		\FHatPi[{\policy[i]}]{t}(y,\omega) 
		\le
		\Fi(y) + \delta_y / 2
		\le
		\alpha - \delta_y / 2 
		<
		\alpha
		.
	\end{align*}
	This implies that $\RHatFunc[\VAR]{\FHatPi[{\policy[i]}]{t}(\omega)} > y$ for all $t \ge T(\omega)$. We get that $\RPiNamed[{\policy[i]}]{\VAR} \ge y$ almost surely, and taking the expectation we get $\EE{\RPiNamed[{\policy[i]}]{\VAR}} \ge y$. Since this holds for all $y < \RHatFunc[\VAR]{\Fi}$, then,
	\begin{align} 
	\label{eq:varOracle01}
	    \EE{\RPiNamed[{\policy[i]}]{\VAR}}
	    \ge
	    \RHatFunc[\VAR]{\Fi}
	    .
	\end{align} 
	On the other hand, using the Law of the iterated logarithm \cite{Klenke2014} we get that
	\begin{align*}
		\limsup_{t \to \infty} 
		    \frac{t}{\lambda\sqrt{2t\log\log t}} \brk*{\FHatPi[{\policy[i]}]{t}(a_i) - \Fi(a_i)}
		    =
		    1
		    \quad a.s
		    ,
	\end{align*}
	where, $\lambda = \Fi(a_i)\brk*{1 - \Fi(a_i)} \neq 0$ since $\Fi$ is non-degenerate. We conclude that
	\begin{align*}
		\FHatPi[{\policy[i]}]{t}(a_i) > \Fi(a_i)
		\ge
		\alpha
		\quad i.o
		,
	\end{align*}
	and thus
	\begin{align}
	\label{eq:ProofVAReq3}
		\RPiNamed[{\policy[i]}]{\VAR}
		= 
		\liminf_{t \to \infty} \RHatFunc[\VAR]{\FHatPi[{\policy[i]}]{t}}
		\le 
		a_i
		\quad a.s.
	\end{align}
	Taking expectation on both sides, we conclude that
	\begin{align*}
		\EE{\RPiNamed[{\policy[i]}]{\VAR}} 
		\le
		\RHatFunc[\VAR]{\Fi}
		,
	\end{align*}
	which together with \cref{eq:varOracle01} proves that 
	$
	    \EE{\RPiNamed[{\policy[i]}]{\VAR}}
	    =
	    \RHatFunc[\VAR]{\Fi}
	    .
    $
    Now, recall that in \cref{prop:varProperties} we showed that $\RHat[\VAR]$ is quasiconvex. We thus have that there exists $i^* \in \actionSet$ such that for all $F \in \DistSetDelta$
	\begin{align} 
	\label{eq:ProofVAReq4}
	    \RHatFunc[\VAR]{F}
	    \le
	    \RHatFunc[\VAR]{\FiOpt}
	    =
	    \EE{\RPiNamed[{\policy[{i^*}]}]{\VAR}} 
	    =
	    a^*
	    .
	\end{align}
	
	\noi\textbf{Global optimizer.} Our purpose will be to show that
	\begin{align}
	\label{eq:ProofVAReq1}
		\FHatPi[\pi]{t}(a^*) - \alpha
		>
		0
		\quad i.o.
	\end{align}
	Similarly to \cref{eq:ProofVAReq3}, this implies that
	\begin{align*}
		\RPiNamed{\VAR} 
		\le
		a^*
		\quad a.s,
	\end{align*}
	and taking the expectation we conclude that, 
	$
	    \EE{\RPiNamed{\VAR}} 
	    \le
	    \EE{\RPiNamed[{\policy[{i^*}]}]{\VAR}}
	    ,
    $
    thus concluding the proof.
	
	By \cref{eq:ProofVAReq4}, we have that
	\begin{align*}
		\FProxPi{t}(a^*)
		=
		\frac{1}{t}\sum_{s=1}^{t}\Fi[\policyAt{s}](a^*)
		\ge
		\alpha
		.
	\end{align*}
	We thus get that,
	\begin{align} 
	\label{eq:ProofVAReq2}
		\FHatPi{t}(a^*) - \alpha 
		\ge
		\FHatPi{t}(a^*) - \FProxPi{t}(a^*)
		= 
		\frac{1}{t} \sum_{s=1}^{t} \indEvent{\XtPi[s]\le a^*} - \Fi[\policyAt{s}](a^*) 
		= 
		\frac{1}{t}\sum_{s=1}^{t}Y_s = \frac{1}{t} W_t
		,
	\end{align}
	where $Y_s = \brk[s]*{\indEvent{\XtPi[s]\le a^*} - \Fi[\policyAt{s}](a^*)}$ and $W_t = \sum_{s=1}^{t}Y_s$.
	We split our remaining analysis into two cases.
	
	The first is when policy $\pi$ chooses some non-degenerate arm infinitely often (i.o). For this case we use the Law of the iterated logarithm for martingales given in \cite{fisher1992law}. We use the same notation as in \cite{fisher1992law} aside for denoting the martingale $W_t$ instead of $U_t$, and its difference sequence by $Y_t$ instead of $X_t$ (to avoid confusion with existing notation). We start by showing $W_t$ is a martingale with respect to its natural filtration
	\begin{align*}
		&\EEBrk{W_{t+1} | W_1, \ldots,W_t}
		\\
		&\quad=
		W_t + \EEBrk{Y_{t+1} | W_1, \ldots,W_t}
		\\
		&\quad= 
		W_t + \EEBrk{\EEBrk{Y_{t+1} | \policyAt{t+1}} | W_1, \ldots,W_t}
		\\
		&\quad=
		W_t + \EEBrk{0 | W_1, \ldots,W_t} 
		=
		W_t
		,
	\end{align*}
	where the second equality is the law of total probability in addition to $Y_{t+1} | \policyAt{t+1}$ being independent of $W_1, \ldots,W_t$. Furthermore, $\EE{\abs{W_t}} \le t < \infty$, so $W_t$ is a martingale.
	
	Next, let $s_t^2 = \sum_{s=1}^{t} \EEBrk{Y_s^2 | W_1, \ldots, W_{s-1}}$. Since $\policy$ chooses a non-degenerate arm infinitely often then, $s_t^2 \to \infty$.
	
	Finally, let $t_0$ denote the first time $\policy$ chooses a non-degenerate arm. So, we can choose $K_t$ in the following way,
	\begin{align*}
		K_t = \varphi(s_{t_0}) / s_{t_0}.
	\end{align*}
	Clearly, there exists $K > 0$ such that, $\limsup_{t \to \infty} K_t < K$. Furthermore,
	\begin{align*}
		\abs{Y_t}
		\le 
		\begin{cases}
			0 ,& t < t_0
			\\
			1 ,& t \ge t_0
		\end{cases}
		\le
		K_t s_t / \varphi(s_t)
		.
	\end{align*}
	So the conditions of Theorem 1 in \cite{fisher1992law} are met and we conclude that
	\begin{align*}
		\limsup_{t \to \infty} W_t / s_t \varphi(s_t)
		>
		0 
		\quad a.s
		.
	\end{align*}
	This means that, $W_t > 0$ infinitely often and substituting into \cref{eq:ProofVAReq2} we conclude that \cref{eq:ProofVAReq1} holds.
	
	In the second case, any non-degenerate arm is chosen a finite number of times. Let $i_b$ denote the index of the largest degenerate arm and $a_b$ be its value. Clearly,
	\begin{align*}
		a_b 
		=
		\EE{\RPiNamed[{\policy[{i_b}]}]{\VAR}} 
		\le
		\EE{\RPiNamed[{\statPolicy[e_{i^*}]}]{\VAR}}
		=
		a^*
		.
	\end{align*}
	Denote, $I_b = \setDef[\Big]{i \given \Fi \text{ is degenerate}}$. Since non-degenerate arms are pulled a finite number of times then,
	\begin{align*}
		\lim_{t \to \infty} \piAt{t}
		=
		0
		\quad
		,\forall i \notin I_b
		,
	\end{align*}
	where $\piAt{t}$ is defined in \cref{eq:ActionDistribution}.
	Since there are finitely many arms, this implies that
	\begin{align*}
		\lim_{t \to \infty} \sum_{i \in I_b} \piAt{t} = 1.
	\end{align*}
	Now, since $a_b$ is such that $\Fi(a_b) = \FHatPi[(i)]{t}(a_b) = 1$ for all $i \in I_b$, we conclude that
	\begin{align*}
		\lim_{t \to \infty} \FHatPi{t}(a^*) 
		&\ge
		\lim_{t \to \infty} \FHatPi{t}(a_b)
		\\
		&=
		\lim_{t \to \infty} \sum_{i=1}^{K} \piAt{t} \FHatPi[(i)]{t}(a_b) 
		\\
		&\ge
		\lim_{t \to \infty} \sum_{i \in I_b} \piAt{t} \FHatPi[(i)]{t}(a_b) 
	    \\
	    &=
	    \lim_{t \to \infty} \sum_{i \in I_b} \piAt{t}
	    =
	    1
	    .
	\end{align*}
	Since $\alpha < 1$, we can clearly conclude \cref{eq:ProofVAReq1} holds, thus finishing the proof.
\end{proof}

\subsubsection{Proof of Lemma~\ref{lemma:atRiskBounds}}
\label{appendix:atRiskBoundsProof}

We prove the individual claims of \cref{lemma:atRiskBounds}.

\textbf{First claim.}
We start by showing that for all $G \in \funcSpace$
\begin{align} \label{eq:varBound1}
    \abs{\RHatFunc[\VAR]{G}} 
    \le
    \frac{\norm{G}}{\min\brk[c]*{\alpha,1-\alpha}}
    .
\end{align}
Suppose that $\RHatFunc[\VAR]{G} \ge 0$, then we have that
\begin{align*}
    \norm{G}
    \ge 
    \int_{0}^{\infty} \brk*{1 - G(y)} dy
    &\ge
    \int_{0}^{\RHatFunc[\VAR]{G}} \brk*{1 - G(y)} dy 
    \\
    &\ge
    \int_{0}^{\RHatFunc[\VAR]{G}} \brk*{1 - \alpha} dy
    =
    \brk*{1-\alpha} \RHatFunc[\VAR]{G} \ge 0
    .
\end{align*}
On the other hand, suppose that $\RHatFunc[\VAR]{G} \le 0$, then we have that
\begin{align*}
    \norm{G}
    \ge 
    \int_{-\infty}^{0} G(y) dy
    &\ge
    \int_{\RHatFunc[\VAR]{G}}^{0} G(y) dy 
    \\
    &\ge
    \int_{\RHatFunc[\VAR]{G}}^{0} \alpha dy
    =
    -\alpha \RHatFunc[\VAR]{G} \ge 0
    ,
\end{align*}
thus concluding \cref{eq:varBound1}. Using this result, we have that for all $F,G \in \funcSpace$
\begin{align*}
    \abs{\RHatFunc[\VAR]{F} - \RHatFunc[\VAR]{G}}
    &\le
    \abs{\RHatFunc[\VAR]{F}} + \abs{\RHatFunc[\VAR]{G}} 
    \\
    &\le
    \frac{\norm{F} + \norm{G}}{\min\brk[c]*{\alpha,1-\alpha}} 
    \\
    &\le
    \frac{2\norm{F} + \norm{F-G}}{\min\brk[c]*{\alpha,1-\alpha}}
    ,
\end{align*}
as desired.

\textbf{Second claim.}	
We have that
\begin{align*}
    &\RHatFunc[\CVAR]{G} - \RHatFunc[\CVAR]{F}
    \\
    &=
    \RHatFunc[\VAR]{G}
    -
    \frac{1}{\alpha} \int_{-\infty}^{\RHatFunc[\VAR]{G}} G(y) dy 
    -
    \RHatFunc[\VAR]{F}
    +
    \frac{1}{\alpha} \int_{-\infty}^{\RHatFunc[\VAR]{F}} F(y) dy 
    \\
    &=
    \RHatFunc[\VAR]{G}
    -
    \RHatFunc[\VAR]{F} 
    +
    \frac{1}{\alpha} \int_{\RHatFunc[\VAR]{G}}^{\RHatFunc[\VAR]{F}} G(y) dy 
    -
    \frac{1}{\alpha} \int_{-\infty}^{\RHatFunc[\VAR]{F}} \brk*{G(y) - F(y)} dy
    \\
    &=
    \frac{1}{\alpha} \int_{\RHatFunc[\VAR]{G}}^{\RHatFunc[\VAR]{F}} \brk*{G(y)-\alpha} dy
    -
    \frac{1}{\alpha} \int_{-\infty}^{\RHatFunc[\VAR]{F}} \brk*{G(y) - F(y)} dy
    ,
\end{align*}
and changing sides we get
\begin{align*}
    \RHatFunc[\CVAR]{G}
    -
    \RHatFunc[\CVAR]{F} 
    +
    \frac{1}{\alpha} \int_{-\infty}^{\RHatFunc[\VAR]{F}} \brk*{G(y) - F(y)} dy
    =
    \frac{1}{\alpha} \int_{\RHatFunc[\VAR]{G}}^{\RHatFunc[\VAR]{F}} \brk*{G(y)-\alpha} dy
    .
\end{align*}
It therefore suffices to show that
\begin{align*}
    0 
    \le
    \frac{1}{\alpha} \int_{\RHatFunc[\VAR]{G}}^{\RHatFunc[\VAR]{F}} \brk*{G(y)-\alpha} dy
    \le 
    \frac{1}{\alpha}\norm{F-G}\abs{\RHatFunc[\VAR]{F} - \RHatFunc[\VAR]{G}}
    .
\end{align*}
Beginning with the left inequality, we have that
\begin{align} 
\label{eq:cvarResPositive}
\begin{aligned}
    &\frac{1}{\alpha} \int_{\RHatFunc[\VAR]{G}}^{\RHatFunc[\VAR]{F}} \brk*{G(y)-\alpha} dy 
    \\
    &\ge 
    \frac{1}{\alpha} \brk*{\RHatFunc[\VAR]{F} - \RHatFunc[\VAR]{G}}\brk*{G\brk*{\min\brk[c]*{\RHatFunc[\VAR]{G}, \RHatFunc[\VAR]{F}}} - \alpha} \\
    &= 
    \begin{cases}
        \frac{1}{\alpha} \underbrace{\brk*{\RHatFunc[\VAR]{F} - \RHatFunc[\VAR]{G}}}_{\le 0}\underbrace{\brk*{G\brk*{\RHatFunc[\VAR]{F}} - \alpha}}_{\le 0} , \RHatFunc[\VAR]{G} \ge \RHatFunc[\VAR]{F}
        \\
        \frac{1}{\alpha} \underbrace{\brk*{\RHatFunc[\VAR]{F} - \RHatFunc[\VAR]{G}}}_{\ge 0}\underbrace{\brk*{G\brk*{\RHatFunc[\VAR]{G}} - \alpha}}_{\ge 0} , \RHatFunc[\VAR]{G} < \RHatFunc[\VAR]{F}
    \end{cases} 
    \\
    & \ge
    0
    ,
\end{aligned}
\end{align}
where the final inequality holds since $G(y) \gtreqless \alpha$ for $y \gtreqless \RHatFunc[\VAR]{G}$. 
Next, for the right hand side inequality we have that
\begin{align*}
    &\frac{1}{\alpha} \int_{\RHatFunc[\VAR]{G}}^{\RHatFunc[\VAR]{F}} \brk*{G(y)-\alpha} dy 
    \\
    &=
    \frac{1}{\alpha} \int_{\RHatFunc[\VAR]{G}}^{\RHatFunc[\VAR]{F}} \brk*{F(y)-\alpha} dy
    +
    \frac{1}{\alpha} \int_{\RHatFunc[\VAR]{G}}^{\RHatFunc[\VAR]{F}} \brk*{G(y)-F(y)} dy 
    \\
    &\le
    \underbrace{\frac{1}{\alpha} \int_{\RHatFunc[\VAR]{G}}^{\RHatFunc[\VAR]{F}} \brk*{F(y)-\alpha}}_{(*)}
    +
    \frac{1}{\alpha}\norm{F-G}_\infty \abs{\RHatFunc[\VAR]{F} - \RHatFunc[\VAR]{G}}
    \\
    &\le
    \frac{1}{\alpha}\norm{F-G} \abs{\RHatFunc[\VAR]{F} - \RHatFunc[\VAR]{G}},
\end{align*}
where $(*) \le 0$ is obtained by exchanging the roles of $F$ and $G$ in \cref{eq:cvarResPositive}.

\textbf{Third claim.}
Consider \cref{eq:positivePDF} with $y = \norm{F-G}_\infty \le \varMa$, and notice that $F \brk*{\RHatFunc[\VAR]{F} + y} \ge \alpha$ for any $y \ge 0$. Then we have that
\begin{align*}
    F \brk*{\RHatFunc[\VAR]{F} + \varBa \norm{F-G}_\infty} - \alpha
    \ge
    \norm{F-G}_\infty
    .
\end{align*}
Using this we get
\begin{align*}
    G \brk*{\RHatFunc[\VAR]{F} + \varBa \norm{F-G}_\infty}
    \ge
    F \brk*{\RHatFunc[\VAR]{F} + \varBa \norm{F-G}_\infty} - \norm{F-G}_\infty
    \ge
    \alpha
    ,
\end{align*}
which by the definition of $\RHat[\VAR]$ in \cref{eq:varDef} implies that
\begin{align*}
    \RHatFunc[\VAR]{G}
    \le
    \RHatFunc[\VAR]{F}
    + 
    \varBa \norm{F-G}_\infty
    ,
\end{align*}
and changing sides we get
\begin{align}
\label{eq:varBound3}
    \RHatFunc[\VAR]{G} - \RHatFunc[\VAR]{F} 
    \le
    \varBa \norm{F-G}_\infty
    .
\end{align}
On the other hand, consider \cref{eq:positivePDF} with $y = -c \norm{F-G}_\infty \ge -\varMa$, where $1 < c \le \frac{\varMa}{\norm{F-G}_\infty}$. Noticing that $F \brk*{\RHatFunc[\VAR]{F} - y} \le \alpha$ for any $y \ge 0$, we have that
\begin{align*}
    -\brk*{F \brk*{\RHatFunc[\VAR]{F} - c \varBa \norm{F-G}_\infty} - \alpha}
    \ge
    c \norm{F-G}_\infty
    ,
\end{align*}
and changing sides we get
\begin{align*}
    F \brk*{\RHatFunc[\VAR]{F} - c \varBa \norm{F-G}_\infty}
    \le
    \alpha - c \norm{F-G}_\infty
    .
\end{align*}
Using this we get
\begin{align*}
    G \brk*{\RHatFunc[\VAR]{F} - c \varBa \norm{F-G}_\infty}
    &\le 
    F \brk*{\RHatFunc[\VAR]{F} - c \varBa \norm{F-G}_\infty} 
    +
    \norm{F-G}_\infty 
    \\
    &\le
    \alpha + (1-c)\norm{F-G}_\infty 
    \\
    &<
    \alpha
    ,
\end{align*}
and using the definition of $\RHat[\VAR]$ in \cref{eq:varDef} and changing sides we get
\begin{align*}
    \RHatFunc[\VAR]{G} - \RHatFunc[\VAR]{F} 
    \ge
    -c \varBa \norm{F-G}_\infty.
\end{align*}
Notice that the constant $c$ may be arbitrarily close to $1$. We thus conclude that
\begin{align*}
    \RHatFunc[\VAR]{G} - \RHatFunc[\VAR]{F} 
    \ge
    -\varBa \norm{F-G}_\infty
    ,
\end{align*}
which combined with \cref{eq:varBound3} implies the desired.

\rev
{
\subsection{Utility-Based Shortfall Risk (UBSR)}
This is a generalization of $\VAR$ that is defined as
\begin{align} 
\label{eq:ubsrDef}
	\RHatFunc[\UBSR]{F}
	=
	\inf_{\xi \in \RR}\brk[c]*{\xi ~\bigg|~ \int_{\RR} l(x-\xi) dF(x) \le \alpha}
	,
\end{align}
where $l: \RR \to \RR$ is an increasing and $G$ Lipschitz utility function. Under these assumptions, \cite{bhat2019concentration} show in their Lemma 4 that $\RHat[\UBSR]$ is $G$ Lipschitz with respect to the 1-Wasserstein metric, i.e.,
\begin{align*}
    \abs{\RHatFunc[\UBSR]{F_1} - \RHatFunc[\UBSR]{F_2}}
    \le
    G \norm{F_1 - F_2}_1
    =
    G \int_R \abs{F_1(x) - F_2(x)} dx
    .
\end{align*}
Moreover, for sub-Gaussian arm distributions, their Lemma 7, which is extracted from \cite{fournier2015rate}, gives the appropriate concentration of the empirical distribution in Wasserstein distance and thus we conclude that it $\RHat[\UBSR]$ is stable. We show that $\RHat[\UBSR]$ is quasi-convex and thus we can apply \cref{prop:stableHorizonGap,theorem:UUCB} to get that $\UUCB$ incurs $\tilde{O}(1 / \sqrt{T})$ regret. We suspect that for $l$ that is also smooth and convex it is possible to verify the smoothness and linear gap assumptions and thus invoke \cref{thm:regretBound} to get $\tilde{O}(\log T / T)$ regret. Unfortunately we did not manage to verify these details formally and thus leave them as a conjecture.
\\
Finally, we verify the quasi-convexity. Let $F_1,F_2$ be two distributions and $\xi_1, \xi_2$ their respective minimizers in $\RHat[\UBSR]$. Assume without loss of generality that $\xi_1 \le \xi_2$. Since $l$ is increasing we have that for any $\epsilon > 0$ and $\lambda \in [0,1]$
\begin{align*}
    \int_{\RR} l(x-(\xi_2 + \epsilon) d(\lambda F_1 + (1-\lambda)F_2)(x)
    &
    =
    \lambda \int_{\RR} l(x-(\xi_2 + \epsilon) d F_1 (x)
    +
    (1-\lambda)\int_{\RR} l(x-(\xi_2 + \epsilon) d F_2(x)
    \\
    &
    \le
    \lambda \int_{\RR} l(x-(\xi_1 + \epsilon) d F_1 (x)
    +
    (1-\lambda)\int_{\RR} l(x-(\xi_2 + \epsilon) d F_2(x)
    \\
    &
    \le
    \lambda \alpha + (1-\lambda) \alpha
    =
    \alpha
    .
\end{align*}
This implies that $\xi_2 + \epsilon$ is inside the feasible set of $\RHatFunc[\UBSR]{\lambda F_1 + (1-\lambda) F_2}$ for all $\epsilon>0$. We conclude that \begin{align*}
    \RHatFunc[\UBSR]{\lambda F_1 + (1-\lambda) F_2}
    \le
    \xi_2
    =
    \RHatFunc[\UBSR]{F_2}
    =
    \max \brk[c]*{\RHatFunc[\UBSR]{F_1},\RHatFunc[\UBSR]{F_2}}
    ,
\end{align*}
as desired.
}
{}

\subsection{Counter examples}
\label{appendix:subsection:counterExamples}

\rev{In this section we discuss performance measures for which \cref{eq:singleOpt} does not hold. It will thus be necessary to compare performance not to the best single arm policy, but the policy with the best arm mixture. To that end, let $\statPolicy$ be a simple randomized policy that at each turn draws an arm $i.i.d$ with distribution $p \in \simplex$, i.e., $\PP{\statPolicyAt{t} = i} = p_i$.}{}

\paragraph{$\mathbf{\RHat[bad1]}$ details.} 
Examples such as $\VAR$ are rather uncommon, but, when encountered, their analysis proves challenging. This fact might motivate a more general framework for \EDRMabbrv s, ideas for which, can be drawn from the proof of \cref{proposition:VAROptPolicy}. The following examples show the types of problems such frameworks could have or would need to address. Define an \EDRMabbrv~ by,
\begin{align*}
    \RHatFunc[bad1]{F}
    =
    \RHatFunc[{\VAR[0.1]}]{F} + \RHatFunc[{\VAR[0.9]}]{F}
    ,
\end{align*}
where the values $0.1,~0.9$ were chosen arbitrarily. When the two components of $\RHat[bad1]$ are stable then it is clear that so is $\RHat[bad1]$. We show that when this is not the case, then it is possible that no \simple\space policy is optimal. Consider a problem with two arms having the following distributions,
\begin{align*}
    \Fi[1](y) 
    =
    \begin{cases}
        0 & y < 0
        \\
        y/10 & 0 \le y < 1 
        \\
        0.1 & 1 \le y < 5 
        \\
        y/50 & 5 \le y < 50 
        \\
        1 & y \ge 50
    \end{cases} 
    \qquad
    \Fi[2](y) 
    = 
    \begin{cases}
        0 & y < 5
        \\
        1 & y \ge 5
        .
    \end{cases}
\end{align*}
Notice that $F^{(1)}$, $F^{(2)}$ satisfy the conditions of \cref{theorem:StableEDRMoptPolicy}. So, using an intermediate result of \cref{theorem:StableEDRMoptPolicy} we have that
$
    \lim_{t \to \infty} \RHatFunc[{\VAR[0.9]}]{\FHatPi[\statPolicy]{t}}
    =
    \RHatFunc[{\VAR[0.9]}]{\Fp}
$
almost surely. Using this convergence we get that
\begin{align*}
    \liminf_{t \to \infty}\RHatFunc[bad1]{\FHatPi[\statPolicy]{t}}
    &\overset{a.s}{=}
    \liminf_{t \to \infty} \RHatFunc[{\VAR[0.1]}]{\FHatPi[\statPolicy]{t}} 
    + \lim_{t \to \infty} \RHatFunc[{\VAR[0.9]}]{\FHatPi[\statPolicy]{t}}
    \\
    &\overset{a.s}{=}
    \RHatFunc[{\VAR[0.1]}]{\Fp}
    +
    \RHatFunc[{\VAR[0.9]}]{\Fp}
    .
\end{align*}
where the convergence of $\RHatFunc[{\VAR[0.1]}]{\FHatPi[\statPolicy]{t}}$ is an intermediate result in \cref{proposition:VAROptPolicy}. Evaluating these last terms we conclude the expression for the performance of a \simple\space policy $\statPolicy$ $(p = (p_1,p_2),~ p_1 = 1 - p_2)$,
\begin{align*}
    \EE{\RPiNamed[\statPolicy]{bad1}} 
    = 
    \begin{cases}
        46 & p_2 = 0 
        \\
        5 + (45 - 50p_2) / (1-p_2) & 0 < p_2 < 8/9 
        \\
        10 & 8/9 < p_2 \le 1
    \end{cases}
\end{align*}
It does not attain a maximum over the simplex and thus there is no optimizer inside the set of \simple\space policies. However, the following non-\simple\space policy is optimal,
\begin{align*} 
    \policyAt[*bad1]{t} 
    = 
    \begin{cases}
        2 & t=1 \quad or \quad \frac{\brk*{t-1}\FHatPi{t-1}(1) + 1}{t} \ge 0.1 
        \\
        1 &  \mbox{otherwise.}
    \end{cases}
\end{align*}
We explain why $\pi^*_{bad1}$ is an oracle policy. Each of the summands in $\RHat[bad1]$ has a \simple\space oracle policy with appropriate optimal performance. Summing these performances provides an upper bound on the performance of $\RHat[bad1]$. More specifically, $\RHat[bad1]$ is bounded by $50$. We show that, $\pi^*_{bad1}$ achieves this value and is thus an oracle policy. It is easy to show by induction that for $(t \ge 1)$, $\FHatPi[\pi^{* bad1}]{t}(1) < 0.1$.
This implies that $\RHatFunc[{\VAR[0.1]}]{\FHatPi[\pi^{* bad1}]{t}} \ge 5$, but $5$ is also an upper bound an so, $\liminf_{t \to \infty} \RHatFunc[{\VAR[0.1]}]{\FHatPi[\pi^{* bad1}]{t}}=5$. Finally, it is a technical result to show that $\lim_{t \to \infty} \FHatPi[\pi^{* bad1}]{t}(5) = 0.1$ almost surely, which implies $\lim_{t \to \infty} \piAt[]{t} \overset{a.s}{=} e_1 = (1, 0)$. Since $\RHat[{\VAR[0.9]}]$ is stable then this implies that
\begin{align*}
    \lim_{t \to \infty} \RHatFunc[{\VAR[0.9]}]{\FHatPi[\pi^{* bad1}]{t}}
    \overset{a.s}{=} 
    \RHatFunc[{\VAR[0.9]}]{\Fi[1]}
    =
    45
    ,
\end{align*}
and taking expectation the result is concluded.

The problem exhibited here is the lack of an optimizer within the set of \simple\space policies. A way of ensuring that this does not occur is to require that the performance of \simple\space policies be upper semi-continuous (with respect to $p$).

\paragraph{$\mathbf{\RHat[bad2]}$ details.} The following example shows that when the performance of \simple\space policies is not lower semi continuous, then while an optimizer exists within the set of \simple\space policies, it might not be a global optimizer. Define an \EDRMabbrv\space by,
\begin{align*}
    \RHatFunc[bad2]{F} 
    =
    \RHatFunc[{\VAR[0.1]^{++}}]{F}
    +
    5 \indEvent{F({10}^{-}) - F({1}^{+}) > 0 ~or~ F({1}^{-}) > 0}
    ,
\end{align*}
where,
\begin{align*}
    &\RHatFunc[+]{F;x} 
    =
    \max_{y \in \RR}\brk[c]*{y \ge x ~\Big|~ F(y) = F(x)} 
    \\
    &\RHatFunc[{\VAR[0.1]^+}]{F} 
    = 
    \begin{cases}
        \RHatFunc[{\VAR[0.1]}]{F} & if~ F(\RHatFunc[{\VAR[0.1]}]{F}) = 1
        \\
        \RHatFunc[+]{F; \RHatFunc[{\VAR[0.1]}]{F}} & otherwise.
    \end{cases} 
    \\
    &\RHatFunc[{\VAR[0.1]^{++}}]{F} 
    = 
    \begin{cases}
        \RHatFunc[{\VAR[0.1]^+}]{F} & if~ F(\RHatFunc[{\VAR[0.1]^+}]{F}) = 1 
        \\
        \RHatFunc[+]{F; \RHatFunc[{\VAR[0.1]^+}]{F}}  & otherwise.
    \end{cases}
\end{align*}
Consider a problem with two arms having the following distributions,
\begin{align*}
    \Fi[1](y)
    = 
    \begin{cases}
        0 & y < 0 
        \\
        0.9 + y/100 & 0 \le y < 10 
        \\
        1 & y \ge 10
    \end{cases}
    \qquad
    ,
    \Fi[2](y) 
    = 
    \begin{cases}
        0 & y < 1 
        \\
        0.1 & 1 \le y < 10 
        \\
        1 & y \ge 10
        .
    \end{cases}
\end{align*}
The performance of a \simple\space policy $\statPolicy$ $(p = (p_1,p_2),~ p_1 = 1 - p_2)$ is given by,
\begin{align*}
    \EE{\RPiNamed[{\statPolicy}]{bad2}} 
    = 
    \begin{cases}
        5 & p_2 < 8/9 
        \\
        -85 + 10 / (1-p_2) & 8/9 \le p_2 < 81/91 
        \\
        6 & 81/91 \le p_2 < 1 
        \\
        10 & p_2 = 1
        ,
    \end{cases}
\end{align*}
which attains a maximum for $p_2=1$. However, the resulting policy is not an oracle policy. The following non-\simple\space policy is an oracle policy,
\begin{align*}
    \policyAt[*bad2]{t} 
    = 
    \begin{cases}
        1 & t=1 
        \\
        2 & otherwise
        .
    \end{cases}
\end{align*}
To show this, proceed as for $\RHat[bad1]$, i.e., see that the individual components are bounded by $10$ and $5$ respectively, and so the optimal performance is at most $15$. It is trivial to verify that $\policy[*bad2]$ obtains this reward, thus showing it is an oracle policy.
Finally, we describe how $\EE{\RPiNamed[{\statPolicy}]{bad2}}$ is obtained. It is easily seen that for all $t \ge 1$,
\begin{align*}
    \indEvent{\FHatPi[\statPolicy]{t}({10}^{-}) - \FHatPi[\statPolicy]{t}({1}^{+}) > 0 ~or~ \FHatPi[\statPolicy]{t}({1}^{-}) > 0} 
    = 
    \begin{cases}
        1 & p_2 < 1 
        \\
        0 & p_2 = 1
        .
    \end{cases}
\end{align*}
As for $\RHatFunc[{\VAR[0.1]^{++}}]{\FHatPi[\statPolicy]{t}}$, when $p_2=1$ then for all $t \ge 1$ we have $\RHatFunc[{\VAR[0.1]^{++}}]{\FHatPi[\statPolicy]{t}} = 10$. If $p_2 < 1$ then the fact that $F^{(1)}$ is strictly increasing causes $\RHatFunc[{\VAR[0.1]^{++}}]{\FHatPi[\statPolicy]{t}}$ to converge to $\RHatFunc[{\VAR[0.1]}]{\FHatPi[\statPolicy]{t}}$. We conclude that,
\begin{align} \label{eq:Rbad2eq1}
    \EE{\RPiNamed[{\statPolicy}]{bad2}}
    = 
    \begin{cases}
        \EE{\RPiNamed[{\statPolicy}]{\VAR[0.1]}} + 5 & p_2 < 1 
        \\
        10 & p_2 = 1
        .
    \end{cases}
\end{align}
Calculating $\EE{\RPiNamed[{\statPolicy[p]}]{\VAR[0.1]}}$ and substituting it into \cref{eq:Rbad2eq1} yields the result.

These examples show that if an \EDRMabbrv\space is either not lower or upper semi-continuous then it might not have a \simple\space oracle policy. However, if it is both lower and upper semi-continuous then it is continuous and thus \cref{theorem:StableEDRMoptPolicy} typically holds.

\end{APPENDICES}

\end{document}